\documentclass[sigconf]{acmart}

\renewcommand\footnotetextcopyrightpermission[1]{} % removes footnote with conference information in first column
\usepackage[utf8]{inputenc}
\usepackage{xcolor}
\usepackage{amsmath}
\usepackage{threeparttable}

\usepackage[acronyms,nonumberlist,nopostdot,nomain,nogroupskip]{glossaries}
\usepackage{booktabs}

\usepackage{tabularx}

\usepackage{tikz}
\usepackage{pgfplots}
\pgfplotsset{compat=newest}
\pgfplotsset{plot coordinates/math parser=false}
\newlength\fheight
\newlength\fwidth
\usetikzlibrary{plotmarks,patterns,decorations.pathreplacing,backgrounds,calc,arrows,arrows.meta,spy,matrix}
\usepgfplotslibrary{patchplots,groupplots,colorbrewer}
\usepackage{tikzscale}
\newif\ifexttikz
\exttikzfalse
\ifexttikz
	\usetikzlibrary{external}
\fi

\usepackage{multirow}
\usepackage[font=footnotesize]{subcaption}
\usepackage[font=footnotesize]{caption}

\usepackage{mathtools}

\newacronym{6g}{6G}{sixth generation}
\newacronym{3gpp}{3GPP}{3rd Generation Partnership Project}
\newacronym{adc}{ADC}{Analog to Digital Converter}
\newacronym{dac}{DAC}{Digital to Analog Converter}
\newacronym{5g}{5G}{5th generation}
\newacronym{aimd}{AIMD}{Additive Increase Multiplicative Decrease}
\newacronym{am}{AM}{Acknowledged Mode}
\newacronym{amc}{AMC}{Adaptive Modulation and Coding}
\newacronym{aoa}{AoA}{Angle of Arrival}
\newacronym{aod}{AoD}{Angle of Departure}
\newacronym{aqm}{AQM}{Active Queue Management}
\newacronym{awgn}{AGWN}{Additive White Gaussian Noise}
\newacronym{balia}{BALIA}{Balanced Link Adaptation}
\newacronym{bdp}{BDP}{Bandwidth-Delay Product}
\newacronym{bf}{BF}{Beamforming}
\newacronym{fpga}{FPGA}{field-programmable gate array}
\newacronym{cc}{CC}{Congestion Control}
\newacronym{cdf}{CDF}{Cumulative Distribution Function}
\newacronym{cn}{CN}{Core Network}
\newacronym{cm}{CM}{confusion matrix}
\newacronym[plural=\gls{cnn}s,firstplural=convolutional neural networks (CNNs)]{cnn}{CNN}{convolutional neural network}
\newacronym{cqi}{CQI}{Channel Quality Information}
\newacronym{cp}{CP}{Control Plane}
\newacronym{csirs}{CSI-RS}{Channel State Information - Reference Signal}
\newacronym{dc}{DC}{Dual Connectivity}
\newacronym{dce}{DCE}{Direct Code Execution}
\newacronym{dci}{DCI}{Downlink Control Information}
\newacronym{dmr}{DMR}{Deadline Miss Ratio}
\newacronym{dmrs}{DMRS}{DeModulation Reference Signal}
\newacronym{e2e}{E2E}{End-to-End}
\newacronym{ecn}{ECN}{Explicit Congestion Notification}
\newacronym{ebs}{EBS}{exhaustive beam sweep}
\newacronym{edf}{EDF}{Earliest Deadline First}
\newacronym{enb}{eNB}{evolved Node Base}
\newacronym{epc}{EPC}{Evolved Packet Core}
\newacronym{es}{ES}{Edge Server}
\newacronym{fdma}{FDMA}{Frequency Division Multiple Access}
\newacronym{fdd}{FDD}{Frequency Division Duplexing}
\newacronym[firstplural=Radio Access Technologies (RATs)]{rat}{RAT}{Radio Access Technology}
\newacronym{fs}{FS}{Fast Switching}
\newacronym{txer}{TX}{transmitter}
\newacronym{rxer}{RX}{receiver}
\newacronym{bt}{BT}{beam tracking}
\newacronym{ftp}{FTP}{File Transfer Protocol}
\newacronym{gnb}{gNB}{Next Generation Node Base}
\newacronym{bs}{BS}{Base Station}
\newacronym{harq}{HARQ}{Hybrid Automatic Repeat reQuest}
\newacronym{hetnet}{HetNet}{Heterogeneous Network}
\newacronym{hh}{HH}{Hard Handover}
\newacronym{hol}{HOL}{Head-of-Line}
\newacronym{ia}{IA}{initial access}
\newacronym{imt}{IMT}{International Mobile Telecommunication}
\newacronym{iot}{IoT}{Internet of Things}
\newacronym{los}{LOS}{Line-of-Sight}
\newacronym{lte}{LTE}{Long Term Evolution}
\newacronym{m2m}{M2M}{Machine to Machine}
\newacronym{ml}{ML}{machine learning}
\newacronym{dl}{DL}{deep learning}
\newacronym{mac}{MAC}{Medium Access Control}
\newacronym{mc}{MC}{Multi-Connectivity}
\newacronym{mcs}{MCS}{Modulation and Coding Scheme}
\newacronym{mec}{MEC}{Mobile Edge Cloud}
\newacronym{mi}{MI}{Mutual Information}
\newacronym{mimo}{MIMO}{Multiple Input, Multiple Output}
\newacronym{mmwave}{mmWave}{millimeter wave}
\newacronym{mmWave}{mmWave}{Millimeter wave}
\newacronym{mptcp}{MPTCP}{Multipath TCP}
\newacronym{mr}{MR}{Maximum Rate}
\newacronym{mss}{MSS}{Maximum Segment Size}
\newacronym{mtd}{MTD}{Machine-Type Device}
\newacronym{mtu}{MTU}{Maximum Transmission Unit}
\newacronym{nfv}{NFV}{Network Function Virtualization}
\newacronym{nlos}{NLOS}{Non-Line-of-Sight}
\newacronym{nr}{NR}{New Radio}
\newacronym{ofdm}{OFDM}{Orthogonal Frequency Division Multiplexing}
\newacronym{pdcch}{PDCCH}{Physical Downlink Control Channel}
\newacronym{pdcp}{PDCP}{Packet Data Convergence Protocol}
\newacronym{pdsch}{PDSCH}{Physical Downlink Shared Channel}
\newacronym{pdu}{PDU}{Packet Data Unit}
\newacronym{pf}{PF}{Proportional Fair}
\newacronym{pgw}{PGW}{Packet Gateway}
\newacronym{phy}{PHY}{Physical}
\newacronym{pbch}{PBCH}{Physical Broadcast Channel}
\newacronym[plural=\gls{mme}s,firstplural=Mobility Management Entities (MMEs)]{mme}{MME}{Mobility Management Entity}
\newacronym{prb}{PRB}{Physical Resource Block}
\newacronym{pss}{PSS}{Primary Synchronization Signal}
\newacronym{pucch}{PUCCH}{Physical Uplink Control Channel}
\newacronym{pusch}{PUSCH}{Physical Uplink Shared Channel}
\newacronym{rach}{RACH}{Random Access Channel}
\newacronym{ran}{RAN}{Radio Access Network}
\newacronym{red}{RED}{Random Early Detection}
\newacronym{rf}{RF}{Radio Frequency}
\newacronym{rlc}{RLC}{Radio Link Control}
\newacronym{rlf}{RLF}{Radio Link Failure}
\newacronym{rrc}{RRC}{Radio Resource Control}
\newacronym{rrm}{RRM}{Radio Resource Management}
\newacronym{rr}{RR}{Round Robin}
\newacronym{rs}{RS}{Remote Server}
\newacronym{rsrp}{RSRP}{Reference Signal Received Power}
\newacronym{rss}{RSS}{Received Signal Strength}
\newacronym{rtt}{RTT}{Round Trip Time}
\newacronym{rw}{RW}{Receive Window}
\newacronym{rx}{RX}{Receiver}
\newacronym{sa}{SA}{standalone}
\newacronym{sack}{SACK}{Selective Acknowledgment}
\newacronym{sap}{SAP}{Service Access Point}
\newacronym{ap}{AP}{Access Point}
\newacronym{sch}{SCH}{Secondary Cell Handover}
\newacronym{scoot}{SCOOT}{Split Cycle Offset Optimization Technique}
\newacronym{sdma}{SDMA}{Spatial Division Multiple Access}
\newacronym{sinr}{SINR}{Signal to Interference plus Noise Ratio}
\newacronym{sm}{SM}{Saturation Mode}
\newacronym{snr}{SNR}{Signal-to-Noise-Ratio}
\newacronym{son}{SON}{Self-Organizing Network}
\newacronym{ss}{SS}{Synchronization Signal}
\newacronym{ssbs}{SSBs}{synchronization signal blocks}
\newacronym{ssb}{SSB}{synchronization signal block}
\newacronym{srs}{SRS}{Sounding Reference Signal}
\newacronym{sss}{SSS}{Secondary Synchronization Signal}
\newacronym{tb}{TB}{Transport Block}
\newacronym{tcp}{TCP}{Transmission Control Protocol}
\newacronym{tdd}{TDD}{Time Division Duplexing}
\newacronym{tdma}{TDMA}{Time Division Multiple Access}
\newacronym{tfl}{TfL}{Transport for London}
\newacronym{tm}{TM}{Transparent Mode}
\newacronym{trp}{TRP}{Transmitter Receiver Pair}
\newacronym{tti}{TTI}{Transmission Time Interval}
\newacronym{ttt}{TTT}{Time-to-Trigger}
\newacronym{tx}{TX}{Transmitter}
\newacronym{ue}{UE}{User Equipment}
\newacronym{ul}{UL}{Uplink}
\newacronym{uml}{UML}{Unified Modeling Language}
\newacronym{um}{UM}{Unacknowledged Mode}
\newacronym{utc}{UTC}{Urban Traffic Control}
\newacronym{vm}{VM}{Virtual Machine}
\newacronym{rsrq}{RSRQ}{Reference Signal Received Quality}
\newacronym{rssi}{RSSI}{Received Signal Strength Indicator}
\newacronym{crs}{CRS}{Cell Reference Signal}
\newacronym{nsa}{NSA}{Non Stand Alone}
\newacronym{mrdc}{MR-DC}{Multi \gls{rat} \gls{dc}}
\newacronym{endc}{EN-DC}{E-UTRAN-\gls{nr} \gls{dc}}
\newacronym{5gc}{5GC}{5G Core}
\newacronym{si}{SI}{Study Item}
\newacronym{iab}{IAB}{Integrated Access and Backhaul}
\newacronym{wf}{WF}{Wired-first}
\newacronym{hqf}{HQF}{Highest-quality-first}
\newacronym{pa}{PA}{Position-aware}
\newacronym{mlr}{MLR}{Maximum-local-rate}
\newacronym{wbf}{WBF}{Wired Bias Function}
\newacronym{mib}{MIB}{Master Information Block}
\newacronym{sib}{SIB}{Secondary Information Block}
\newacronym{kpi}{KPI}{Key Performance Indicator}
\newacronym{ppp}{PPP}{Poisson Point Process}
\newacronym{gtp}{GTP}{GPRS Tunneling Protocol}
\newacronym{amf}{AMF}{Access and Mobility Management Function}
\newacronym{dash}{DASH}{Dynamic Adaptive Streaming over HTTP}
\newacronym{http}{HTTP}{HyperText Transfer Protocol}
\newacronym{qos}{QoS}{Quality of Service}
\newacronym{udp}{UDP}{User Datagram Protocol}
\newacronym{cu}{CU}{Central Unit}
\newacronym{du}{DU}{Distributed Unit}
\newacronym{mt}{MT}{Mobile Termination}
\newacronym{sdap}{SDAP}{Service Data Adaptation Protocol}
\newacronym{tdm}{TDM}{Time Division Multiplexing}
\newacronym{fdm}{FDM}{Frequency Division Multiplexing}
\newacronym{sdm}{SDM}{Space Division Multiplexing}
\newacronym{st}{ST}{Spanning Tree}
\newacronym{ummimo}{UM-MIMO}{Ultra-massive Multiple Input, Multiple Output}
\newacronym{uavs}{UAVs}{Unmanned Aerial Vehicles}
\newacronym{wlan}{WLAN}{Wireless LAN}
\newacronym{rlnc}{RLNC}{Random Linear Network Coding}
\newacronym{drx}{DRX}{Discontinuous Reception}
\newacronym{cpu}{CPU}{Central Processing Unit}
\newacronym{txb}{TXB}{transmitter's beam}
\newacronym{rxb}{RXB}{receiver's beam}
\newacronym{sifs}{SIFS}{Short Interframe Space}
\newacronym{difs}{DIFS}{DCF Interframe Space}

\newacronym{rfid}{RFID}{Radio Frequency Identification}
\newacronym{rfp}{RFP}{radio fingerprinting}
\newacronym{sdr}{SDR}{software-defined radio}
\newacronym{fml}{FML}{federated machine learning}
\newacronym{da}{DAG}{data augmentation}
\tikzstyle{startstop} = [rectangle, rounded corners, minimum width=2cm, minimum height=0.5cm,text centered, draw=black]
\tikzstyle{io} = [trapezium, trapezium left angle=70, trapezium right angle=110, minimum width=3cm, minimum height=1cm, text centered, draw=black]
\tikzstyle{process} = [rectangle, minimum width=2cm, minimum height=0.5cm, text centered, draw=black, alignb=center]
\tikzstyle{decision} = [ellipse, minimum width=2cm, minimum height=1cm, text centered, draw=black]
\tikzstyle{arrow} = [thick,<->,>=stealth]
\tikzstyle{line} = [thick,>=stealth]
\tikzstyle{darrow} = [thick,<->,>=stealth,dashed]
\tikzstyle{sarrow} = [thick,->,>=stealth]
\tikzstyle{larrow} = [line width=0.1mm,dashdotted,->,>=stealth]

\makeatletter
\def\grd@save@target#1{%
  \def\grd@target{#1}}
\def\grd@save@start#1{%
  \def\grd@start{#1}}
\tikzset{
  grid with coordinates/.style={
    to path={%
      \pgfextra{%
        \edef\grd@@target{(\tikztotarget)}%
        \tikz@scan@one@point\grd@save@target\grd@@target\relax
        \edef\grd@@start{(\tikztostart)}%
        \tikz@scan@one@point\grd@save@start\grd@@start\relax
        \draw[minor help lines] (\tikztostart) grid (\tikztotarget);
        \draw[major help lines] (\tikztostart) grid (\tikztotarget);
        \grd@start
        \pgfmathsetmacro{\grd@xa}{\the\pgf@x/1cm}
        \pgfmathsetmacro{\grd@ya}{\the\pgf@y/1cm}
        \grd@target
        \pgfmathsetmacro{\grd@xb}{\the\pgf@x/1cm}
        \pgfmathsetmacro{\grd@yb}{\the\pgf@y/1cm}
        \pgfmathsetmacro{\grd@xc}{\grd@xa + \pgfkeysvalueof{/tikz/grid with coordinates/major step x}}
        \pgfmathsetmacro{\grd@yc}{\grd@ya + \pgfkeysvalueof{/tikz/grid with coordinates/major step y}}
        \foreach \x in {\grd@xa,\grd@xc,...,\grd@xb}
        \node[anchor=north] at (\x,\grd@ya) {\pgfmathprintnumber{\x}};
        \foreach \y in {\grd@ya,\grd@yc,...,\grd@yb}
        \node[anchor=east] at (\grd@xa,\y) {\pgfmathprintnumber{\y}};
      }
    }
  },
  minor help lines/.style={
    help lines,
    gray,
    line cap =round,
    xstep=\pgfkeysvalueof{/tikz/grid with coordinates/minor step x},
    ystep=\pgfkeysvalueof{/tikz/grid with coordinates/minor step y}
  },
  major help lines/.style={
    help lines,
    line cap =round,
    line width=\pgfkeysvalueof{/tikz/grid with coordinates/major line width},
    xstep=\pgfkeysvalueof{/tikz/grid with coordinates/major step x},
    ystep=\pgfkeysvalueof{/tikz/grid with coordinates/major step y}
  },
  grid with coordinates/.cd,
  minor step x/.initial=.5,
  minor step y/.initial=.2,
  major step x/.initial=1,
  major step y/.initial=1,
  major line width/.initial=1pt,
}
\makeatother

\definecolor{desireRed}{RGB}{230,57,60}%
\definecolor{darkPurple}{RGB}{59,31,43}%
\definecolor{springGreen}{RGB}{37,223,145}%
\definecolor{queenBlue}{RGB}{69,123,157}%
\definecolor{spaceCadet}{RGB}{29,53,87}%
\newcommand{\datasetname}{RFID16-2021}

\acmYear{2021}
\copyrightyear{2021} 
\setcopyright{rightsretained} 
\acmConference[Mobihoc '21]{The Twenty-first ACM International Symposium on Theory, Algorithmic Foundations, and Protocol Design for Mobile Networks and Mobile Computing}{July 26--30, 2021}{Shanghai, China} 
\acmBooktitle{The Twenty-second ACM International Symposium on Theory, Algorithmic Foundations, and Protocol Design for Mobile Networks and Mobile Computing (Mobihoc '21), July 26--30, 2021, Shanghai, China} 
\begin{document}

\title{The Tags Are Alright: Robust Large-Scale RFID Clone Detection Through Federated Data-Augmented Radio Fingerprinting \vspace{-0.3cm}}

\author{Mauro Piva$^\dagger$, Gaia Maselli$^\dagger$, and Francesco Restuccia$^*$}
\affiliation{%
  \institution{$^\dagger$Department of Computer Science, Sapienza University, Italy \\
  $^*$Department of Electrical and Computer Engineering, Northeastern University, United States}
}

\glsresetall

\begin{abstract}
Millions of RFID tags are pervasively used all around the globe to inexpensively identify a wide variety of everyday-use objects. One of the key issues of RFID is that tags cannot use energy-hungry cryptography, and thus can be easily cloned. For this reason, \gls{rfp} is a compelling approach that leverages the unique imperfections in the tag's wireless circuitry to achieve large-scale RFID clone detection. Recent work, however, has unveiled that time-varying channel conditions can significantly decrease the accuracy of the \gls{rfp} process. Prior art in RFID identification does not consider this critical aspect, and instead focuses on custom-tailored feature extraction techniques and data collection with static channel conditions. For this reason, we propose the first large-scale investigation into \gls{rfp} of RFID tags with dynamic channel conditions.  Specifically, we perform a massive data collection campaign on a testbed composed by 200 off-the-shelf identical RFID tags and a \gls{sdr} tag reader. We collect data with different tag-reader distances in an over-the-air configuration. To emulate implanted RFID tags, we also collect data with two different kinds of porcine meat inserted between the tag and the reader. We use this rich dataset to train and test several \gls{cnn}-based classifiers in a variety of channel conditions. Our investigation reveals that training and testing on different channel conditions drastically degrades the classifier's accuracy. For this reason, we propose a novel training framework based on \gls{fml} and \gls{da} to boost the accuracy. Extensive experimental results indicate that (i) our \gls{fml} approach improves accuracy by up to 48\%; (ii) our \gls{da} approach improves the \gls{fml} performance by up to 19\% and the single-dataset performance by 31\%. To the best of our knowledge, this is the first paper experimentally demonstrating the efficacy of \gls{fml} and \gls{da} on a large device population. To allow full replicability, we are sharing with the research community our fully-labeled 200-GB RFID waveform dataset, as well as the entirety of our code and trained models, concurrently with our submission.\vspace{-0.3cm}
\end{abstract}




% %%
% %% Keywords. The author(s) should pick words that accurately describe
% %% the work being presented. Separate the keywords with commas.
% \keywords{datasets, neural networks, gaze detection, text tagging}

\maketitle

\glsresetall

\section{Introduction}

The cost-effectiveness of \gls{rfid} tags is becoming a fundamental driver for their significant expansion into the  \gls{iot} ecosystem \cite{welbourne2009building,al2015internet}. In short, an \gls{rfid} tag consists of a tiny radio transceiver, usually consisting of a few thousands logic gates \cite{ranasinghe2006security}, with a total cost reaching as low as few cents \cite{saila2014magic}. When triggered by an electromagnetic interrogation pulse from a nearby \gls{rfid} reader, the tag transmits digital data back to the reader. Nowadays, \glspl{rfid} are so pervasive that it is hard to realize they are around us. Among others, applications include object monitoring \cite{chung2008rfid,chen2006system,bridgelall2006object,maselli2019battery}, mobile payments \cite{qadeer2009novel}, and continuous health monitoring through intrabody implantation \cite{scotti2020body}. Attesting to their ever-increasing popularity, a recent study indicates that the \gls{rfid} market will reach \$16.23B by 2029, a 34\% increase in 10 years \cite{RFIDMarket}. 

Since \gls{rfid} can be attached to cash and other valuable objects, and also implanted into animals and people, their widespread usage has raised serious security and privacy concerns \cite{rfidCash,juels2006rfid,ahson2017rfid}. Critically, low-cost \gls{rfid} tags cannot support complex cryptography beyond hash functions \cite{mansoor2019securing}, and lightweight cryptographic techniques have been proven as insecure \cite{safkhani2014cryptanalysis,tan2008secure,zhou2010lightweight}.  For this reason, \gls{rfid} tags can be easily tampered with and their functionality compromised \cite{WiredHackRFID,wang2018challenge}. In this paper, we consider the \textit{cloning} security vulnerability, where tag data is eavesdropped during the reading process by an adversary and then replayed by a rogue device, for example, a \gls{sdr} \cite{ai2020nowhere}. It is intuitive that the resilience of \gls{rfid} tags to cloning attacks is strongly correlated to their applicability in critical applications. For example, if thousands of cloned tags are injected into the supply chain, companies would lose track of their assets and thus sustain severe financial loss \cite{maleki2017new}. Perhaps even more worrisome, tags cloning could have serious consequences to the well-being of individuals, as they are extensively used in credit cards, passports, badges, and health care, among others \cite{bu2017you}.

\begin{figure}[!h]
    \centering
    \vspace{-0.2cm}
  \includegraphics[width=\columnwidth]{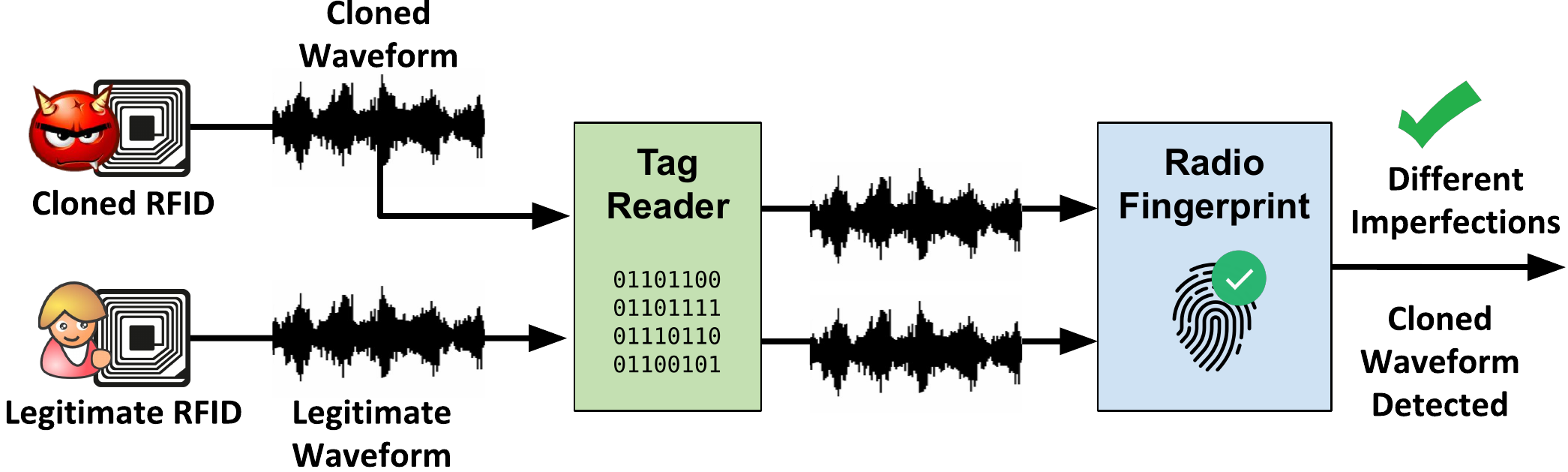}
  \caption{Radio fingerprinting leverages unique circuitry imperfections that are hardly replicable by an adversary to identify cloned RFID tags.  }\label{fig:rfid_finger} 
    \vspace{-0.3cm}
\end{figure}

\textbf{Approach.}~To address the tag cloning issue without relying on cryptography, some prior work has proposed approaches based on \gls{rfp} \cite{danev2009physical,romero2009electromagnetic,periaswamy2010fingerprinting,zanetti2010physical,zanetti2011practicality,bertoncini2011wavelet,danev2012towards,han2015geneprint,chen2018combating}. Figure \ref{fig:rfid_finger} summarizes the \gls{rfid} tag cloning issue and how \gls{rfp} addresses the issue. In short, \gls{rfp} leverages small-scale hardware-level imperfections typically found in off-the-shelf RFID front-end circuitry, such as frequency and sampling offset, I/Q imbalance, phase noise, and harmonic distortions \cite{johnson1966physical}. By estimating the impairments on the received waveform and associating them to a given device, a unique identification of the device can be obtained \cite{1_xu2016device}. Prior work  -- discussed in details in Section \ref{sec:rw} -- relies on protocol-specific feature-extraction techniques such as dynamic wavelet fingerprint \cite{bertoncini2011wavelet} or minimum power response at multiple frequencies \cite{periaswamy2010fingerprinting} to extract hardware impairments, which can only be applied to a given family of \gls{rfid} tags. In stark contrast, in this paper we leverage \gls{ml}-based techniques -- specifically, \glspl{cnn} -- to create feature-agnostic, general-purpose, and optimizable \gls{rfp} classifiers \cite{5_8466371}. The key advantages of \glspl{cnn} with respect to traditional \gls{ml} are that (i) \glspl{cnn} employ a very high number of parameters and thus can distinguish very high population of devices; and (ii) by learning filters operating over unprocessed I/Q samples, they avoid application-specific computational-expensive feature extraction/selection algorithms \cite{restucciacommag2020,OShea-ieeejstsp2018}. 

\textbf{Existing Issues.}~Prior research has hinted that dynamic propagation environments may cause the CNN's accuracy to plummet significantly \cite{shawabka2020exposing}. To validate the severity of the problem, \textit{we experimentally show in Section \ref{sec:learning_channel} that training and testing on different datasets reduce accuracy by 90\% on the average.} This is not without a reason. Indeed, one of key assumptions of \glspl{cnn}, is that training and testing datasets are independent and identically distributed (i.i.d.). On the other hand, this is hardly the case in the wireless domain, where (i) time-varying oscillations in temperature and voltage may cause the hardware impairments to change over time; and (ii) interference and noise levels can modify waveforms in dynamic and unpredictable manner. Existing strategies leverage the computation of finite input response (FIR) filters applied at the transmitter's side to partially compensate the channel action \cite{restuccia2019deepradioid}. However, this strategy cannot be used for \gls{rfid}, \textit{since tags are not software-defined and cannot modify their physical-layer waveforms in any circumstance}. Also, these strategies maximize the fingerprinting accuracy for a given wireless node only, while we want to improve the performance of the whole population of \gls{rfid} tags. For this reason, we need to find alternative strategies. This critical issue is further complicated by the \textit{current lack of rich, large-scale datasets for \gls{rfp} of \gls{rfid} tags, which makes the replication of existing research impossible}. It is easy to observe that without a common benchmark, all the innovation in the field will be stymied, since every paper can claim to be ``better than the previous one''.  This calls for the generation of public-domain datasets that can be used by the research community at large.

\textbf{Technical Contributions.}~We summarize the novel contributions to the state of the art provided by this paper:

$\bullet$ We perform a massive data collection campaign on a testbed composed by 200 off-the-shelf identical EPC RFID tags \cite{global2008epc} and a USRP2 \gls{sdr} tag reader. We collect data with different tag-reader distances (20cm, 50cm and 100cm) in an over-the-air configuration. To obtain even more challenging propagation scenarios and emulate implanted RFID tags, we also collect data with two different kinds of porcine meat (i.e., with different texture and thickness) inserted between the tag and the reader, and two reader distances (20cm and 50cm). This dataset is used to train and test several \gls{cnn}-based classifiers based on classic cross-entropy minimization. Our experimental results on the collected 7 datasets reveals that training and testing on different channel conditions decreases the \gls{rfp} performance by up to 90\%; 

$\bullet$ We propose a federated data-augmented training framework for \gls{rfid} fingerprinting to address the lack of generalization of standard techniques. Our framework is based on \gls{fml}, which has recently emerged as a powerful tool at the intersection of artificial intelligence and edge computing \cite{mcmahan2017communication}. In short, \gls{fml} allows the periodical fusion of locally-trained models by averaging the parameters over a number of training epochs, thus eliminating the need to stream waveforms captured by the tag readers to a centralized server.  \textit{To the best of our knowledge, no prior work has ever explored the usage of \gls{fml} for \gls{rfp} improvement}. Our framework also leverages the concept of \gls{da} to improve the robustness of the original models \cite{perez2017effectiveness,shorten2019survey,mikolajczyk2018data}. In short, the key rationale of \gls{da} is that by inserting into the dataset ``noisy'' inputs obtained by modified original inputs, the classifier will be stronger to changing spectrum environments;

$\bullet$ We perform an exhaustive training and testing campaign where our two training strategies are extensively evaluated and compared with baselines. Experiments results indicate that (i) our \gls{fml}-based training improves accuracy by up to 48\% with respect to the single-dataset scenario; and (ii) our \gls{da}-based training improves the \gls{fml} performance by up to 19\% and the global performance by 31\%. \textit{To the best of our knowledge, this is the first paper demonstrating the efficacy on such rich datasets and large population of devices}. In stark contrast with existing work, \textit{we pledge to share with the research community our fully-labeled 200-GB RFID waveform dataset, as well as the entirety of our code and trained models.} This will allow complete replicability and verification of results as well as a benchmark for further work in the field.\vspace{-0.3cm}

\section{Background and Related Work}\label{sec:rw}

Passive \gls{rfid} tags are divided into three categories, depending on the frequency of operation: 125-134 KHz, also known as Low Frequency (LF), 13.56 MHz, also known as High Frequency (HF), and 865-928 MHz, also known as Ultra High Frequency (UHF). The functionality of RFID is simple in nature; when a tag is scanned by a reader, the reader transmits energy to the tag which powers it enough for the chip and antenna to relay information back to the reader. UHF frequencies typically offer much better read range -- between 1 and 12 meters, according to the setup -- and can transfer data faster. Thus, UHF-based tags are among the most popular type of RFID technology \cite{want2006introduction}, and are thus the focus of our investigation. Differently from active and semi passive devices, these are equipped with no battery, and are strongly power constrained. In order to work these tags need to be powered by a reader, which allows the tags to get the required energy to produce simple elaborations and to communicate through backscattering.

Radio fingerprinting (RFP) has been extensively used on the WiFi \cite{10_brik2008wireless,12_vo2016fingerprinting,shawabka2020exposing,restuccia2019deepradioid} and ZigBee standards \cite{16-Peng-ieeeiotj2018}.  \gls{rfid} anti-cloning through \gls{rfp} has also been investigated over the last years 
\cite{danev2009physical,romero2009electromagnetic,periaswamy2010fingerprinting,zanetti2010physical,zanetti2011practicality,bertoncini2011wavelet,danev2012towards,han2015geneprint,chen2018combating}. For an excellent and exhaustive survey on the topic, the reader is referred to \cite{bu2017you}. The seminal work \cite{danev2009physical} applies \gls{rfp} to 50 HF tags, achieving 2.43\% error rate. Among others, the features used are based on the Hilbert transform. Romero \emph{et al} \cite{romero2009electromagnetic} analyze how electromagnetic signatures can help detect counterfeit HF tags. However, their analysis leverages an oscilloscope with a sampling rate of 20 GHz, which is beyond the capability of off-the-shelf readers. Periaswamy \emph{et al.} \cite{periaswamy2010fingerprinting} leverage the minimum power response at multiple frequencies (MPRMF)x to distinguish RFID tags, achieving an accuracy of
90.5\% with a population of 100 tags. Zanetti \emph{et al.} \cite{zanetti2010physical} use time- and spectrum-level domain to  fingerprint 70 UHF tags, achieving 71\% accuracy.  Bertoncini \emph{et al.} \cite{bertoncini2011wavelet} consider 146 tags belonging to 3 different manufacturers, with a sampling rate of 1.5MS/s. Protocol-specific techniques such as wavelet packet decomposition and higher order statistics are leveraged to transform the signal into an image which is then processed by a support vector machine (SVM). Conversely from us, they leverage the full EPC transmission, while we focus on the RN16 portion to avoid learning protocol-dependent features. More recently, Han \emph{et al.} \cite{han2015geneprint} proposed a study where tag-reader distance and orientation are considered in evaluating the robustness of the \gls{rfp} process. However, these experiments were conducted with a low population of 30 devices, while we consider 200 tags in this paper. Chen \emph{et al.} proposed a large-scale study by considering tags using C1G2 standard \cite{chen2018combating}.  Unfortunately, we cannot compare our learning-based approach with \cite{han2015geneprint,chen2018combating} and the rest of previous work since the datasets were not shared with the community. 

Convolutional neural networks (CNNs) have been recently used for a number of different wireless applications \cite{luong2018applications,jagannath2019machine,restucciacommag2020}. Although some work has explored the usage of \glspl{cnn} for \gls{rfp} purposes \cite{sankhe2019no,shawabka2020exposing,riyaz2018deep,restuccia2019deepradioid}, to the best of our knowledge no prior work has explored the usage of \glspl{cnn} for \gls{rfp} of \gls{rfid} tags. Moreover, as far as we know, no work has proposed the usage of federated machine learning (FML) to improve the performance of \gls{cnn}-based \gls{rfp}.  Although this problem is extremely relevant, only recently it has received attention from the community \cite{restuccia2019deepradioid,xie2020data,soltani2020more}. Although very similar in target, the solution proposed in \cite{restuccia2019deepradioid} assumes that the transmitter is able to somehow modify the transmitted waveform, which is not the case for \gls{rfid} tags since they are passive devices. Most importantly, the approach in \cite{restuccia2019deepradioid} optimizes the accuracy for a single device at a time. In this paper, we propose techniques that improve the model accuracy for all the devices in the dataset. Closer to our paper, Xie \emph{et al.} \cite{xie2020data} and Soltani \emph{et al.} \cite{soltani2020more} presented \gls{da} schemes for \gls{rfp}. The latter introduced a scheme that involves \gls{da} at the transmitter's side, which is not applicable to RFID tags. Moreover, the schemes in \cite{xie2020data,soltani2020more}  are validated on simulated datasets with a population of only 10 devices. Neither of the datasets are shared, including the large-scale dataset in \cite{soltani2020more}, which makes the results not replicable.

\textit{
To the best of our knowledge, ours is the first paper proposing training techniques for \gls{rfp} models based on \gls{fml} and \gls{da} that are validated with a large-scale experimental dataset of 200 devices.  Departing from prior work in the field, we are the first to share all the datasets, the code and trained models for complete replicability and results verification concurrently with the submission of the paper.\vspace{-0.3cm}}

\section{Proposed Training Framework}

We now present our framework for federated data-augmented radio fingerprinting. We first describe the considered neural network in Section \ref{sec:nn_descr}, then we introduce federated \gls{rfp} and data-augmented \gls{rfp} in Section \ref{sec:fed_rfp} and \ref{sec:augmentation}, respectively.

\subsection{Neural Network Description}\label{sec:nn_descr}

In line with recent existing work on WiFi \gls{rfp}  \cite{restuccia2019deepradioid,sankhe2019oracle,sankhe2019no}, we leverage a convolutional neural network (CNN) to extract and classify the unique imperfections of \gls{rfid} tags. Specifically, a convolutional layer (CVL) drastically reduces the number of trainable parameters in dense networks by computing filters that ``move'' across an input tensor and ``react'' to a given pattern in the input \cite{lecun1989generalization}.  More formally, by defining $d$ and $w$ as depth and width, a convolutional layer consists of a set of $F$ filters $\mathbf{Q}_f \in \mathbb{R}^{d \times w}, 1 \le f \le F$, where $F$ is also called the layer depth. Each filter generates  a \textit{feature map} $\mathbf{Y}^f \in \mathbb{R}^{n' \times m'}$  from an input matrix $\mathbf{X} \in \mathbb{R}^{n \times m}$ according to the following:

\begin{equation}
    \mathbf{Y}^{f}_{i,j} = \sum_{k=0}^{h-1}\sum_{\ell=0}^{w-1} \mathbf{Q}^f_{h-k,w-\ell} \cdot \mathbf{X}_{1+s\cdot(i-1)-k, 1+s\cdot(j-1)-\ell}
    \label{eq:filters}
\end{equation}

\noindent where $1 \le i \le n'$ and $1 \le j \le m'$. For simplicity, Equation \eqref{eq:filters} assumes input and filter dimension equal to 2. This formula can be generalized for tensors having dimension greater than 2. The $s \ge 1$ is called \textit{stride}, $n' = 1 + \left \lfloor{n + d - 2}\right \rfloor$ and $m' = 1 + \left \lfloor{m+w-2}\right \rfloor$. The matrix $\mathbf{X}$ is assumed to be padded with zeros, \textit{i.e.}, $X_{ij} = 0\ \forall i \not\in [1, n],\ j \not\in [1, m]$. For more details on CNNs, the reader may refer to \cite{goodfellow2016deep}.

\begin{figure}[h!]
    \centering
  \includegraphics[width=\columnwidth]{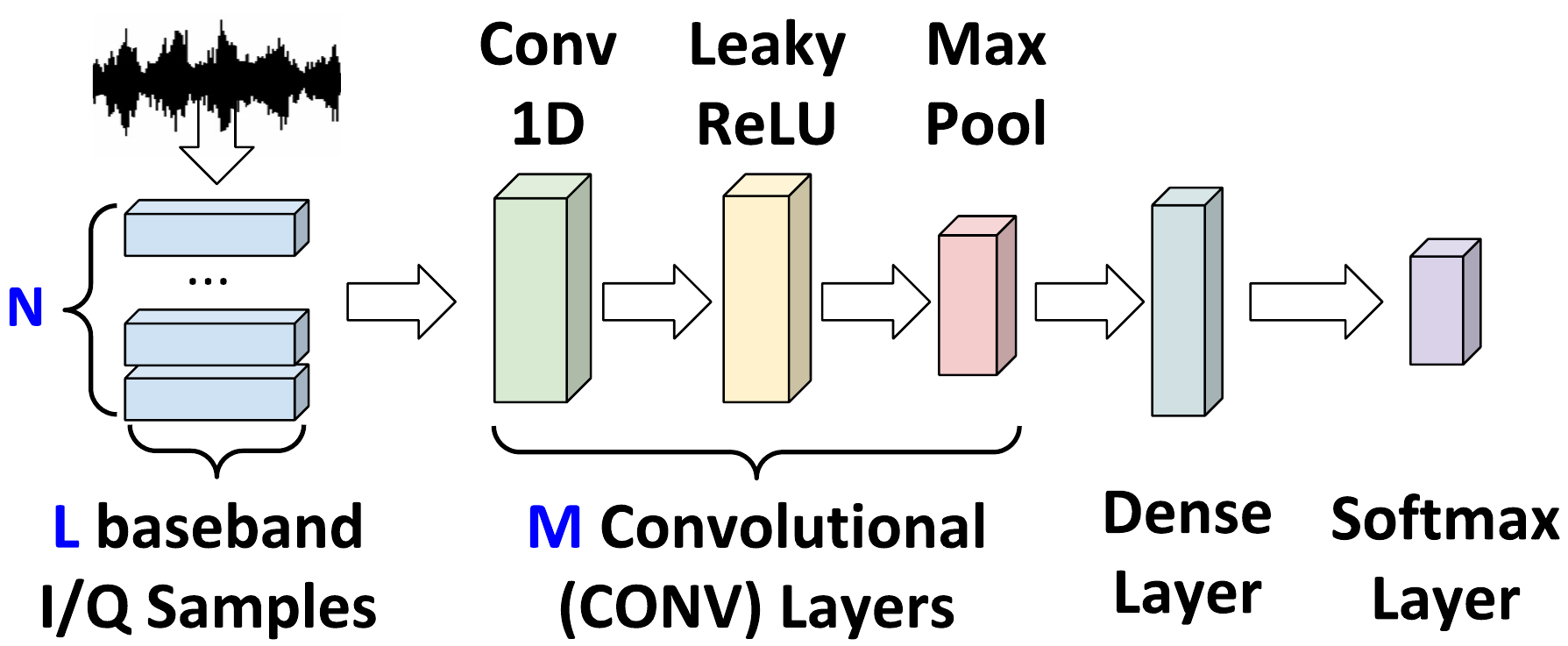}
  \caption{Our CNN model for RFID RFP. }\label{fig:nn_model} 
\end{figure}

Figure \ref{fig:nn_model} depicts the structure of the \gls{cnn} we utilize. Our proposed CNN has been implemented in Pytorch and is available at \cite{dataset-website}.  Our network takes as input a tensor of NxL size,  where $L$ is the number of baseband I/Q samples in each slice, $N$ is the number of available signal slices. To adapt the \gls{cnn} concept to our problem, as well as reducing computational burden and number of parameters, we will use a series of convolutional (CONV) layers, each composed by a Conv1D layer, a leaky rectified linear unit (LeakyReLU) as activation function, \textit{i.e.}, $ \sigma(x) = 0.1 \cdot x$ if $x<0$ and $x$ otherwise, and a \textit{max pooling layer} (MaxPool), which computes the maximum value out of $1 \times 2$ regions of the output (and thus, cuts the output in half). The input of the network is thus convolved by a series of $M$ CONV layers. Conv1D layers are composed by 25 filters with kernel size 3.

The last MaxPool is connected to a dense layer which has as input the dimension of the last MaxPool and as output the size of the population of RFID tags. We also add a flattening layer, which converts the data into a 1-dimensional vector. This vector is finally fed to the fully-connected dense layer, which is connected to a softmax layer. The loss value of each iteration is calculated with the Cross Entropy Loss, which represents a combination of a softmax and a negative log likelihood, and can be formalized as
% %
% \begin{equation*}
% loss(pred, target) = - pred[target] + 
% log \left( \sum_{i=0}^{len(pred)} e^{pred[i]} \right)
% \end{equation*}

% with $pred$ as the array of the scores of each class, and target as the index of the correct class.

\begin{equation}
\mathcal{L}(\Psi, \tau) = - \Psi[\tau] + 
log \left( \sum_{i=0}^{|\Psi|} e^{\Psi[i]} \right)
\end{equation}

\noindent with $\Psi$ as the array of the scores of each class, and $\tau$ as the index of the correct class. This means that effectively, in the released implementation, the softmax layer is not included in the network definition, but is executed at the loss calculation. We train our network using an Adam optimizer \cite{kingma2014adam}  with a learning rate of 0.001. The network has been trained for up to 100 epochs.

\subsection{Federated Radio Fingerprinting}\label{sec:fed_rfp}

To boost the accuracy of locally-trained \gls{cnn} models, we propose a training framework based on the concept of federated machine learning (FML). Specifically, \gls{fml} attempts to address distributed learning problems where the data is non identical and independently distributed (i.i.d.). Our key intuition is to leverage this capability to address the channel problem in \gls{rfp}, since any particular tag reader's local dataset cannot capture all the possible channel distributions. Finally, \gls{fml} is perfect in scenarios where tag readers are handled by different entities, which may not want to share their local dataset for privacy reasons.

Another key reason is that training the model locally and sending only the parameters reduces drastically the bandwidth utilization, which can be a critical parameter in an edge-based environment. To give an example, in this paper we sample the RFID signal at 5 mega samples per second (MS/s) and produce for each sample a complex number representing the I/Q value, which occupies 4 bytes. This yields a throughput of 20 MB/s. Such a throughput would severely congestion the network, especially in an IoT scenario. The  weights of our models occupy up to 2.48 MB, and are only exchanged periodically as explained below.

\begin{figure}[h!]
    \centering
  \includegraphics[width=\columnwidth]{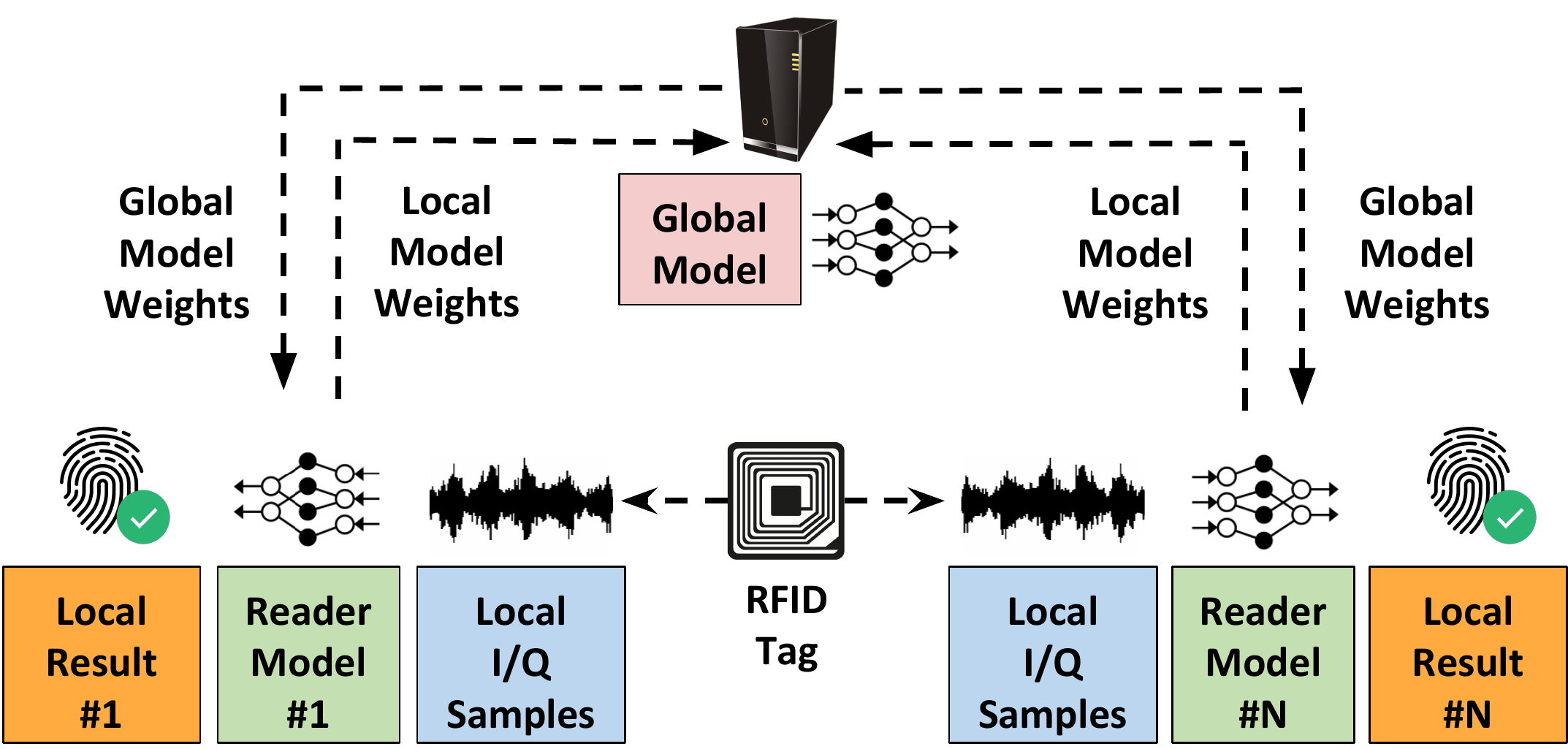}
  \caption{Federated Radio Fingerprinting framework.\vspace{-0.3cm} }\label{fig:nn_federated} 
\end{figure}

Figure \ref{fig:nn_federated} summarizes our federated \gls{rfp} approach.
We consider a scenario where $R$ tag readers (i.e. clients) collect I/Q samples from a tag, train a local model $M_r$ and adopt the following synchronous update schema.
Intuitively, at each round all the readers perform a training of the local model $M_r$ with the locally available data. Each reader then sends the weights of $M_r$ to a central server. The weights received from all the readers are averaged by the server and then sent back to the tag readers, which update the local models with the updated weights.
\\
More formally, for the training we assume that each reader $r \in R$ has a set of inputs $X_r$ and the corresponding set of labels $Y_r$, and that $X = \bigcup_{r\in R} X_r$, $Y = \bigcup_{r\in R} Y_r$.
The target of the server is to minimize a global loss function $f(w) = loss(X,Y,w)$, where $loss$ refers to the prediction done on $(X, Y)$ with weights $w$.
Without \gls{rfp}, we could describe the global loss function with the next two equations:

\begin{equation}
\min_{w \in \Re^d} f(X,Y,w) 
\quad \mathrm{and} \quad
f(X,Y,w) = \frac{1}{|X|} \sum_{(x,y) \in (X,Y)} f(x, y, w)
\label{eq:nn1}
\end{equation}

Introducing the \gls{rfp}, we can reformulate formula \ref{eq:nn1} as follows:

\begin{equation}
f(w) = \sum_{r \in R} \frac{|X_r|}{|X|} F_i(X_r, Y_r, w)
\quad \mathrm{with} 
\label{eq:nn2}
\end{equation}.

\begin{equation}
F_i(X_r, Y_r, w) = \frac{1}{|X_r|} \sum_{(x, y) \in (X_r,Y_r)} f(x,y, w)
\end{equation}

In our implementation of the federated averaging we expect the whole set of readers to be always active, as differently from other federated scenario they are not power constrained. For each tag reader we defined a fixed learning rate $\eta = 0.001$. Each tag reader then computes $fp_r = \nabla F_r(w_t)$, as the gradient obtained with the weight $w_t$ at time $t$.
The central server aggregates these weights producing $$w_{t+1}= \sum_{r \in R} \frac{1}{R} F_r(w_t).$$ This means that for each round of the federated learning all the readers run an epoch of the model training, and then share the resulting weights with the central server.
This approach however produces high instability through the first epochs, with the local accuracy which is drastically reduced at every federated average. 
In order to early stabilize the model, we let a single reader produce an initial model training for up to 10 epochs. 
This model is then shared with the other readers, and used as initialization for the training, letting local models start from this early trained model, and not from an empty one.
Starting local models from a random state seems to produce local models that have a very weak result when federated. 
This bootstrap produced an improvement in the number of epochs required to make the accuracy converge, with a reduction of up to 20 epochs.
We also evaluated a trade off between data segregation and sharing, sharing a subset of all the data and centrally training a model, which is then used as starting point for the local model training. 
While this change produces an increased initial improvement on the accuracy, it does not produces an effective advantage, as both the models with data shared and without converge to a comparable accuracy.

\emph{We show in Section \ref{sec:federated_results} that our federated \gls{rfp} improves accuracy by up to 48\% with respect to a local-only learning scenario.}

\subsection{Data-Augmented Radio Fingerprinting}\label{sec:augmentation}

The federated learning approach introduced in Section \ref{sec:fed_rfp} is able to improve the robustness of a model merging data coming from multiple sources. To further improve the robustness of local models and learn channel- and interference-independent features, we leverage federated data augmentation (DAG). \textit{We will show in Section \ref{sec:aug_results} that through our approach, we are able to make our model more robust, and to dramatically improve the performance of the federated approach up to 19\%}.

\gls{da} is a machine learning technique that produces a slightly modified version of a dataset to increase its size and improve robustness \cite{mikolajczyk2018data,shorten2019survey,perez2017effectiveness}. To the best of our knowledge, this is the first time that this technique is applied to RFID tag fingerprinting and with a population of 200 devices. To perform \gls{rfp}, it is possible to exploit \gls{da} on both the devices involved in a communication, i.e. tag and reader \cite{soltani2020more}. However, to apply data augmentation on the tag side we would have to modify it, reducing the real applicability of this work.

\begin{figure}[h!]
    \centering
  \includegraphics[width=0.9\columnwidth]{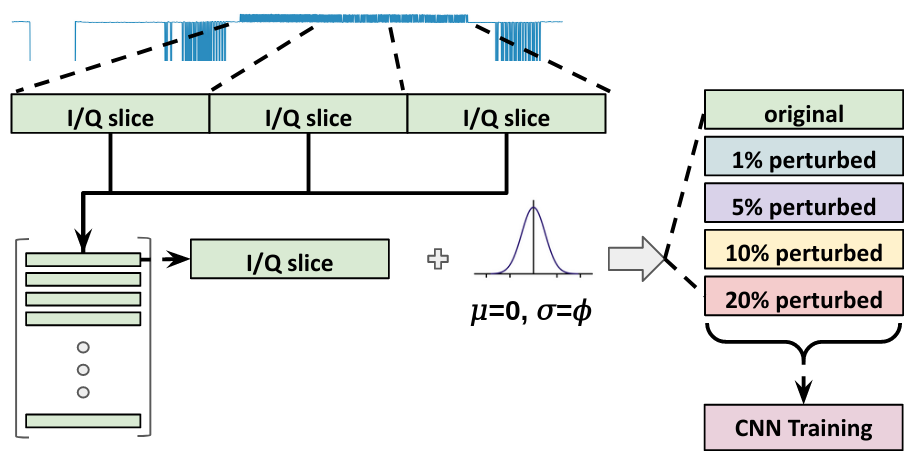}
  \caption{Data-augmented Radio Fingerprinting. }\label{fig:nn_augmentation} 
\end{figure}

Figure \ref{fig:nn_augmentation} shows our data-augmented \gls{rfp} approach. As we do not want to modify commercial RFID tags, we can apply data augmentation only at the tag reader's side, and we did it through Additive White Gaussian Noise (AWGN). Specifically,  for each tag $t$ and for each scenario we collected around 2000 communications. Each communication, $C_t$, is composed by 3400 I/Q sampled at 5 MS/s.
We defined an array of perturbation coefficients $\Phi=\{0.20,0.10,0.05,0.01\}$. To transform the signal in a machine learning readable input, we take a communication $C$, composed by  $s$ complex numbers, and decompose the real and the imaginary components in a matrix of shape ($s$,2). Depending on the $L$ window size, we slice these matrices in contiguous sub-matrices of length $L$. Each of these matrices represents a tensor that will be feed to the machine learning model. These sub-matrices compose $X$, which is the set containing the input of the machine learning model. Thus, considering $X$ as the model input:

\begin{equation}
    X_{aug} = \bigcup_{\phi \in \Phi} \{A(x,\phi)_{\forall{x \in X} }  \}
\end{equation}

\begin{equation}
A(x,\phi) =
    \begin{cases}
      x^r = x^r+\mathcal{N}(\mu = 0,\sigma = \phi*avg(x^r)) \\
      x^i = x^i+\mathcal{N}(\mu = 0,\sigma = \phi*avg(x^i))
    \end{cases}
\end{equation}

\noindent with $x^r$ and $x^i$ representing the real and imaginary parts of the sampled inputs, relatively the first and the second columns of the matrices in $X$, and $norm$ the normal distribution drawn with $\mu = 0$ and $\sigma = \phi*avg(x) $, with $avg(x)$ as the mean of $x$. The final input for the machine learning model is composed by the union of the sampled signal and the augmented one, $X = X \cup X_{aug}$.

\section{Data Collection Campaign}\label{sec:testbed}

We first introduce some background information on the EPC-Gen2 standard in Section \ref{sec:epc-gen2}, then we introduce our experimental setup  and dataset structure in Section \ref{sec:dataset}.

\subsection{Background on EPC-Gen2}\label{sec:epc-gen2}

On the software side, almost all the UHF RFID tags communicate through the EPC-Gen2 standard \cite{global2008epc}. In short, the reader delivers power to the tags by emitting a Continuous Wave (CW) and begins all reader-to-tags signaling with either a preamble or a frame synchronization. The preamble precedes a Query command and denotes the start of an inventory round, specifying the number of slots that will be issued in the frame. Each slot is signaled by the reader with a frame-sync message. Each tag randomly selects one slot and replies by sending a 16-bit random or pseudo-random number (RN16) when such slot occurs. If no tags reply, the slot is idle, the reader truncates it and signals the next slot. If one or more tags reply, the reader sends back the received RN16 that can be single (if only one tag transmitted) or colliding (if multiple tags transmitted concurrently). In case of single transmission, the transmitting tag recognizes its RN16 and sends back its unique EPC ID. The slot then terminates with the reader acknowledging the received ID. If instead multiple tags replied together, the reader sends back a RN16 that is the sum of the different tags sequences. As no tag recognizes its RN16, the slot terminates without the transmission of any EPC ID. Each message includes a cyclic redundancy checks (CRC) code to enable error detection. Specifically, the RN16 includes a 5-bit CRC, while the EPC ID includes a 16-bit CRC.

\begin{figure}[!h]
    \centering
  \includegraphics[width=.9\columnwidth]{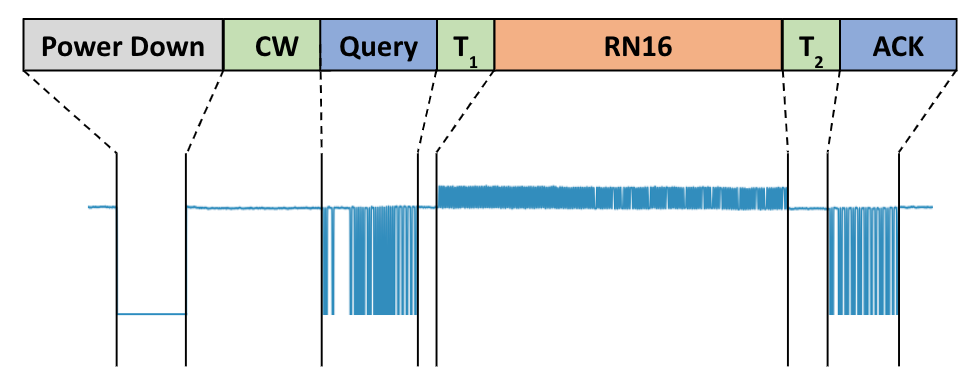}
  \caption{Reader-tag communication during the sampling process.\vspace{-0.2cm} }\label{fig:many_s} 
\end{figure}

\textit{In this work, we exploit the randomness of RN16 to sample each tag over multiple different transmitted messages. } Figure \ref{fig:many_s} shows the structure of a communication between our reader and a EPC-Gen2 tag during the sampling process. The interaction begins with the reader emitting a CW to energize the tag and allow it to receive and act on reader commands. After receiving this message, the tag takes time $T_1$ to react and reply by sending a RN16. The reader after receiving the response needs a reaction time $T_2$ before being able to acknowledge the tag by issuing an ACK. The interaction repeats multiple times --- depending on the number of samplings per tags --- with the reader energizing again the tag and issuing a new query. Reaction time depends on device characteristics and measures the reaction time from the end of a message reception to the start of a message transmission. On the tag side, it is estimated as $T_1= 10/R$, while on the reader side it is estimated as $T_2=1/R$, where $R$ is the datarate (according to EPC specifications, see \cite{global2008epc}, p. 43)). 
The sampling process relies only on RN16 collection. This is not without a reason. Sampling a tag over a sequence that changes at each transmission allows the model to effectively learn hardware specific features. Instead, sampling a tag that transmitted always the same sequence (i.e., EPC ID) would make the model learn features related to a software defined ID.

\subsection{The \datasetname\ Dataset} \label{sec:dataset}

\begin{figure}[h!]
    \centering
  \includegraphics[width=0.9\columnwidth]{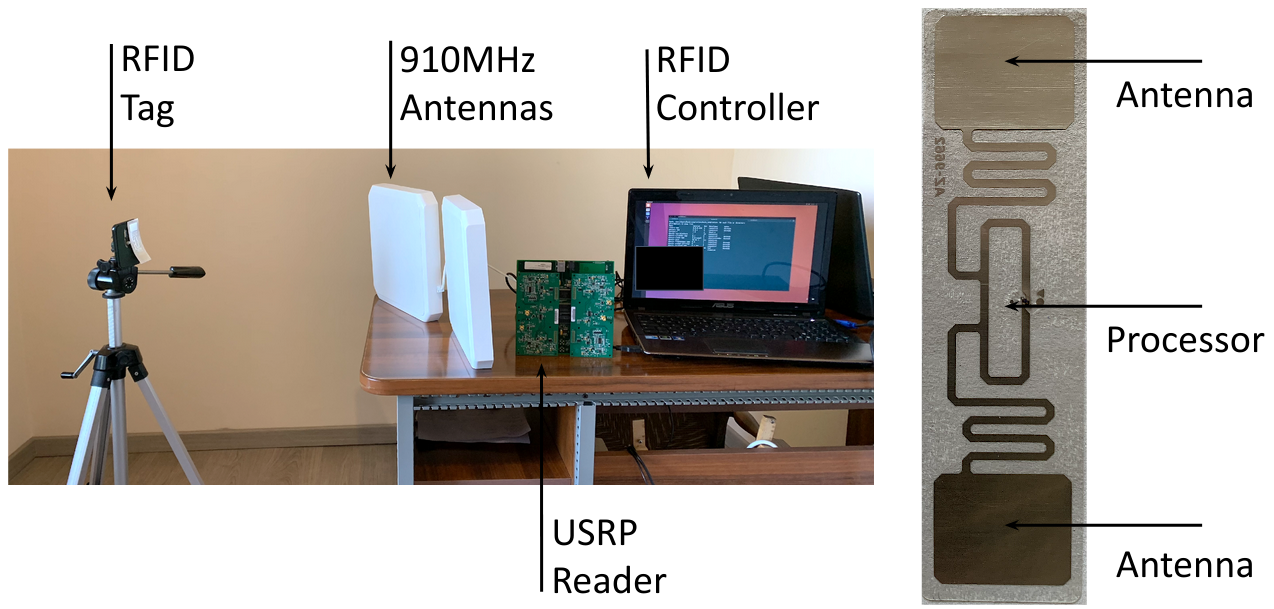}
  \caption{(left) Experimental testbed; (right) RFID tag.\vspace{-0.1cm} }\label{fig:testbed} 
    
\end{figure}

We performed a massive waveform collection campaign with an experimental setup composed by an Ettus USRP 4 and two Flex 900 daughterboards, as depicted in figure \ref{fig:testbed}. We used 200 AZ 9662 Alien H3 73.5x21.2mm UHF tags \cite{YaronTechRFID}, depicted in figure \ref{fig:testbed}. These tags are produced by YaronTech, widely adopted and easily purchasable over the web. Tags have been interrogated with the EPC-Gen2 standard \cite{global2008epc}, through a USRP implementation \cite{buettner2010gen}. The antennas used in our setup are 902-928 MHZ RFID antennas produced by Laird Connectivity, model S9028PCL \cite{lairdantennas}.

\begin{figure}[h!]
    \centering
  \includegraphics[width=0.8\columnwidth]{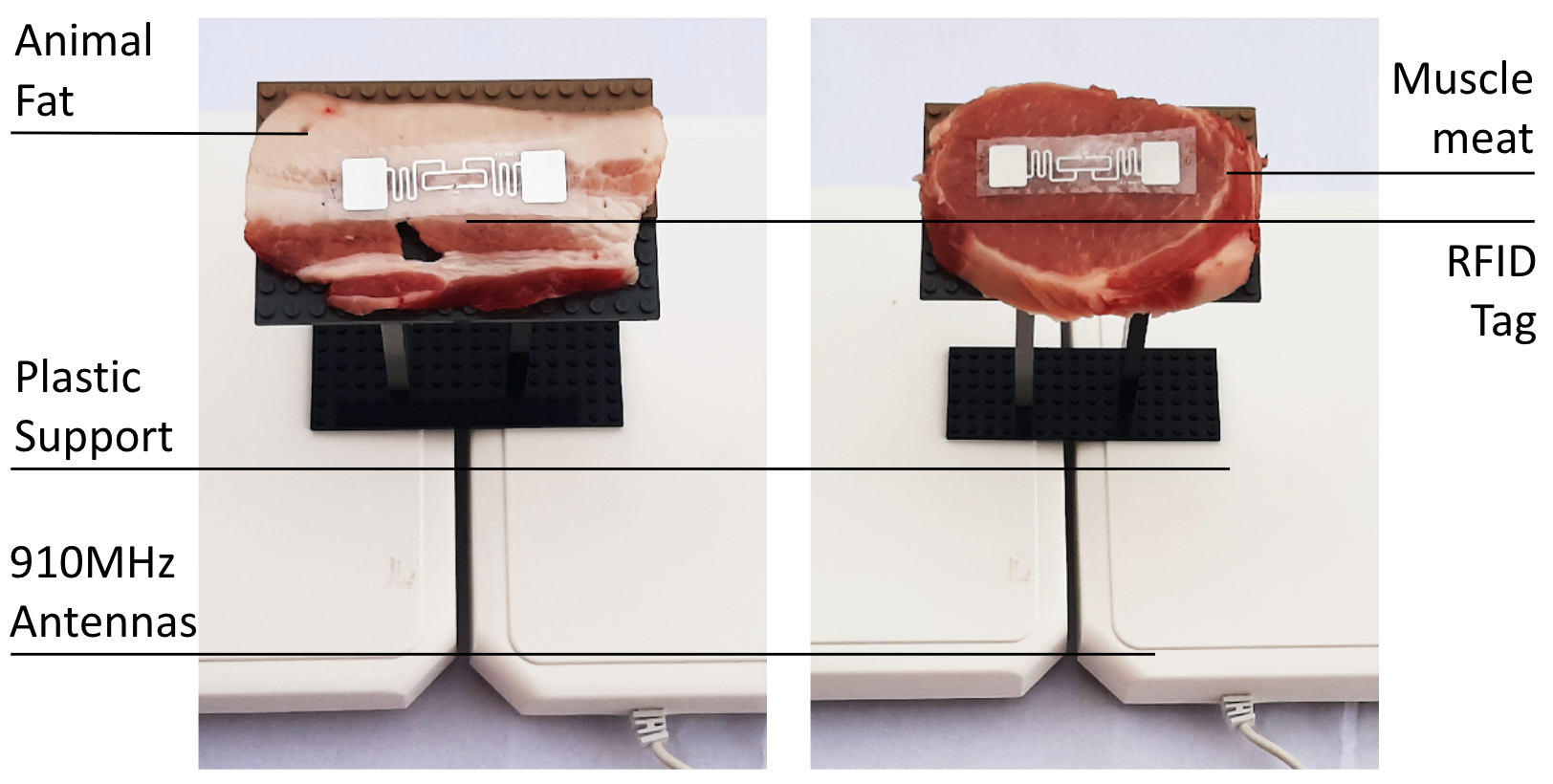}
  \caption{Data collection with animal fat(left), and animal muscle meat(right). }\label{fig:tower_bacon} 
\end{figure}

The RFID16-2021 dataset contains RN16 exchange signals collected in 7 different scenarios. Each scenario, encoded as {\em SCEN-dist-obst}, is defined by 2 features: (i) the tag-reader distance (encoded by {\em dist}) and (ii) the eventual obstacle present between a tag and the reader (encoded by {\em obst}). We performed data collection at three distances, 20cm, 50cm and 100cm, relatively encoded as {\em 020, 050, 100}. We defined three different scenario: {\em Over-The-Air (OTA)}, which represents a data collection without obstacles between the reader and a tag, {\em Porcine Meat 0(PM0)} which represents a data collection with a fat porcine meat 0.5cm thick obstacle between the reader and the tag, as depicted in figure \ref{fig:tower_bacon}, and {\em PM1} in which we use a low-fat meat of 3cm thick as obstacle. As an example, {SCEN-050-PM1} represents a data collection at a distance of 50 centimeters between the reader and the tag an a low fat porcine meat as obstacle. The PM0 and PM1 subdatasets have 20 tags only, while OTA subdatasets have up to 200 tags each.

We used porcine meat as propagation media because it has similar properties to human muscular tissues, so as to simulate the presence of an under skin sensor (PM0) and a more deeply implanted one (PM1), as the PM1 porcine meat of 3 cm of thick simulates the presence of a muscle. The tags are the same throughout the whole data collection, meaning that the tag with label 0 is the same hardware in all the subsets. For our results we split every subset in the ratio $0.8, 0.1, 0.1$ for train, validation, and test. The data collection process has been performed over a 2-week periods for the subsets $OTA20, OTA50, OTA100$, where the first 100 tags have been collected through the first week and the other 100 during the successive one. The data collection process have been done in a normal office, and data are thus affected by uncontrolled and real interference. The \datasetname\ dataset is publicly available\cite{dataset-website}.

\section{Experimental Results}\label{sec:results}

In this section, we present extensive experimental results obtained from the {\datasetname} dataset. To the best of our knowledge, this is the first publicly available dataset for \gls{rfp} with RFID samples. This makes these experimental results completely reproducible, and a benchmark which today is not available. We recall that the {\datasetname} is composed by 7 subdatasets, $OTA20$, $OTA50$, $OTA100$, $PM0-20$, $PM0-50$, $PM1-20$, $PM1-50$, as described in Section \ref{sec:dataset}, and that the tags are the same through the whole data collection, meaning that the tag with label 0 is the same hardware in all the subsets. We split each subdataset following the $0.8/0.1/0.1$ ratio for training, validation and testing, respectively. 
In the following, we will refer to the term "accuracy" to indicate the accuracy computed on the test set.

\subsection{Hyperparameter Evaluation}

\begin{figure}[h!]
         \begin{subfigure}[b]{0.48\columnwidth}
                 \centering
                 \includegraphics[width=\textwidth]{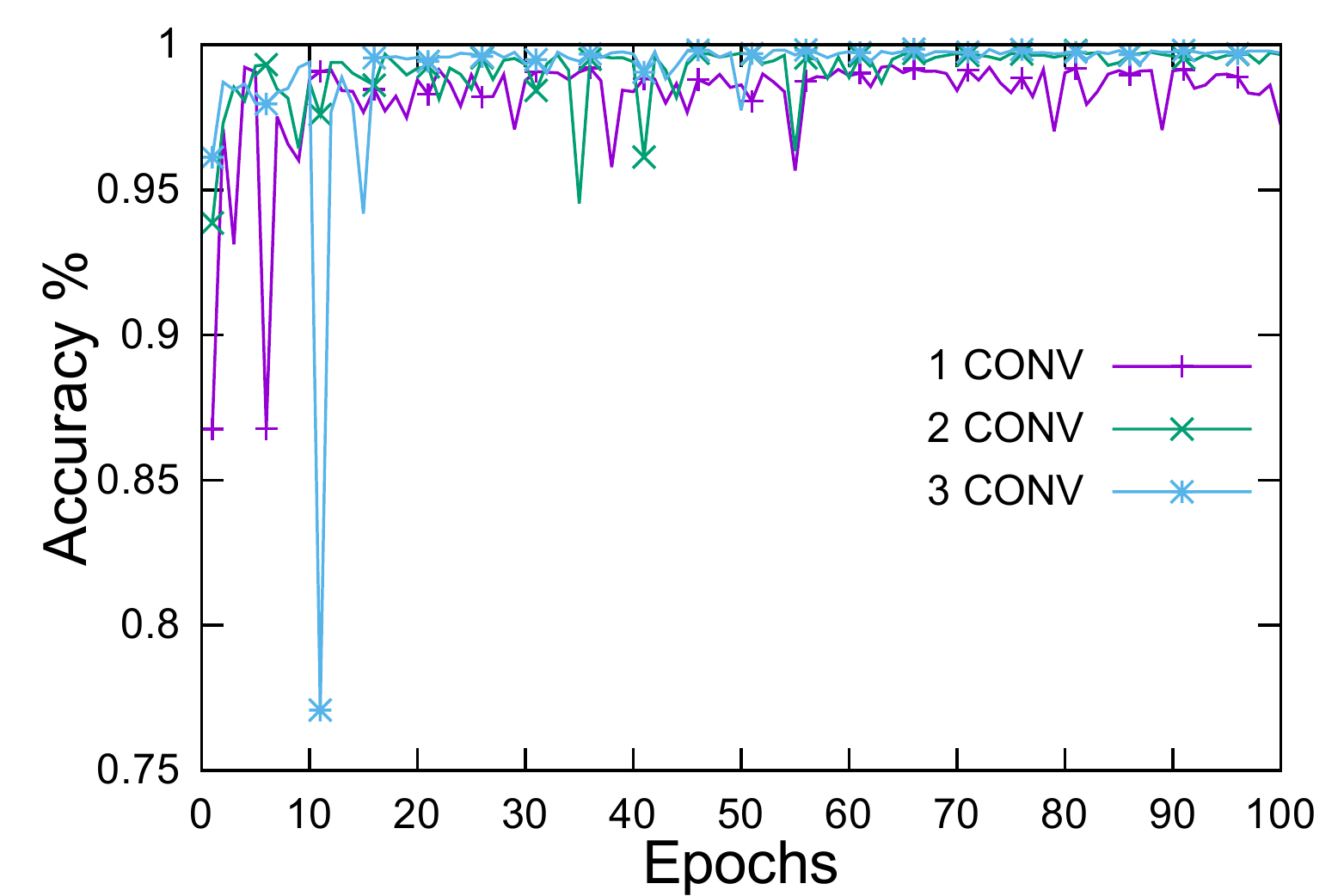}
                 \caption{20 Tags over OTA20}
                 \label{fig:res_varioconvoluzioni_a}
         \end{subfigure}
              \begin{subfigure}[b]{0.48\columnwidth}
             \centering
             \includegraphics[width=\textwidth]{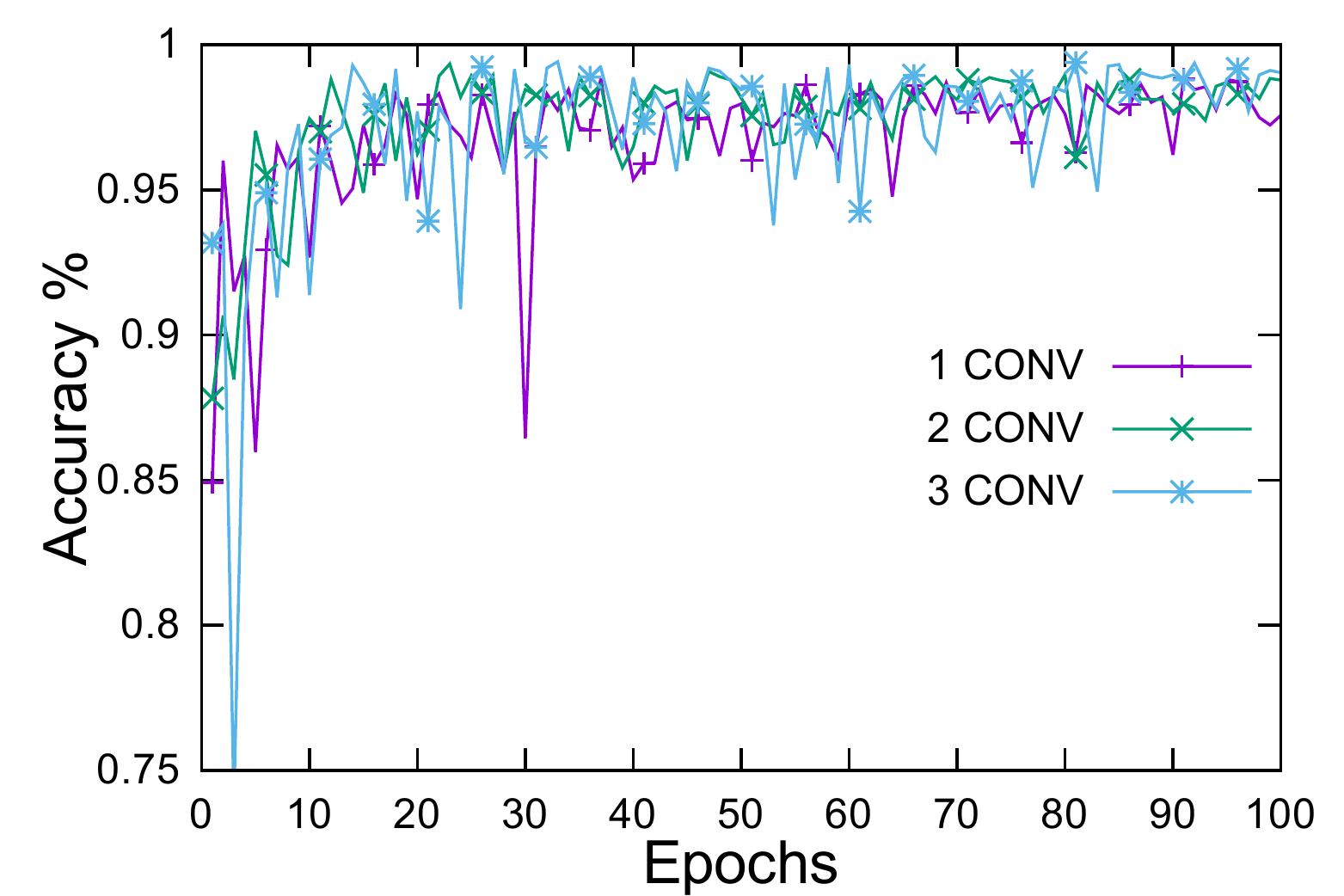}
             \caption{PM1-20}
             \label{fig:res_varioconvoluzioni_c}
     \end{subfigure}

         \begin{subfigure}[b]{0.48\columnwidth}
                 \centering
                 \includegraphics[width=\textwidth]{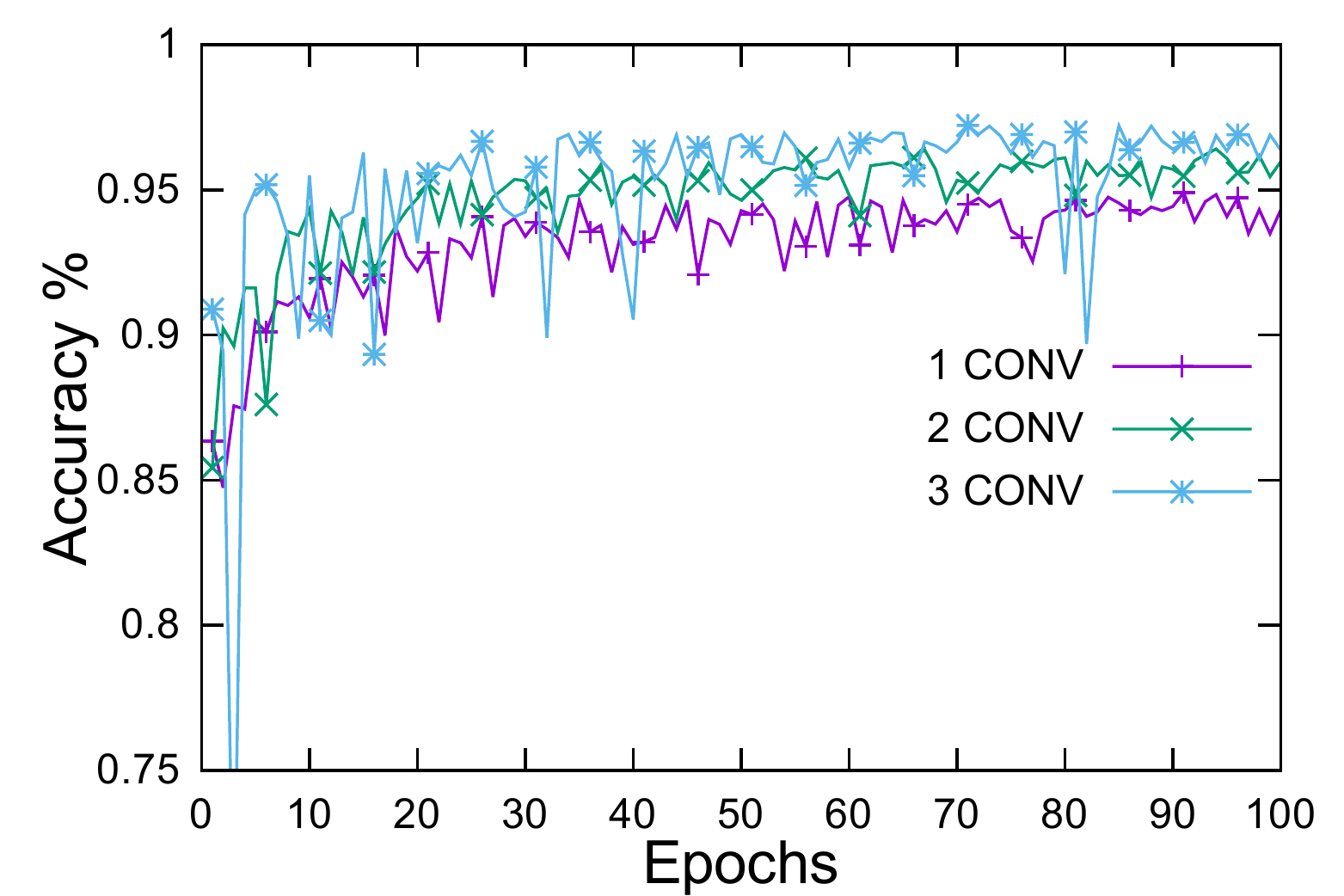}
                 \caption{200 Tags over OTA20}
                 \label{fig:res_varioconvoluzioni_b}
         \end{subfigure}
     \begin{subfigure}[b]{0.48\columnwidth}
             \centering
             \includegraphics[width=\textwidth]{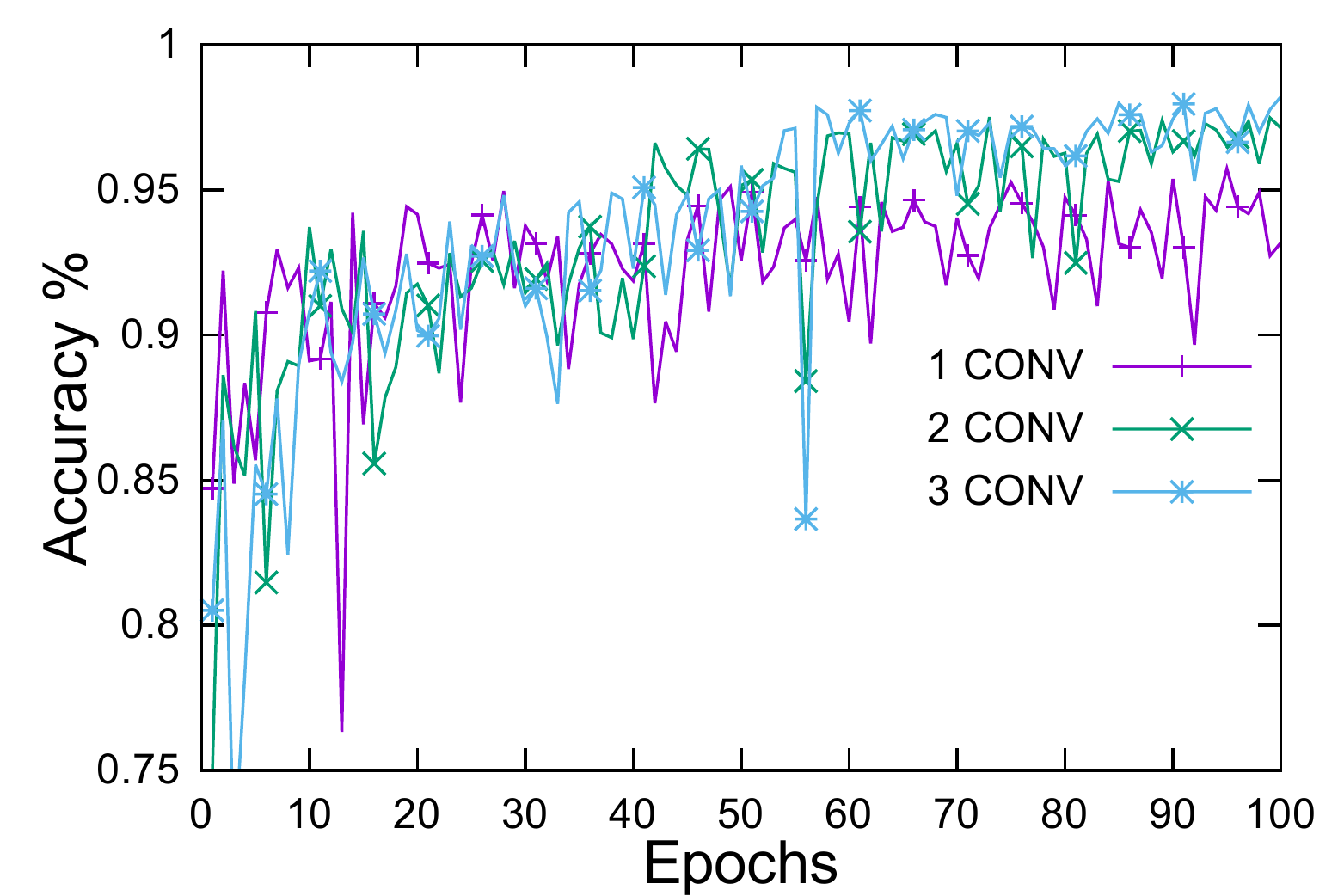}
             \caption{PM1-50}
             \label{fig:res_varioconvoluzioni_d}
      \end{subfigure}
\caption{Results varying the number of convolutions, on OTA20 and PM1 } \label{fig:res_varioconvoluzioni}
\end{figure}

We first investigate the impact of the number of CONV layers in the \gls{cnn} described in Section \ref{sec:nn_descr} on the accuracy. Figure \ref{fig:res_varioconvoluzioni} shows the accuracy by changing the number of convolutional boxes with different scenarios. We made a test choosing the first twenty tags form OTA20. The model with 1 convolution has a lower accuracy with respect to the other two, but with an accuracy drop of less then 2\%, making the results of the three models comparable. We tried again with twenty tags but with porcine meat as obstacle, using the set PM1 at 20 centimeters of distance between the tag and the reader, and still the results of the three model are comparable, even if with a general accuracy drop of few points with respect to the first test we presented. 

With 200 tags and OTA20, we notice a general drop of few points which is interestingly more accentuated in the model with 1 convolution. Models with 2 and 3 convolutions are instead comparable. The most complex case for the neural network is instead represented by the scenario with porcine meat at 50 cm, PM1-50, where the accuracy of the 1 convolution model keep dropping. The accuracy of the models with 2 and 3 CONV layers is instead stable an comparable, and we thus decided to utilize the 2-CONV model.

\subsection{Impact of the Wireless Channel}\label{sec:learning_channel}

\begin{figure}[h!]
         \begin{subfigure}[b]{0.48\columnwidth}
                \centering
                \includegraphics[width=\textwidth]{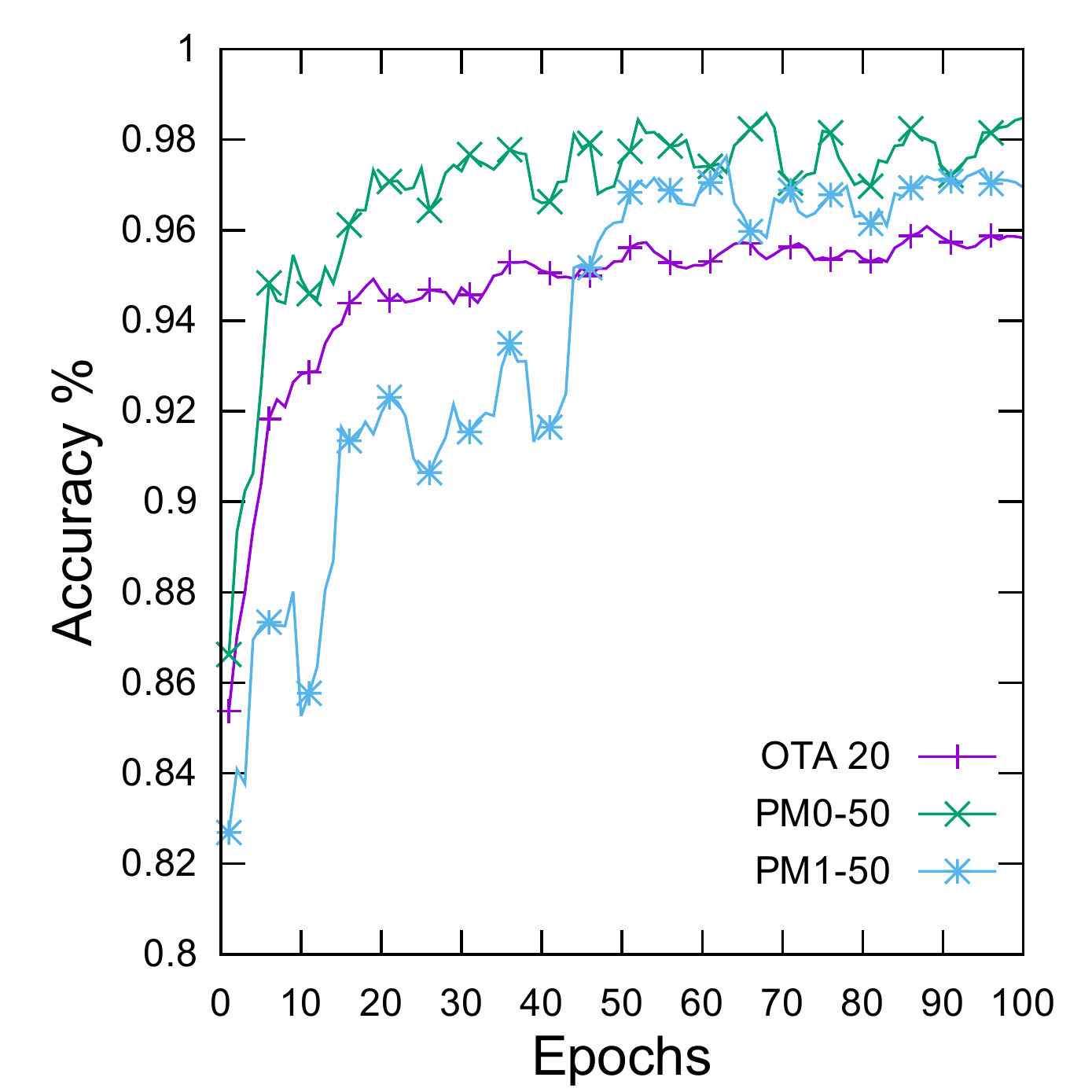}
                \caption{Test accuracy on three subsets of \datasetname. }\label{fig:traintestsameset} 
         \end{subfigure}
         \begin{subfigure}[b]{0.48\columnwidth}
                 \centering
    \captionsetup{justification=centering}

    \setlength\fwidth{0.65\columnwidth}
    \setlength\fheight{0.65\columnwidth}
    \begin{tikzpicture}
\pgfplotsset{every tick label/.append style={font=\tiny}}

\begin{axis}[
enlargelimits=false,
colorbar,
colormap/Purples,
width=\fwidth,
height=\fheight,
at={(0\fwidth,0\fheight)},
scale only axis,
tick align=inside,
xlabel={Train Set},
xmin=-0.5,
xmax=6.5,
xtick style={draw=none},
xtick = {0,1,2,3,4,5,6},
xticklabels = {OTA20,OTA50,OTA100,PM0-20,PM1-20,PM0-50,PM1-50},
xlabel style={font=\scriptsize\color{white!15!black}},
ylabel style={font=\scriptsize\color{white!15!black}},
ylabel={Test Set},
ymin=-0.5,
ymax=6.5,
xlabel shift=-5pt,
ylabel shift=-5pt,
ytick style={draw=none},
ytick = {0,1,2,3,4,5,6},
yticklabels = {OTA20,OTA50,OTA100,PM0-20,PM1-20,PM0-50,PM1-50},
axis background/.style={fill=white},
xticklabel style = {rotate=45,anchor=east},
colorbar horizontal,
colorbar style={
at={(0,1.05)},               % <-- (changed)
anchor=below south west,    % <-- (changed)
% change the width of the colorbar relative to the main `axis' environment
width=\pgfkeysvalueof{/pgfplots/parent axis width},
xtick={0, 0.5, 1},
xmin=0,
xmax=1,
axis x line*=top,
xticklabel shift=-1pt,
point meta min=0,
point meta max=1,
},
colorbar/width=2mm,
]\addplot [matrix plot,point meta=explicit]
 coordinates {
(0,0) [0.805191134977836] (0,1) [0.05509973666165741] (0,2) [0.001166707416009723] (0,3) [0.05944215065257555] (0,4) [0.0] (0,5) [0.046439048231321285] (0,6) [0.03266122206059998] 

(1,0) [0.07192173483121793] (1,1) [0.7747331837365004] (1,2) [0.077713667632193] (1,3) [0.01605528276413074] (1,4) [0.017960517811038208] (1,5) [0.0027535277810029923] (1,6) [0.0388620854439166] 

(2,0) [0.07898955644865033] (2,1) [0.05448606935396333] (2,2) [0.7079213008695214] (2,3) [0.028735493415597183] (2,4) [0.01969474302410467] (2,5) [0.03464129634785029] (2,6) [0.0755315405403129] 

(3,0) [0.035255754032421226] (3,1) [0.07174495319118653] (3,2) [0.0026118006220879385] (3,3) [0.7411869631126164] (3,4) [0.03949488053662497] (3,5) [0.0659406945478594] (3,6) [0.04376495395720359] 

(4,0) [0.011785504270030419] (4,1) [0.039026448645579796] (4,2) [0.04754141349893379] (4,3) [0.10628879021827067] (4,4) [0.7679421494316813] (4,5) [0.0006642436841433443] (4,6) [0.026751450251360675] 

(5,0) [0.05706311638600783] (5,1) [0.07923360075362468] (5,2) [0.0026617942768758928] (5,3) [0.07147696950156951] (5,4) [0.0053171558619136] (5,5) [0.7824377079983589] (5,6) [0.0018096552216496067] 

(6,0) [0.046605965523989976] (6,1) [0.0] (6,2) [0.06870261273158942] (6,3) [0.024828410957547276] (6,4) [0.05020402493255895] (6,5) [0.0001467172237318132] (6,6) [0.8095122686305826]

};
\node[text=white] at (axis cs: 0,0) {\scriptsize 96\%};
\node[text=black] at (axis cs: 0,1) {\scriptsize 5\%};
\node[text=black] at (axis cs: 0,2) {\scriptsize 1\%};
\node[text=black] at (axis cs: 0,3) {\scriptsize 6\%};
\node[text=black] at (axis cs: 0,4) {\scriptsize 0\%};
\node[text=black] at (axis cs: 0,5) {\scriptsize 4\%};
\node[text=black] at (axis cs: 0,6) {\scriptsize 3\%};
\node[text=white] at (axis cs: 1,1) {\scriptsize 97\%};
\node[text=white] at (axis cs: 2,2) {\scriptsize 94\%};
\node[text=white] at (axis cs: 3,3) {\scriptsize 95\%};
\node[text=white] at (axis cs: 4,4) {\scriptsize 98\%};
\node[text=white] at (axis cs: 5,5) {\scriptsize 99\%};
\node[text=white] at (axis cs: 6,6) {\scriptsize 98\%};

\node[text=black] at (axis cs: 1,0) {\scriptsize 7\%};
\node[text=black] at (axis cs: 1,2) {\scriptsize 8\%};
\node[text=black] at (axis cs: 1,3) {\scriptsize 2\%};
\node[text=black] at (axis cs: 1,4) {\scriptsize 2\%};
\node[text=black] at (axis cs: 1,5) {\scriptsize 0\%};
\node[text=black] at (axis cs: 1,6) {\scriptsize 3\%};

\node[text=black] at (axis cs: 2,0) {\scriptsize 7\%};
\node[text=black] at (axis cs: 2,1) {\scriptsize 5\%};
\node[text=black] at (axis cs: 2,3) {\scriptsize 3\%};
\node[text=black] at (axis cs: 2,4) {\scriptsize 2\%};
\node[text=black] at (axis cs: 2,5) {\scriptsize 3\%};
\node[text=black] at (axis cs: 2,6) {\scriptsize 7\%};

\node[text=black] at (axis cs: 3,0) {\scriptsize 3\%};
\node[text=black] at (axis cs: 3,1) {\scriptsize 4\%};
\node[text=black] at (axis cs: 3,2) {\scriptsize 5\%};
\node[text=black] at (axis cs: 3,4) {\scriptsize 10\%};
\node[text=black] at (axis cs: 3,5) {\scriptsize 0\%};
\node[text=black] at (axis cs: 3,6) {\scriptsize 2\%};

\node[text=black] at (axis cs: 4,0) {\scriptsize 1\%};
\node[text=black] at (axis cs: 4,1) {\scriptsize 4\%};
\node[text=black] at (axis cs: 4,2) {\scriptsize 5\%};
\node[text=black] at (axis cs: 4,3) {\scriptsize 10\%};
\node[text=black] at (axis cs: 4,5) {\scriptsize 0\%};
\node[text=black] at (axis cs: 4,6) {\scriptsize 3\%};

\node[text=black] at (axis cs: 5,0) {\scriptsize 6\%};
\node[text=black] at (axis cs: 5,1) {\scriptsize 8\%};
\node[text=black] at (axis cs: 5,2) {\scriptsize 0\%};
\node[text=black] at (axis cs: 5,3) {\scriptsize 7\%};
\node[text=black] at (axis cs: 5,4) {\scriptsize 0\%};
\node[text=black] at (axis cs: 5,6) {\scriptsize 0\%};

\node[text=black] at (axis cs: 6,0) {\scriptsize 5\%};
\node[text=black] at (axis cs: 6,1) {\scriptsize 0\%};
\node[text=black] at (axis cs: 6,2) {\scriptsize 7\%};
\node[text=black] at (axis cs: 6,3) {\scriptsize 2\%};
\node[text=black] at (axis cs: 6,4) {\scriptsize 5\%};
\node[text=black] at (axis cs: 6,5) {\scriptsize 0\%};

\end{axis}
\end{tikzpicture}

%%0,96	0,97	0,94	0,95	0,98	0,99	0,98
    \caption{Cross accuracy testing between differen subsets.}
        \label{fig:cm_traintest}
         \end{subfigure}

\caption{Impact of the wireless channel.\vspace{-0.3cm}} \label{fig:train_rest_same}
\end{figure}
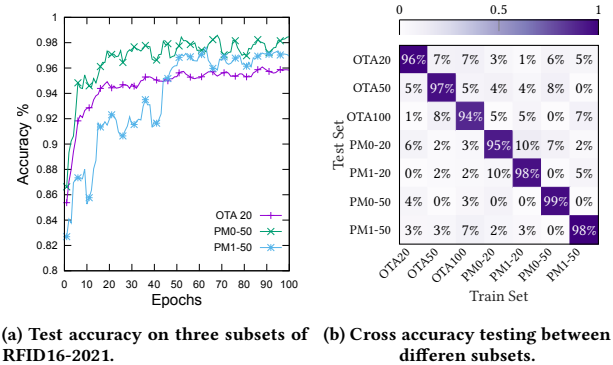

Our next investigation is aimed at understanding the impact of the wireless channel on the classifier's performance. Therefore,  we trained the neural network described in Section \ref{sec:nn_descr} on all the 7 subdatasets, and tested it only on the test part of the same subset $S$, e.g. we trained the model on the train part of $OTA20$ and tested it on the test part of $OTA20$. As shown in Figure \ref{fig:traintestsameset}, the neural network reaches an accuracy of up to 96\% in free air communication with 200 tags and 98\% in porcine meat with 20 tags.
Interestingly, the neural network does not seem to be affected by the presence of obstacles like the porcine meat.
The test on PM1-50 reports an accuracy of $98\%$, higher than the one achieved performing the same test with the tags and reader closer and without the porcine meat obstacle.
Increasing the number of tags (200) the accuracy lowers ($95\%$) with respect to the other scenarios.

These results confirm findings in prior literature, where the radio fingerprinting is performed without specifying the channel condition. However, a simple cross-test shown in Figure  \ref{fig:cm_traintest} enlightens the limits of previous approaches, as a model trained on a sub-dataset $X$ is not able to generalize and recognize the same tag with a model trained on another sub-dataset. Indeed, Figure \ref{fig:cm_traintest} shows that whenever training and testing are performed over different sets - which is the case of every element that is not on the diagonal - the accuracy is very low (color intensity in the confusion matrix shows accuracy), down to 0\% in some cases. 
\textbf{These results conclude that an RFID tag can be classified correctly if both the train and test data have been collected with the same channel conditions. When instead a model is trained on a channel condition and tested, with the same tag, in another channel condition, we have a complete disruption of the results.}

\subsection{Federated \gls{rfp} Evaluation}\label{sec:federated_results}

The key aspect of \gls{fml} is that a set of clients, with different data, can work together to train a shared model, without sharing the data directly. We made experiments without and with obstacles. In the first case we defined three clients, each one representing a different reader, which communicate with the same tag but at different distances.  We created this scenario assigning the subset OTA20 to the first client, OTA50 to the second and OTA100 to the third. Each client trains a local model and, every epoch, sends the model weights to the central server, as shown in figure \ref{fig:nn_federated}. This central server produces a federated average of the weights it receives, and shares back the federated weights with the clients. For the federated learning we adopted the same neural network described in the precedent section \ref{sec:nn_descr}.

To evaluate the accuracy of the federated model \textit{we defined both an optimum scenario and a lower bound}. In the optimum scenario, which we call Union, we have a unique client which can access all the three dataset, making training and testing on the union of the datasets.
For the lower bound, which we call Baseline, we have $n$ clients, each one with a model $m_n$ and a dataset $d_n$. The Baseline is equal to the average of the accuracy of all the $m_n$ models tested on all the $d_n$ datasets.

%%%FEDERATO CON 10 PERC

\begin{figure}[h]
         \begin{subfigure}[b]{0.48\columnwidth}
                 \centering
                 \includegraphics[width=\textwidth]{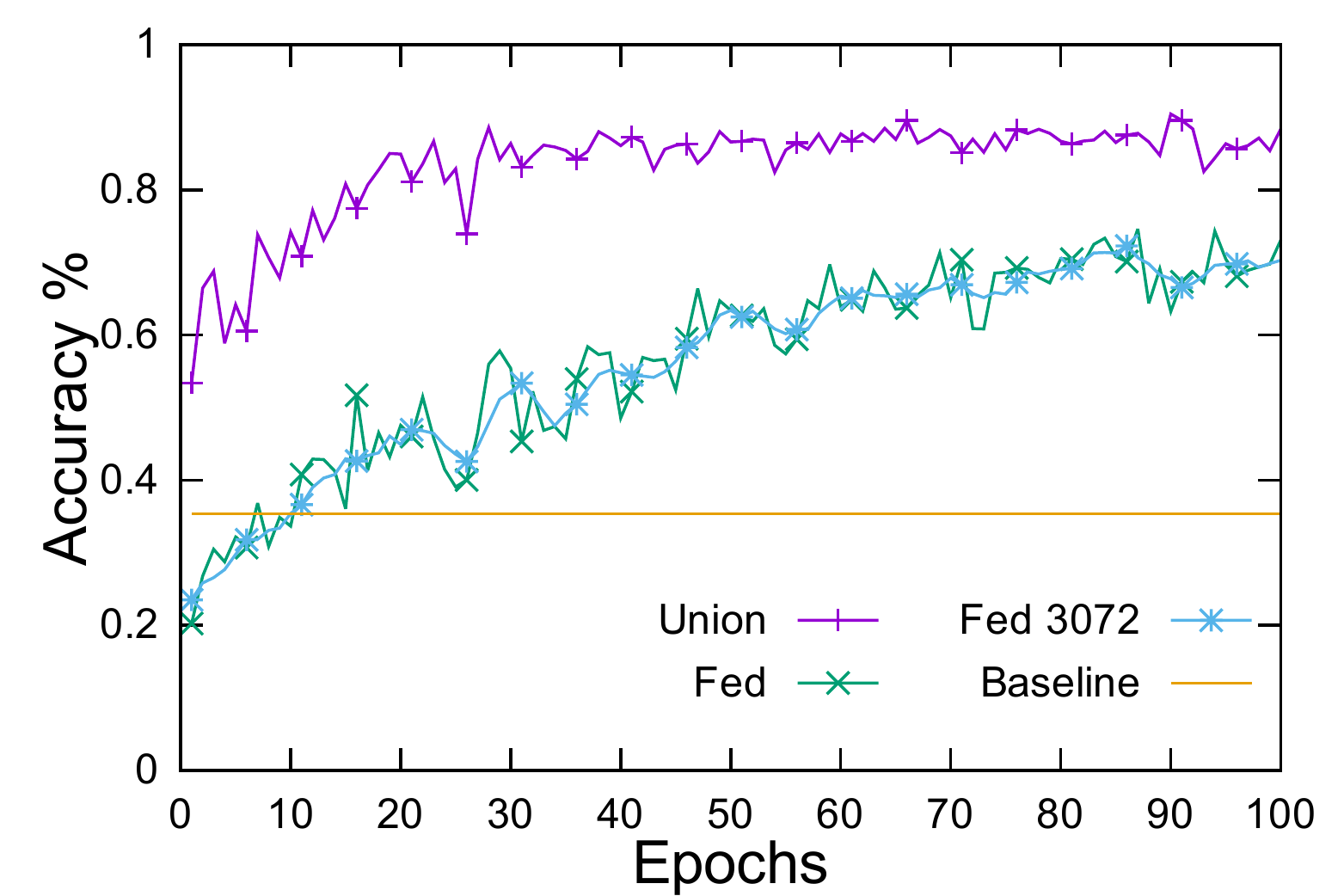}
                 \caption{20 Tags}
                 %\label{}
         \end{subfigure}
         \begin{subfigure}[b]{0.48\columnwidth}
                 \centering
                 \includegraphics[width=\textwidth]{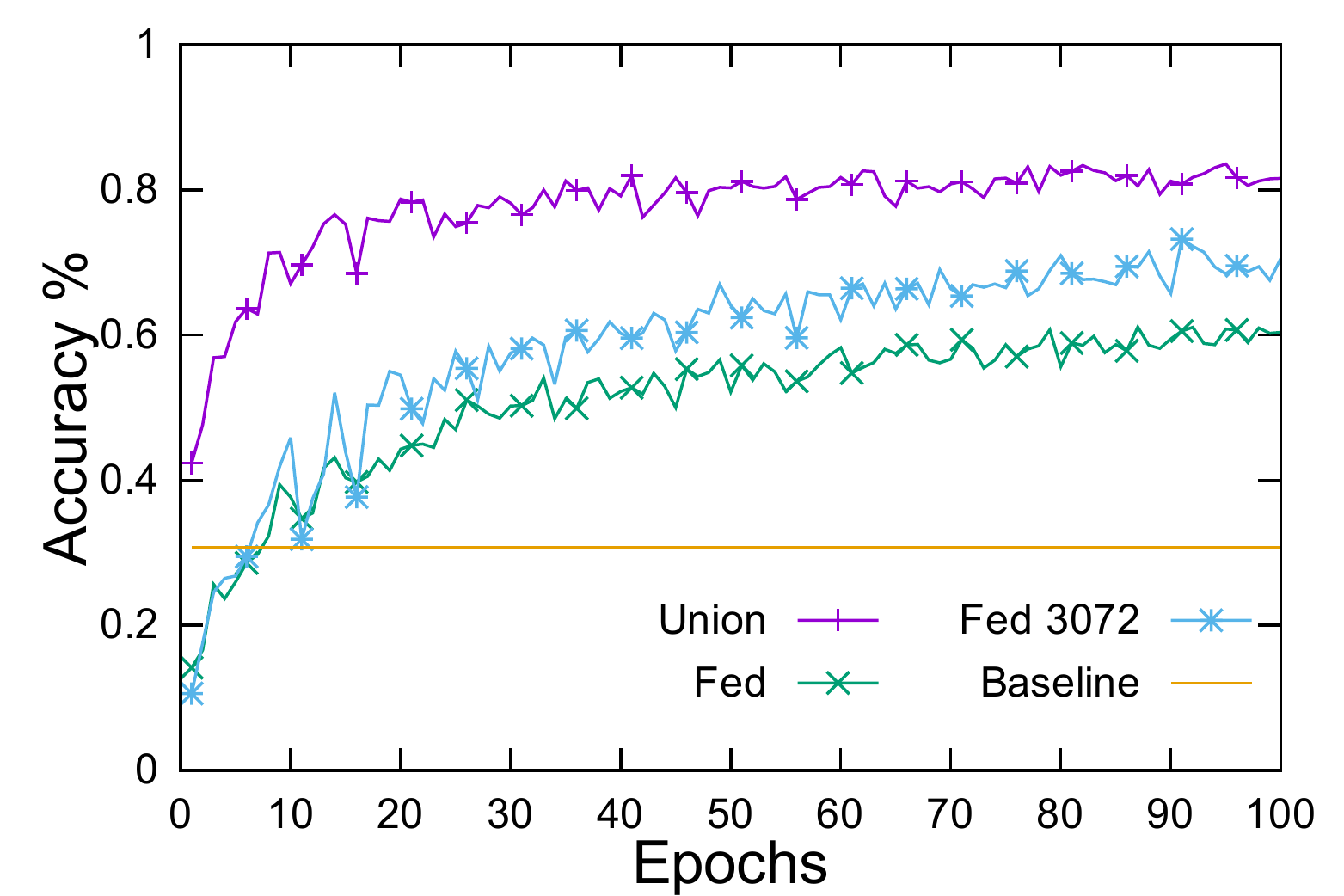}
                 \caption{50 Tags}
                 %\label{}
         \end{subfigure}
% leave a blank line to change row         

     \begin{subfigure}[b]{0.48\columnwidth}
             \centering
             \includegraphics[width=\textwidth]{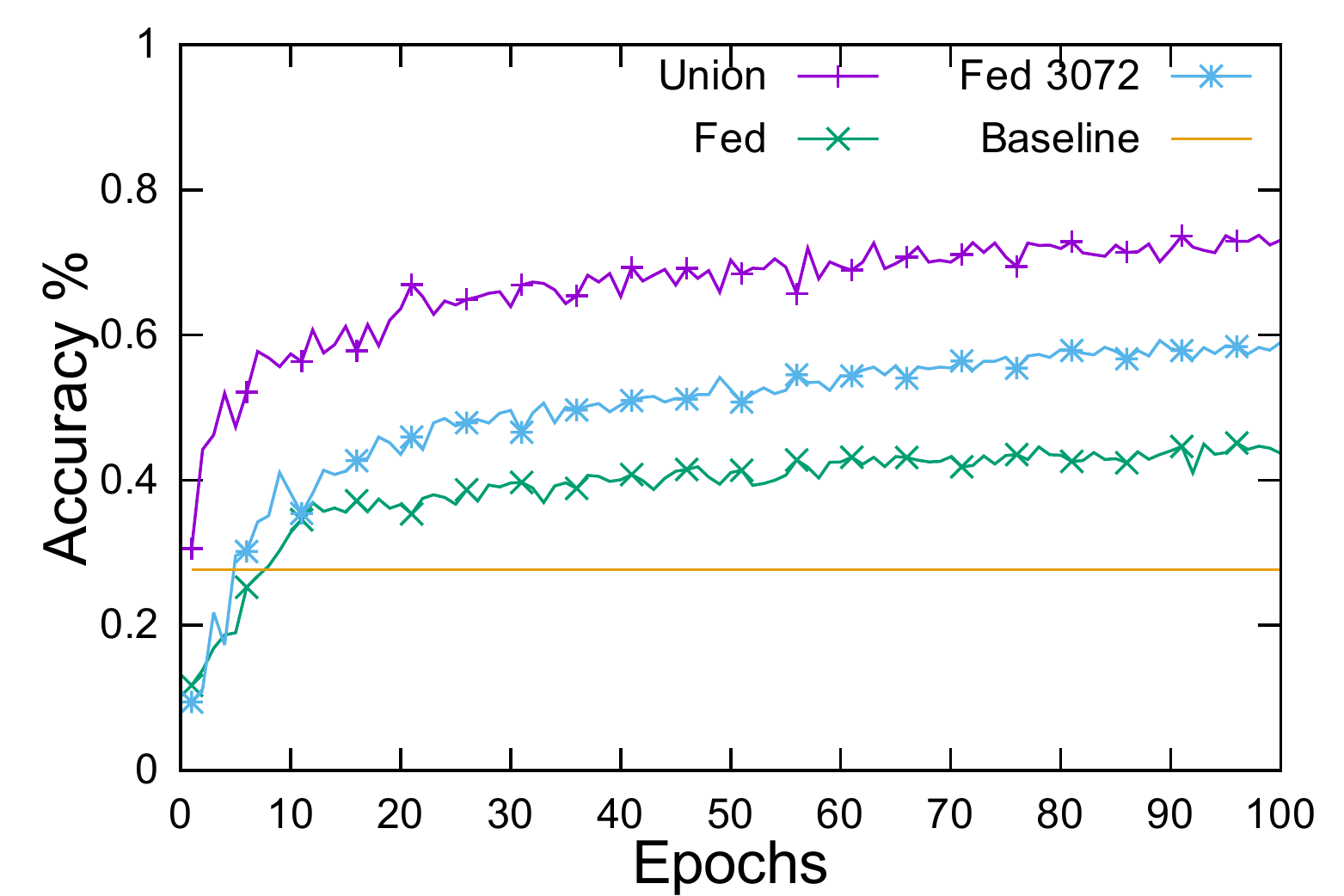}
             \caption{100 Tags}
             %\label{}
     \end{subfigure}
     \begin{subfigure}[b]{0.48\columnwidth}
             \centering
             \includegraphics[width=\textwidth]{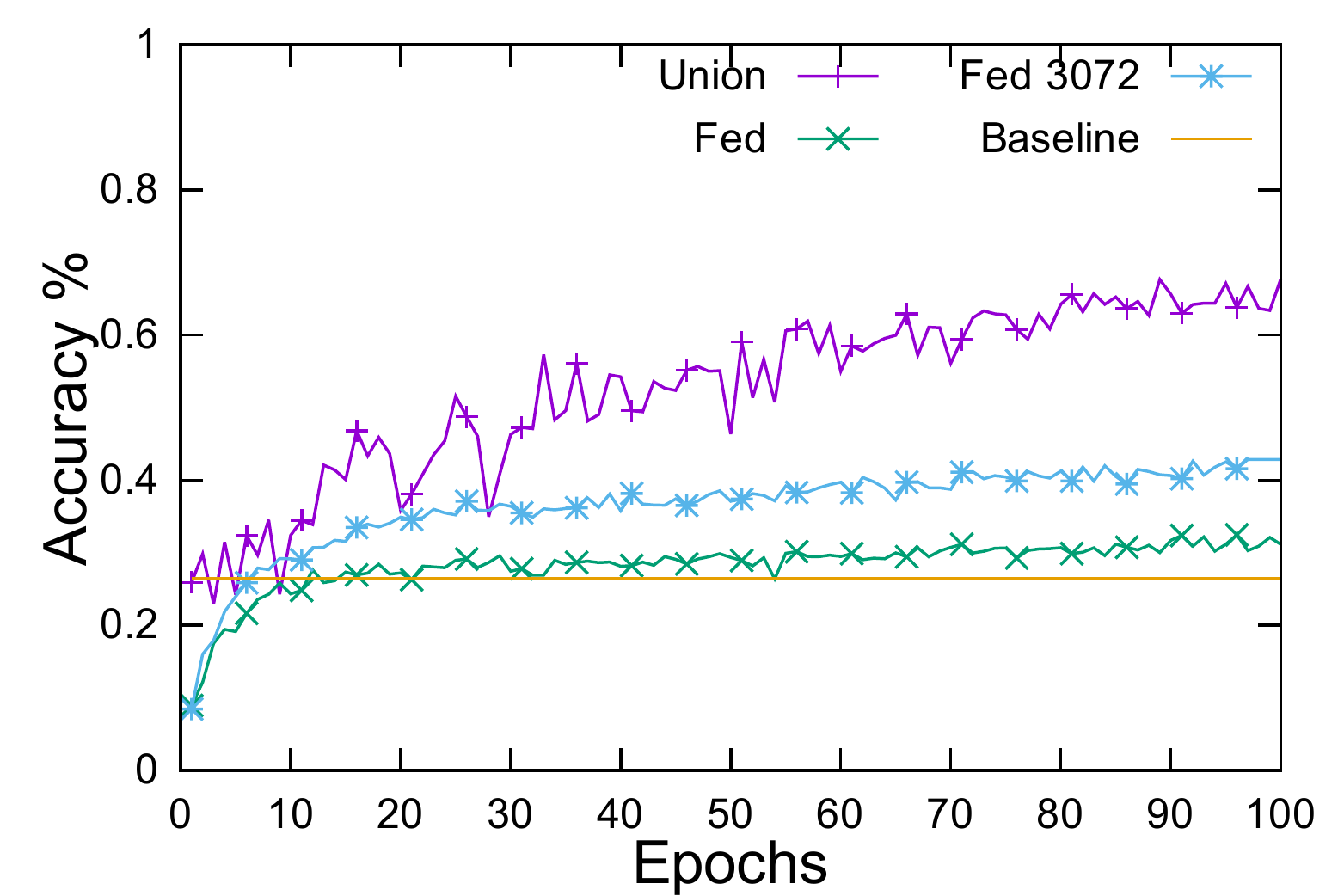}
             \caption{200 Tags}
             %\label{}
      \end{subfigure}
\caption{Accuracy of Federated RFP with 10 percent of data.\vspace{-0.4cm}} \label{fig:res_federated10percent}
\end{figure}

\begin{figure}[h]
         \begin{subfigure}[b]{0.48\columnwidth}
                 \centering
                 \includegraphics[width=\textwidth]{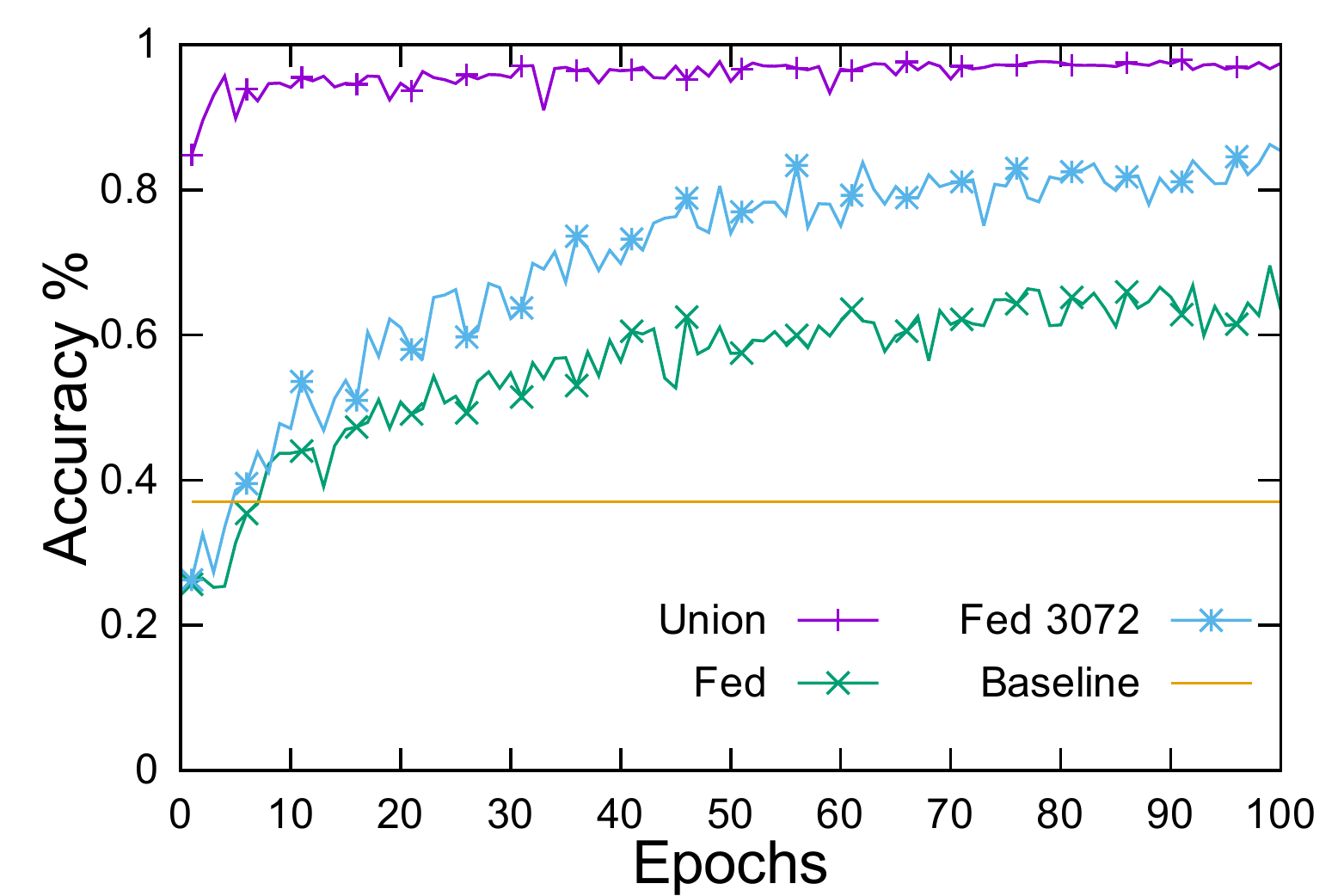}
                 \caption{20 Tags}
                 %\label{}
         \end{subfigure}
         \begin{subfigure}[b]{0.48\columnwidth}
                 \centering
                 \includegraphics[width=\textwidth]{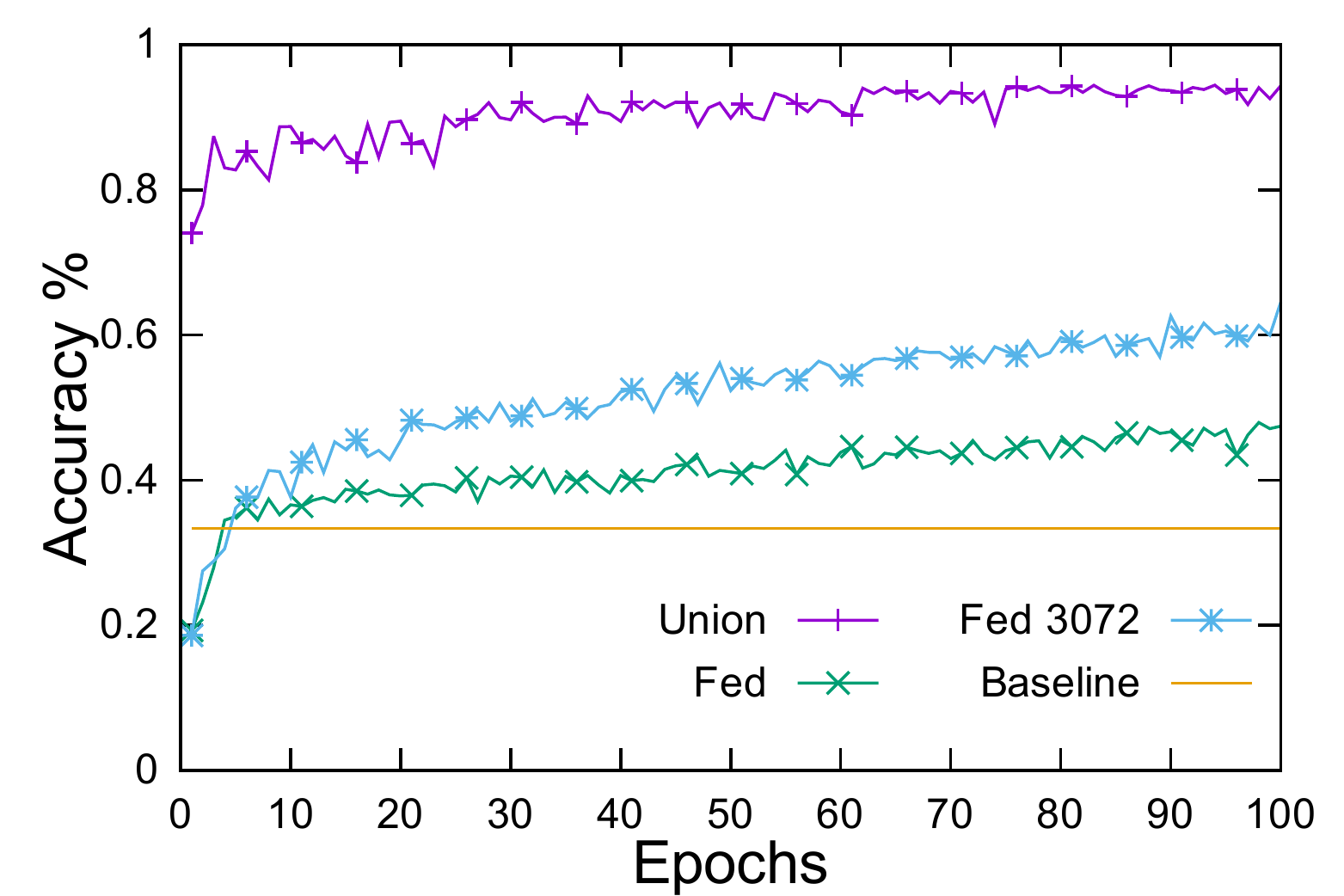}
                 \caption{50 Tags}
                 %\label{}
         \end{subfigure}
% leave a blank line to change row         

     \begin{subfigure}[b]{0.48\columnwidth}
             \centering
             \includegraphics[width=\textwidth]{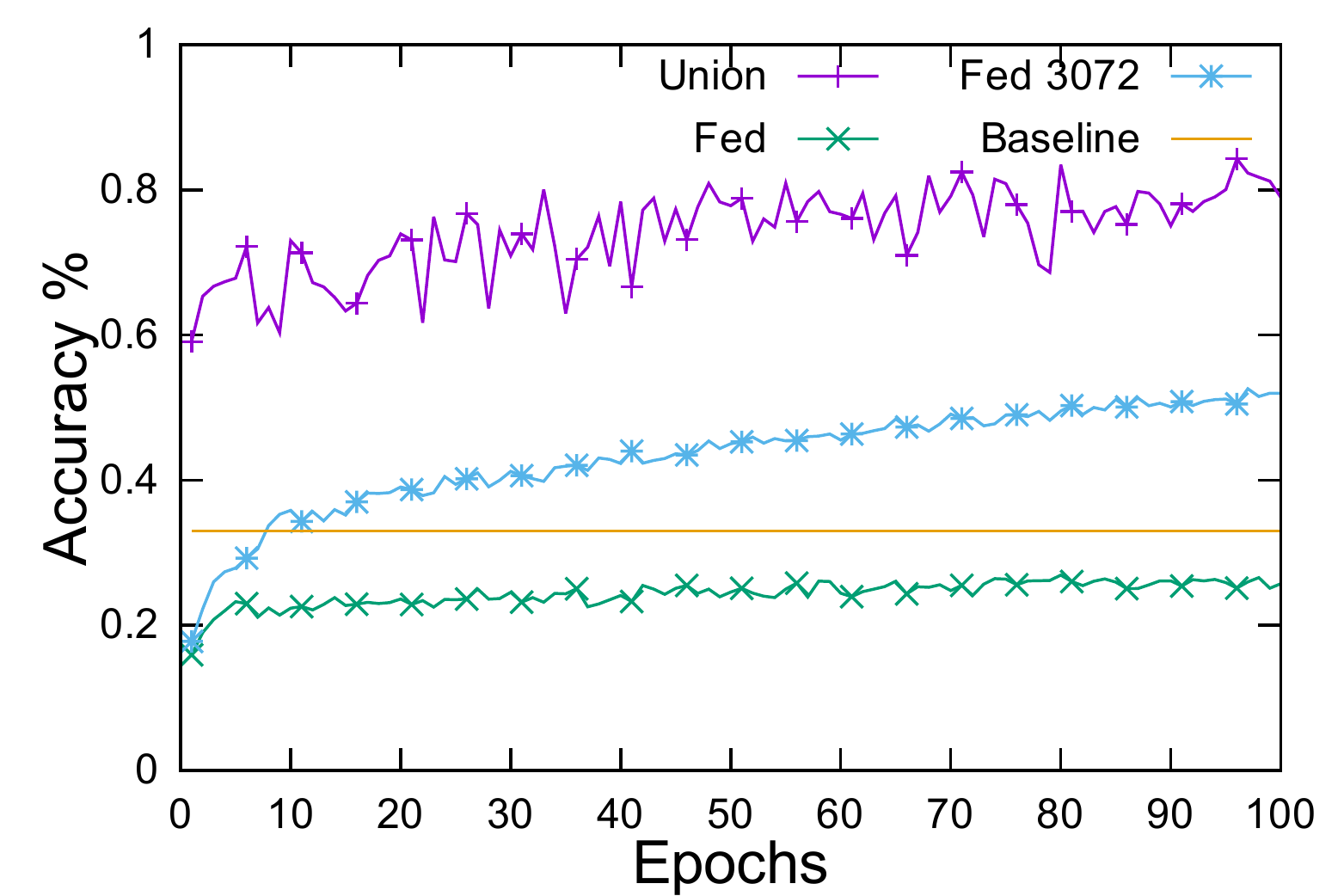}
             \caption{100 Tags}
             %\label{}
     \end{subfigure}
     \begin{subfigure}[b]{0.48\columnwidth}
             \centering
             \includegraphics[width=\textwidth]{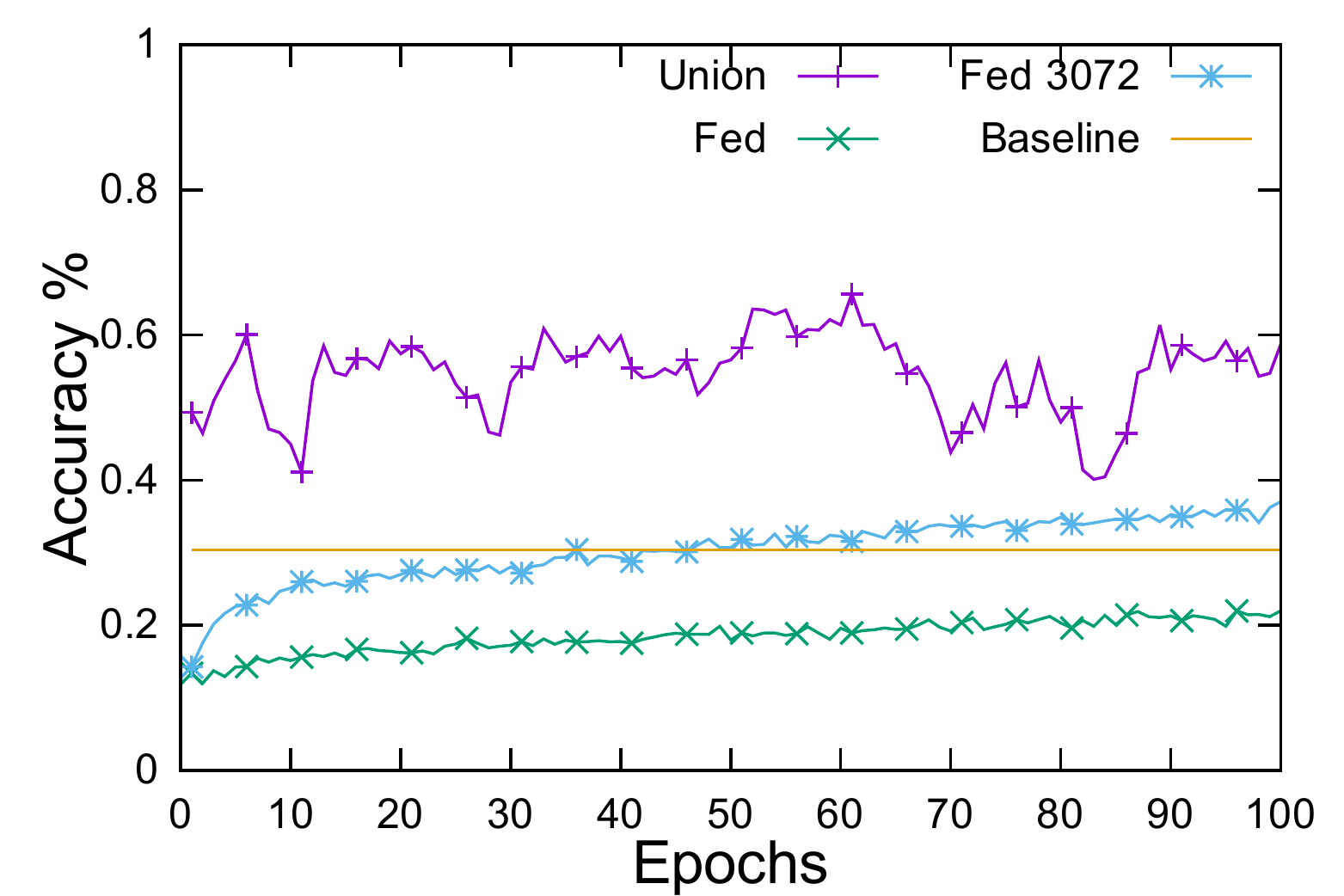}
             \caption{200 Tags}
             %\label{}
      \end{subfigure}
\caption{Accuracy of Federated RFP with  100 percent of data.\vspace{-0.3cm}} \label{fig:res_federated100percent}
\end{figure}

 Figure \ref{fig:res_federated10percent} and \ref{fig:res_federated100percent} depict the performance of federated \gls{rfp} with a portion (10\%) and the total (100\%) of the dataset. This was done to evaluate its performance with  fewer data available. As depicted in Figure \ref{fig:res_federated10percent}, Union is able to generalize and recognize the tags at different distances, with an accuracy that is always at least two times better than the lower bound. This demonstrate how sharing the entire dataset produces a benefit on the accuracy. The results show that the federated \gls{rfp} approach, with only the 10\% of the data available, improves more than 10 times with respect to the Baseline. With 20 tags, federated RFP is less then 10\% from Union. This distance however increases with the number of tags. However, the increase on the number of tags produces a drop of the accuracy even on the optimum, which highlights a limit on the data available more than on the federated approach.

For this reason, we also investigated the size of the learning windows, increasing it from 1024 to 3072 samples. With this change, the model is able to find more impairments per input, and thus we obtain advantages mainly in the scenarios with more nodes and more data. This suggests that more I/Q samples are needed to distinguish signals when the population is higher, which is consistent with prior art. In scenarios with less tags, like depicted in fig.\ref{fig:res_federated10percent}, is instead easier to find unique tags features and the window size increase does not produce benefits. However, this approach still seems to be not really robust, as with 200 tags and 100\% data it reaches up to 40\% of accuracy. This drop may still be related to the model learning channel features and interference. 
\textbf{It is thus clear that the federated \gls{rfp} approach, with only the 10\% of the data available, produces a huge improvements on results, obtaining an accuracy more than 10 times higher with respect to the Baseline.
On the other hand, this approach does not appear to be robust to nodes population.}

%%FINE FED 100 PERC
\subsection{Data Augmentation}\label{sec:aug_results}

\begin{figure}[h!]
         \begin{subfigure}[b]{0.48\columnwidth}
                 \centering
                 \includegraphics[width=\textwidth]{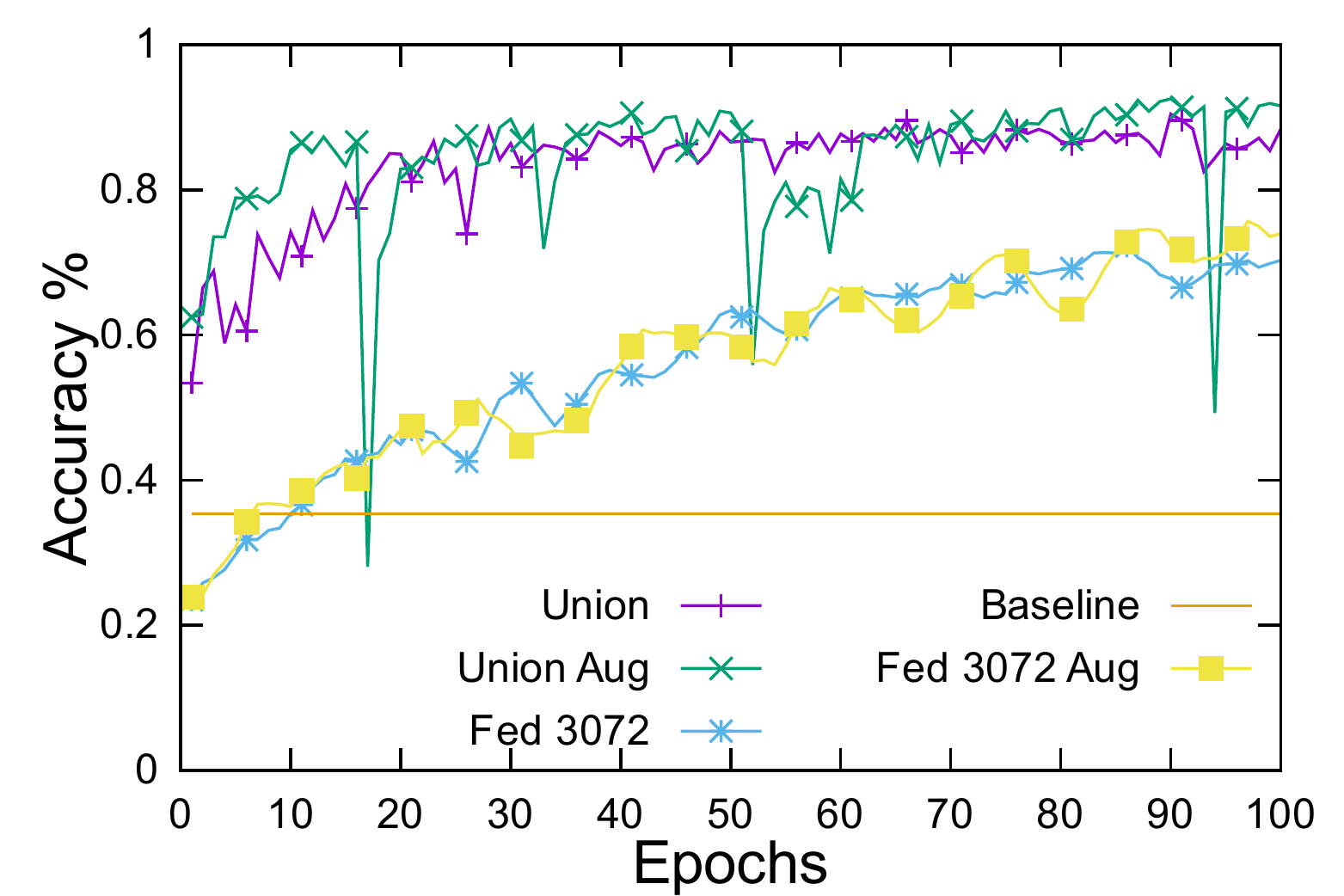}
                 \caption{20 Tags}
                 %\label{}
         \end{subfigure}
         \begin{subfigure}[b]{0.48\columnwidth}
                 \centering
                 \includegraphics[width=\textwidth]{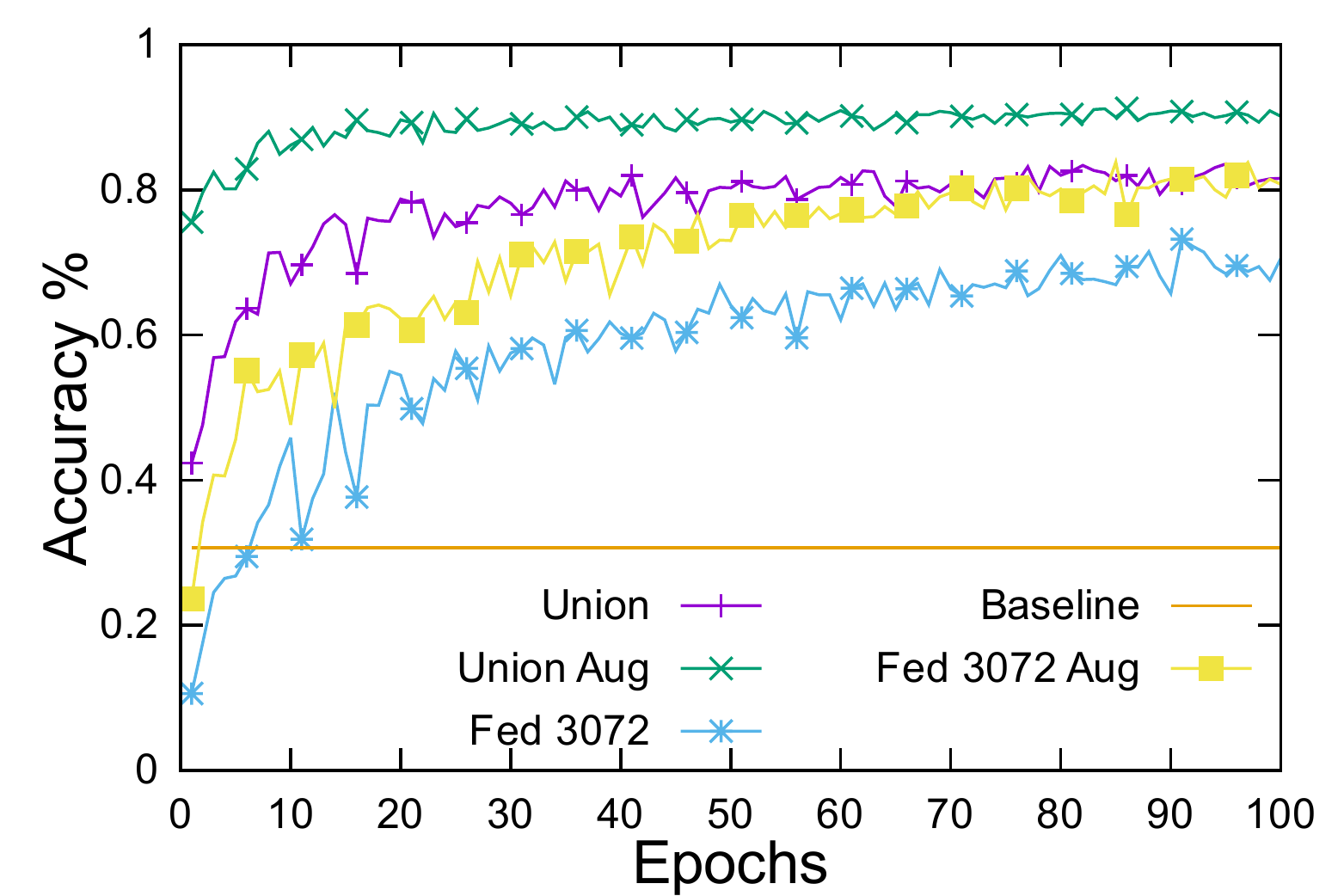}
                 \caption{50 Tags}
                 %\label{}
         \end{subfigure}
% leave a blank line to change row         

     \begin{subfigure}[b]{0.48\columnwidth}
             \centering
             \includegraphics[width=\textwidth]{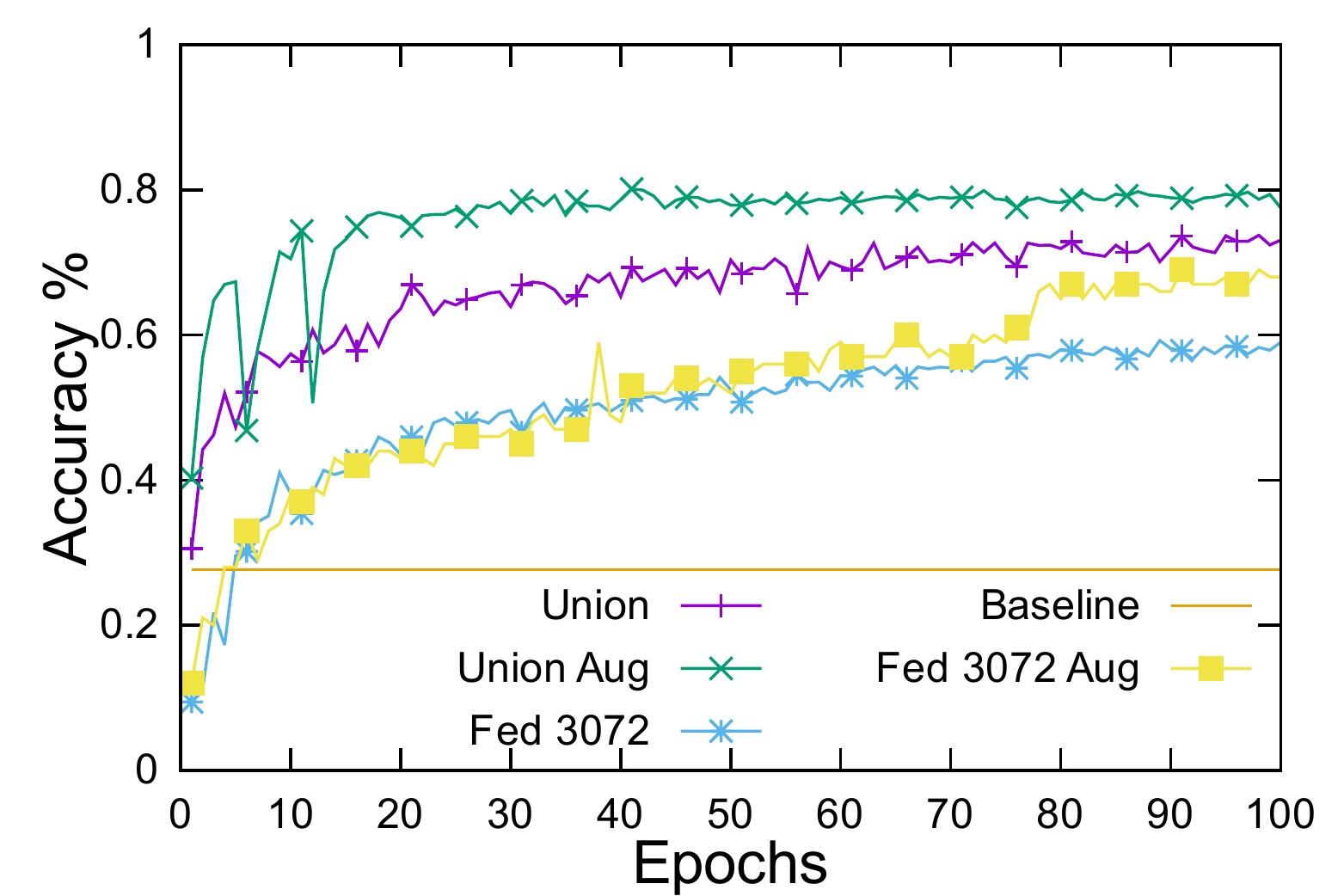}
             \caption{100 Tags}
             %\label{}
     \end{subfigure}
     \begin{subfigure}[b]{0.48\columnwidth}
             \centering
             \includegraphics[width=\textwidth]{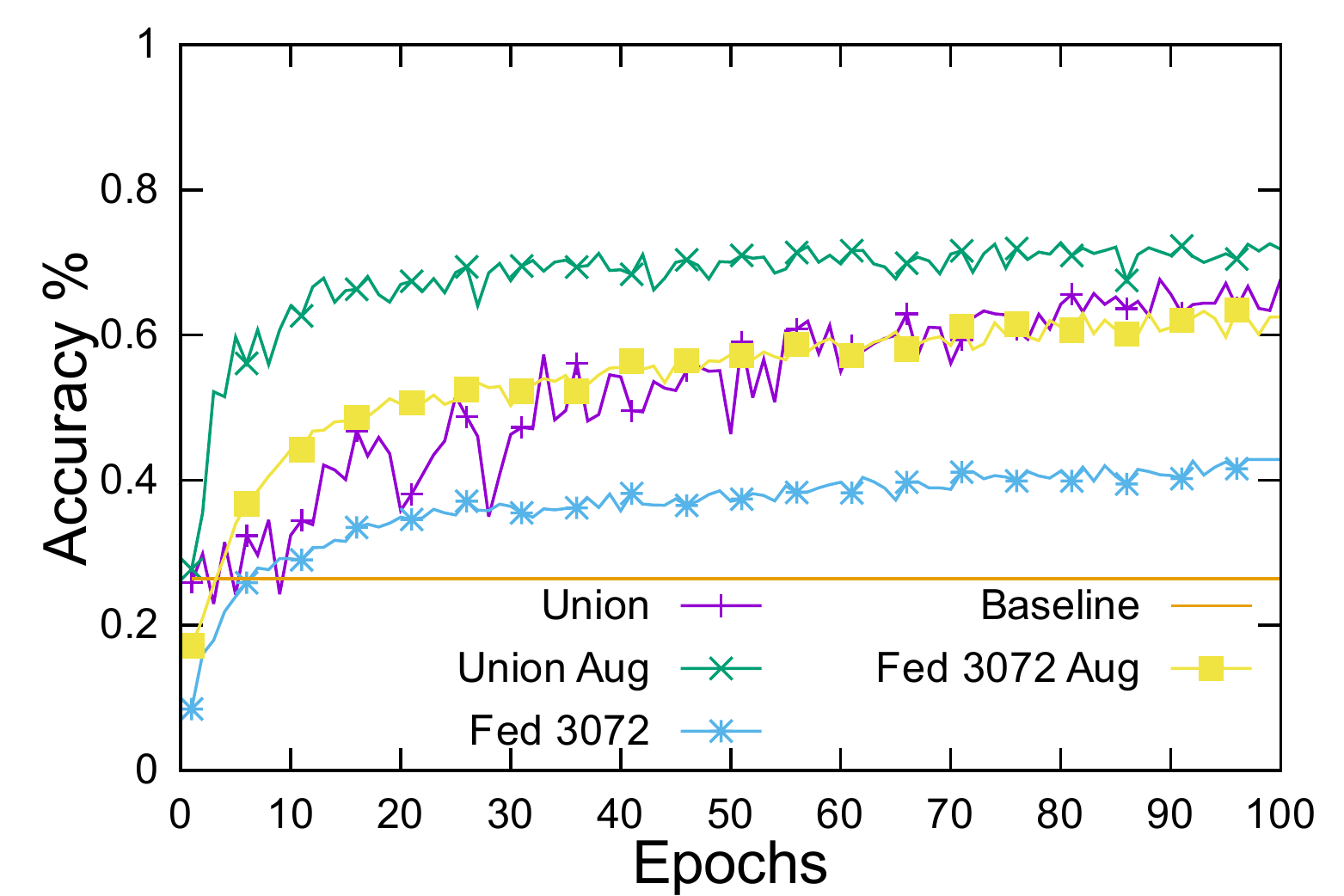}
             \caption{200 Tags}
             %\label{}
      \end{subfigure}
\caption{Accuracy of Data-Augmented RFP with 10 percent of data.} \label{fig:res_augment10percent}
\end{figure}

We now evaluate the data-augmented \gls{rfp} technique described in Section \ref{sec:augmentation}. Figure \ref{fig:res_augment10percent} shows the results with 10\% of the data obtained on the OTA subdatasets. 
As done earlier, we report the results obtained with Union and the Baseline. We can see the advantages of this approach, especially when the population size increases. 
\textbf{These results show that with only 10\% of the dataset, the accuracy of federated \gls{rfp} is increased by an impressive 20\% with 200 tags, becoming comparable with the accuracy in the Union dataset.}
This improvement makes the accuracy obtained by a federated model comparable with the accuracy of the optimum model, confirming the capabilities of the federated approach to deal with noisy and variable channel environments. This impressive effectiveness of data augmentation is also confirmed with the optimum model. Applying data augmentation on that model guarantees an improvement in the accuracy of up to 10\%, as shown with 50 tags. This improvement pattern may be related to the extensive number of different channels recorded with many tags. We recall that the tags collection has been performed over 2 weeks, and that the first tags may have a completely different set of interference as the data collection has been performed in a normal office.  Thus, data augmentation seems to be particularly efficient in realistic scenarios, as it is able to deal with temporary and different interference.

\begin{figure}[h!]
         \begin{subfigure}[b]{0.48\columnwidth}
                 \centering
                 \includegraphics[width=\textwidth]{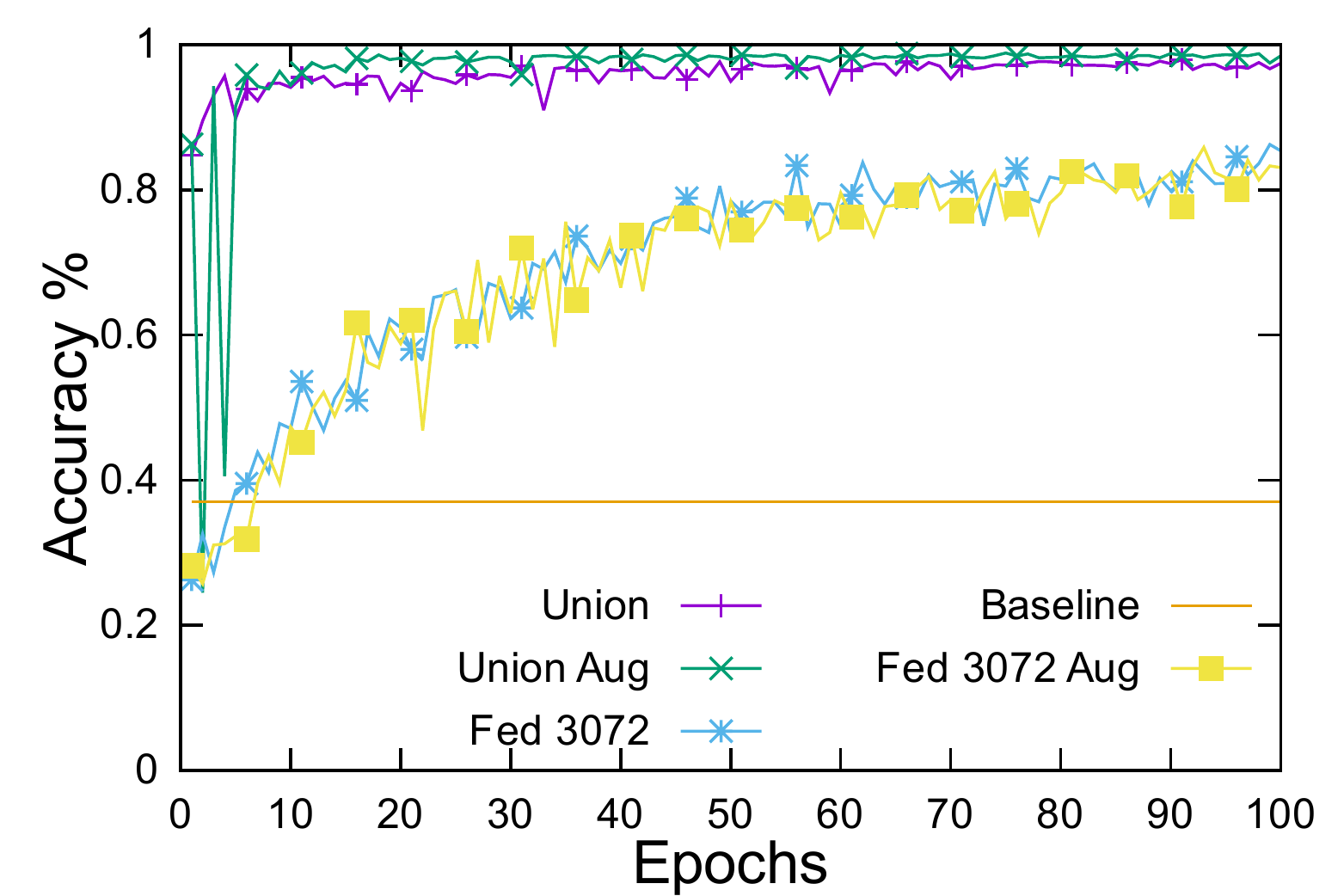}
                 \caption{20 Tags}
                 %\label{}
         \end{subfigure}
         \begin{subfigure}[b]{0.48\columnwidth}
                 \centering
                 \includegraphics[width=\textwidth]{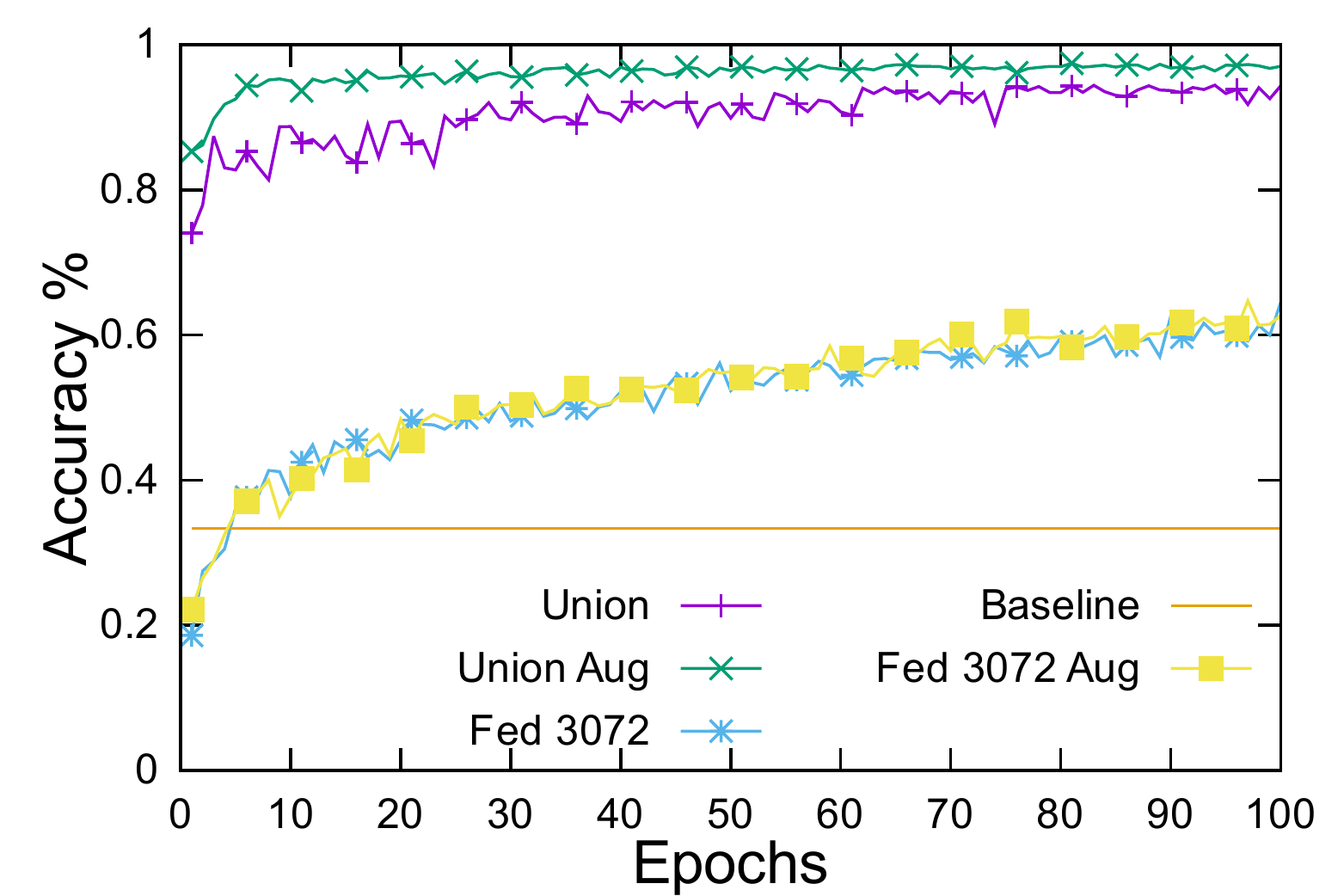}
                 \caption{50 Tags}
                 %\label{}
         \end{subfigure}
% leave a blank line to change row         

     \begin{subfigure}[b]{0.48\columnwidth}
             \centering
             \includegraphics[width=\textwidth]{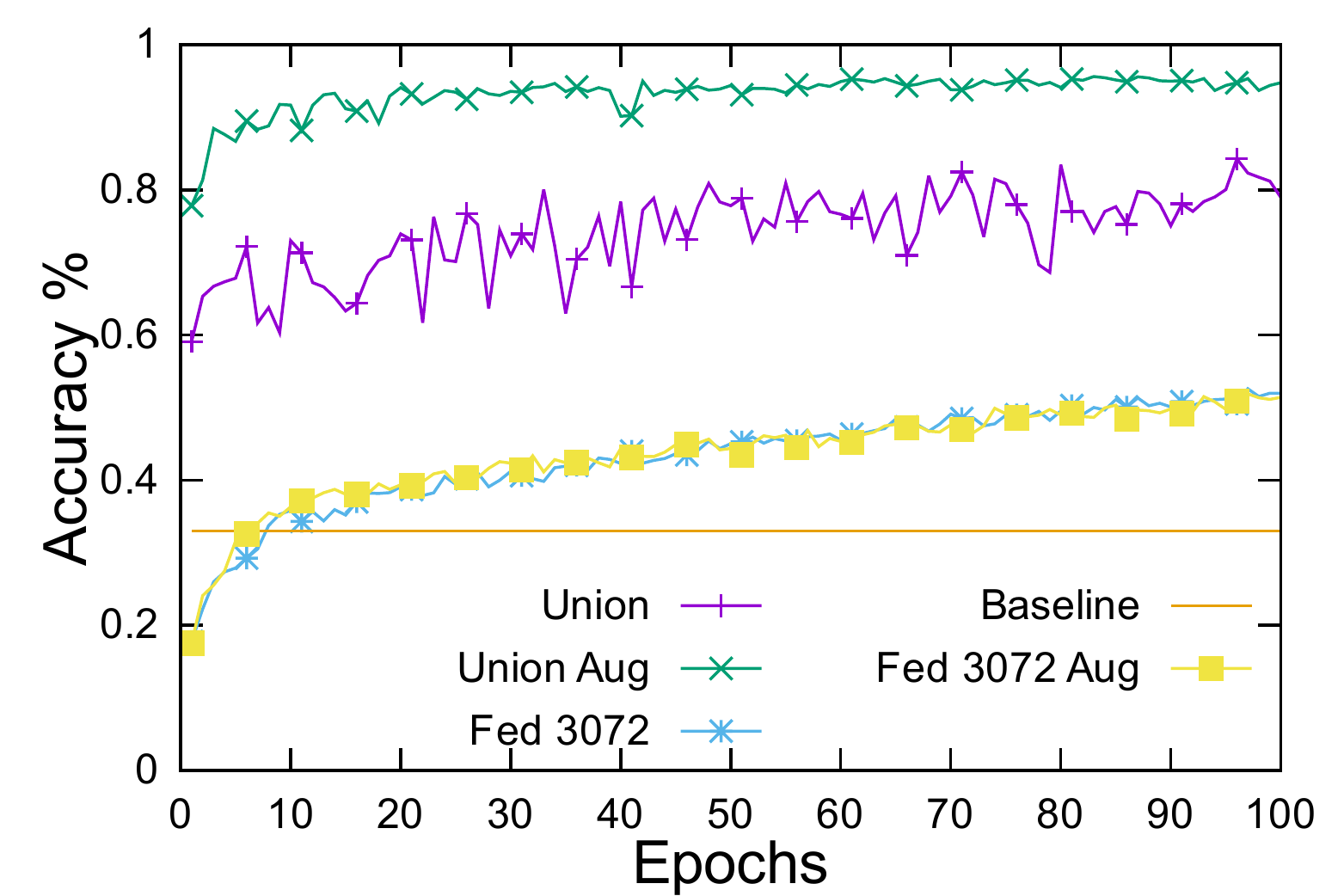}
             \caption{100 Tags}
             %\label{}
     \end{subfigure}
     \begin{subfigure}[b]{0.48\columnwidth}
             \centering
             \includegraphics[width=\textwidth]{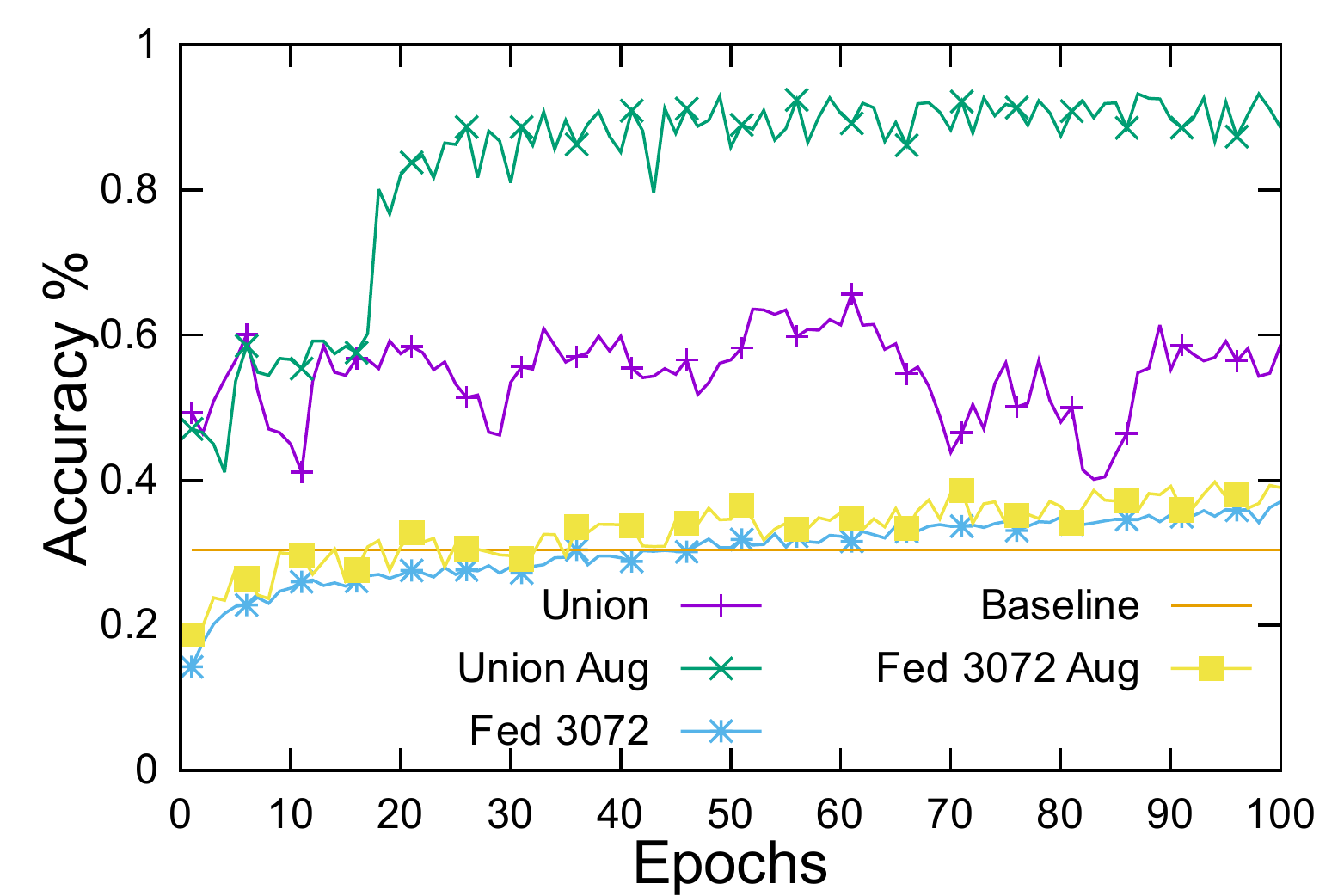}
             \caption{200 Tags}
             %\label{}
      \end{subfigure}
\caption{Accuracy of Data-Augmentated RFP with 100 percent of data.} \label{fig:res_augment100percent}
\end{figure}

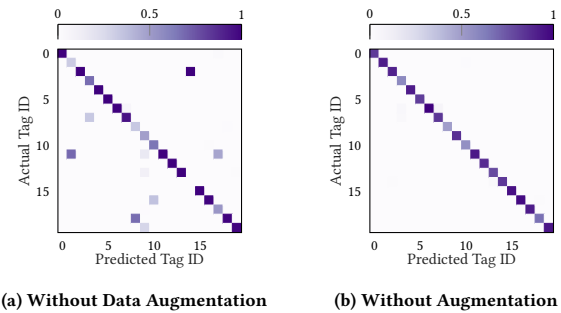
\begin{figure}[h]
    \centering
    \captionsetup{justification=centering}
    \begin{subfigure}[t]{0.48\columnwidth}
    \centering
    \setlength\fwidth{0.6\columnwidth}
    \setlength\fheight{0.6\columnwidth}
    \begin{tikzpicture}
\pgfplotsset{every tick label/.append style={font=\tiny}}

\begin{axis}[
enlargelimits=false,
colorbar,
colormap/Purples,
width=\fwidth,
height=\fheight,
at={(0\fwidth,0\fheight)},
scale only axis,
tick align=inside,
xlabel={Predicted Tag ID},
xmin=-0.5,
xmax=19.5,
xtick style={draw=none},
xlabel style={font=\scriptsize\color{white!15!black}},
ylabel style={font=\scriptsize\color{white!15!black}},
ylabel={Actual Tag ID},
ymin=-0.5,
ymax=19.5,
xlabel shift=-5pt,
ylabel shift=-5pt,
ytick style={draw=none},
axis background/.style={fill=white},
colorbar horizontal,
colorbar style={
at={(0,1.05)},               % <-- (changed)
anchor=below south west,    % <-- (changed)
% change the width of the colorbar relative to the main `axis' environment
width=\pgfkeysvalueof{/pgfplots/parent axis width},
xtick={0, 0.5, 1},
xmin=0,
xmax=1,
axis x line*=top,
xticklabel shift=-1pt,
point meta min=0,
point meta max=1,
},
colorbar/width=2mm,
]\addplot [matrix plot,point meta=explicit]
 coordinates {
(0,0) [1.0] (0,1) [0.0] (0,2) [0.0] (0,3) [0.0] (0,4) [0.0] (0,5) [0.0] (0,6) [0.0] (0,7) [0.0] (0,8) [0.0] (0,9) [0.0] (0,10) [0.0] (0,11) [0.0] (0,12) [0.0] (0,13) [0.0] (0,14) [0.0] (0,15) [0.0] (0,16) [0.0] (0,17) [0.0] (0,18) [0.0] (0,19) [0.0] 

(1,0) [0.0] (1,1) [0.31451612903225806] (1,2) [0.0] (1,3) [0.0] (1,4) [0.0] (1,5) [0.0] (1,6) [0.0] (1,7) [0.0] (1,8) [0.0] (1,9) [0.0] (1,10) [0.0] (1,11) [0.6854838709677419] (1,12) [0.0] (1,13) [0.0] (1,14) [0.0] (1,15) [0.0] (1,16) [0.0] (1,17) [0.0] (1,18) [0.0] (1,19) [0.0] 

(2,0) [0.0] (2,1) [0.0] (2,2) [1.0] (2,3) [0.0] (2,4) [0.0] (2,5) [0.0] (2,6) [0.0] (2,7) [0.0] (2,8) [0.0] (2,9) [0.0] (2,10) [0.0] (2,11) [0.0] (2,12) [0.0] (2,13) [0.0] (2,14) [0.0] (2,15) [0.0] (2,16) [0.0] (2,17) [0.0] (2,18) [0.0] (2,19) [0.0] 

(3,0) [0.0] (3,1) [0.0] (3,2) [0.0] (3,3) [0.672053872053872] (3,4) [0.0] (3,5) [0.0] (3,6) [0.0] (3,7) [0.32794612794612793] (3,8) [0.0] (3,9) [0.0] (3,10) [0.0] (3,11) [0.0] (3,12) [0.0] (3,13) [0.0] (3,14) [0.0] (3,15) [0.0] (3,16) [0.0] (3,17) [0.0] (3,18) [0.0] (3,19) [0.0] 

(4,0) [0.0] (4,1) [0.0] (4,2) [0.0] (4,3) [0.0] (4,4) [0.9986559139784946] (4,5) [0.0] (4,6) [0.0] (4,7) [0.0] (4,8) [0.0006720430107526882] (4,9) [0.0] (4,10) [0.0006720430107526882] (4,11) [0.0] (4,12) [0.0] (4,13) [0.0] (4,14) [0.0] (4,15) [0.0] (4,16) [0.0] (4,17) [0.0] (4,18) [0.0] (4,19) [0.0] 

(5,0) [0.0] (5,1) [0.0] (5,2) [0.0] (5,3) [0.0] (5,4) [0.0] (5,5) [1.0] (5,6) [0.0] (5,7) [0.0] (5,8) [0.0] (5,9) [0.0] (5,10) [0.0] (5,11) [0.0] (5,12) [0.0] (5,13) [0.0] (5,14) [0.0] (5,15) [0.0] (5,16) [0.0] (5,17) [0.0] (5,18) [0.0] (5,19) [0.0] 

(6,0) [0.0] (6,1) [0.0] (6,2) [0.0] (6,3) [0.0] (6,4) [0.0] (6,5) [0.0] (6,6) [1.0] (6,7) [0.0] (6,8) [0.0] (6,9) [0.0] (6,10) [0.0] (6,11) [0.0] (6,12) [0.0] (6,13) [0.0] (6,14) [0.0] (6,15) [0.0] (6,16) [0.0] (6,17) [0.0] (6,18) [0.0] (6,19) [0.0] 

(7,0) [0.0] (7,1) [0.0] (7,2) [0.0] (7,3) [0.002688172043010753] (7,4) [0.0] (7,5) [0.0] (7,6) [0.051075268817204304] (7,7) [0.946236559139785] (7,8) [0.0] (7,9) [0.0] (7,10) [0.0] (7,11) [0.0] (7,12) [0.0] (7,13) [0.0] (7,14) [0.0] (7,15) [0.0] (7,16) [0.0] (7,17) [0.0] (7,18) [0.0] (7,19) [0.0] 

(8,0) [0.0] (8,1) [0.0] (8,2) [0.0] (8,3) [0.0] (8,4) [0.0] (8,5) [0.0] (8,6) [0.0] (8,7) [0.0] (8,8) [0.32796780684104626] (8,9) [0.0] (8,10) [0.0] (8,11) [0.0] (8,12) [0.0] (8,13) [0.0] (8,14) [0.0] (8,15) [0.0] (8,16) [0.0] (8,17) [0.0] (8,18) [0.6720321931589537] (8,19) [0.0] 

(9,0) [0.0] (9,1) [0.0] (9,2) [0.0] (9,3) [0.0] (9,4) [0.0] (9,5) [0.0] (9,6) [0.0] (9,7) [0.0] (9,8) [0.0] (9,9) [0.48148148148148145] (9,10) [0.0] (9,11) [0.15218855218855218] (9,12) [0.0] (9,13) [0.09427609427609428] (9,14) [0.0] (9,15) [0.0] (9,16) [0.0] (9,17) [0.0047138047138047135] (9,18) [0.0] (9,19) [0.2673400673400673] 

(10,0) [0.0] (10,1) [0.0] (10,2) [0.0] (10,3) [0.0] (10,4) [0.0] (10,5) [0.0] (10,6) [0.0] (10,7) [0.0] (10,8) [0.0] (10,9) [0.0] (10,10) [0.6457912457912458] (10,11) [0.0] (10,12) [0.0] (10,13) [0.0] (10,14) [0.0] (10,15) [0.0] (10,16) [0.3542087542087542] (10,17) [0.0] (10,18) [0.0] (10,19) [0.0] 

(11,0) [0.0] (11,1) [0.0] (11,2) [0.0] (11,3) [0.0] (11,4) [0.0] (11,5) [0.0] (11,6) [0.0] (11,7) [0.0] (11,8) [0.0] (11,9) [0.0] (11,10) [0.0] (11,11) [0.9959677419354839] (11,12) [0.0] (11,13) [0.004032258064516129] (11,14) [0.0] (11,15) [0.0] (11,16) [0.0] (11,17) [0.0] (11,18) [0.0] (11,19) [0.0] 

(12,0) [0.0] (12,1) [0.0] (12,2) [0.0] (12,3) [0.0] (12,4) [0.0] (12,5) [0.0006706908115358819] (12,6) [0.0] (12,7) [0.0] (12,8) [0.0] (12,9) [0.0] (12,10) [0.0] (12,11) [0.0] (12,12) [0.9993293091884641] (12,13) [0.0] (12,14) [0.0] (12,15) [0.0] (12,16) [0.0] (12,17) [0.0] (12,18) [0.0] (12,19) [0.0] 

(13,0) [0.0] (13,1) [0.0] (13,2) [0.0] (13,3) [0.0] (13,4) [0.0] (13,5) [0.0] (13,6) [0.0] (13,7) [0.0] (13,8) [0.0] (13,9) [0.0] (13,10) [0.0] (13,11) [0.0013468013468013469] (13,12) [0.0] (13,13) [0.9986531986531987] (13,14) [0.0] (13,15) [0.0] (13,16) [0.0] (13,17) [0.0] (13,18) [0.0] (13,19) [0.0] 

(14,0) [0.0] (14,1) [0.0] (14,2) [1.0] (14,3) [0.0] (14,4) [0.0] (14,5) [0.0] (14,6) [0.0] (14,7) [0.0] (14,8) [0.0] (14,9) [0.0] (14,10) [0.0] (14,11) [0.0] (14,12) [0.0] (14,13) [0.0] (14,14) [0.0] (14,15) [0.0] (14,16) [0.0] (14,17) [0.0] (14,18) [0.0] (14,19) [0.0] 

(15,0) [0.0] (15,1) [0.0] (15,2) [0.0] (15,3) [0.0] (15,4) [0.0] (15,5) [0.0] (15,6) [0.0] (15,7) [0.0] (15,8) [0.0] (15,9) [0.0] (15,10) [0.0] (15,11) [0.0] (15,12) [0.0] (15,13) [0.0] (15,14) [0.0] (15,15) [1.0] (15,16) [0.0] (15,17) [0.0] (15,18) [0.0] (15,19) [0.0] 

(16,0) [0.0] (16,1) [0.0] (16,2) [0.0] (16,3) [0.0] (16,4) [0.0] (16,5) [0.002699055330634278] (16,6) [0.0] (16,7) [0.0] (16,8) [0.0] (16,9) [0.0] (16,10) [0.010796221322537112] (16,11) [0.0] (16,12) [0.0] (16,13) [0.0] (16,14) [0.0] (16,15) [0.0] (16,16) [0.9865047233468286] (16,17) [0.0] (16,18) [0.0] (16,19) [0.0] 

(17,0) [0.021462105969148222] (17,1) [0.0] (17,2) [0.0] (17,3) [0.0] (17,4) [0.0] (17,5) [0.0] (17,6) [0.0] (17,7) [0.0] (17,8) [0.0] (17,9) [0.0] (17,10) [0.0] (17,11) [0.4607645875251509] (17,12) [0.0] (17,13) [0.0] (17,14) [0.0] (17,15) [0.0] (17,16) [0.0] (17,17) [0.5177733065057009] (17,18) [0.0] (17,19) [0.0] 

(18,0) [0.0] (18,1) [0.0] (18,2) [0.0] (18,3) [0.0] (18,4) [0.0] (18,5) [0.0] (18,6) [0.0] (18,7) [0.0] (18,8) [0.0053872053872053875] (18,9) [0.0] (18,10) [0.0] (18,11) [0.0] (18,12) [0.0] (18,13) [0.0] (18,14) [0.0] (18,15) [0.0] (18,16) [0.0] (18,17) [0.0] (18,18) [0.9946127946127946] (18,19) [0.0] 

(19,0) [0.0] (19,1) [0.0] (19,2) [0.0] (19,3) [0.0] (19,4) [0.0] (19,5) [0.0] (19,6) [0.0] (19,7) [0.0] (19,8) [0.0] (19,9) [0.002688172043010753] (19,10) [0.0] (19,11) [0.0020161290322580645] (19,12) [0.0] (19,13) [0.010752688172043012] (19,14) [0.0] (19,15) [0.0] (19,16) [0.0] (19,17) [0.0] (19,18) [0.0] (19,19) [0.9845430107526881] 

};
\end{axis}
\end{tikzpicture}
    \caption{Without Data Augmentation}
    \label{fig:cm_3}
    \end{subfigure}
    \begin{subfigure}[t]{0.48\columnwidth}
    \centering
    \setlength\fwidth{0.6\columnwidth}
    \setlength\fheight{0.6\columnwidth}
    \begin{tikzpicture}
\pgfplotsset{every tick label/.append style={font=\tiny}}

\begin{axis}[
enlargelimits=false,
colorbar,
colormap/Purples,
width=\fwidth,
height=\fheight,
at={(0\fwidth,0\fheight)},
scale only axis,
tick align=inside,
xlabel={Predicted Tag ID},
xmin=-0.5,
xmax=19.5,
xtick style={draw=none},
xlabel style={font=\scriptsize\color{white!15!black}},
ylabel style={font=\scriptsize\color{white!15!black}},
ylabel={Actual Tag ID},
ymin=-0.5,
ymax=19.5,
xlabel shift=-5pt,
ylabel shift=-5pt,
ytick style={draw=none},
axis background/.style={fill=white},
colorbar horizontal,
colorbar style={
at={(0,1.05)},               % <-- (changed)
anchor=below south west,    % <-- (changed)
% change the width of the colorbar relative to the main `axis' environment
width=\pgfkeysvalueof{/pgfplots/parent axis width},
xtick={0, 0.5, 1},
xmin=0,
xmax=1,
axis x line*=top,
xticklabel shift=-1pt,
point meta min=0,
point meta max=1,
},
colorbar/width=2mm,
]\addplot [matrix plot,point meta=explicit]
 coordinates {
(0,0) [0.8188552188552188] (0,1) [0.0] (0,2) [0.0] (0,3) [0.0] (0,4) [0.0] (0,5) [0.0] (0,6) [0.0] (0,7) [0.0] (0,8) [0.0] (0,9) [0.0] (0,10) [0.0] (0,11) [0.0] (0,12) [0.0] (0,13) [0.0] (0,14) [0.0] (0,15) [0.0] (0,16) [0.0] (0,17) [0.0] (0,18) [0.0] (0,19) [0.0] 

(1,0) [0.0] (1,1) [0.9086021505376344] (1,2) [0.0] (1,3) [0.0] (1,4) [0.0] (1,5) [0.0] (1,6) [0.0] (1,7) [0.0] (1,8) [0.0] (1,9) [0.0] (1,10) [0.0] (1,11) [0.0] (1,12) [0.0] (1,13) [0.0] (1,14) [0.0] (1,15) [0.0] (1,16) [0.0] (1,17) [0.0] (1,18) [0.0] (1,19) [0.0] 

(2,0) [0.0] (2,1) [0.0] (2,2) [0.892018779342723] (2,3) [0.0] (2,4) [0.0] (2,5) [0.0] (2,6) [0.0] (2,7) [0.0] (2,8) [0.0] (2,9) [0.0] (2,10) [0.0] (2,11) [0.0] (2,12) [0.0] (2,13) [0.0013413816230717639] (2,14) [0.009389671361502348] (2,15) [0.0] (2,16) [0.0] (2,17) [0.0] (2,18) [0.0] (2,19) [0.0] 

(3,0) [0.0] (3,1) [0.0] (3,2) [0.0] (3,3) [0.5703703703703704] (3,4) [0.0] (3,5) [0.0] (3,6) [0.035016835016835016] (3,7) [0.028956228956228958] (3,8) [0.0] (3,9) [0.0] (3,10) [0.0] (3,11) [0.0] (3,12) [0.0] (3,13) [0.0] (3,14) [0.0] (3,15) [0.0] (3,16) [0.0] (3,17) [0.0] (3,18) [0.0] (3,19) [0.0] 

(4,0) [0.0] (4,1) [0.0] (4,2) [0.0] (4,3) [0.0] (4,4) [0.9119623655913979] (4,5) [0.0] (4,6) [0.0] (4,7) [0.0] (4,8) [0.0] (4,9) [0.0] (4,10) [0.0] (4,11) [0.0] (4,12) [0.0] (4,13) [0.0] (4,14) [0.0] (4,15) [0.0] (4,16) [0.0] (4,17) [0.0] (4,18) [0.0] (4,19) [0.0] 

(5,0) [0.0] (5,1) [0.0] (5,2) [0.0] (5,3) [0.0] (5,4) [0.0] (5,5) [0.8078982597054887] (5,6) [0.0] (5,7) [0.0] (5,8) [0.0] (5,9) [0.0] (5,10) [0.0] (5,11) [0.0] (5,12) [0.0] (5,13) [0.0] (5,14) [0.0] (5,15) [0.0] (5,16) [0.0] (5,17) [0.0] (5,18) [0.0] (5,19) [0.0] 

(6,0) [0.0] (6,1) [0.0] (6,2) [0.0] (6,3) [0.0] (6,4) [0.0] (6,5) [0.0] (6,6) [0.999330655957162] (6,7) [0.0] (6,8) [0.0] (6,9) [0.0] (6,10) [0.0] (6,11) [0.0] (6,12) [0.0] (6,13) [0.0] (6,14) [0.0] (6,15) [0.0] (6,16) [0.0] (6,17) [0.0] (6,18) [0.0] (6,19) [0.0] 

(7,0) [0.0] (7,1) [0.0] (7,2) [0.0] (7,3) [0.002688172043010753] (7,4) [0.0] (7,5) [0.0] (7,6) [0.05913978494623656] (7,7) [0.8286290322580645] (7,8) [0.0] (7,9) [0.0] (7,10) [0.0] (7,11) [0.0] (7,12) [0.0] (7,13) [0.0] (7,14) [0.0] (7,15) [0.0] (7,16) [0.0] (7,17) [0.0] (7,18) [0.0] (7,19) [0.0] 

(8,0) [0.0] (8,1) [0.0] (8,2) [0.0] (8,3) [0.0] (8,4) [0.0] (8,5) [0.0] (8,6) [0.0] (8,7) [0.0] (8,8) [0.4869215291750503] (8,9) [0.0] (8,10) [0.0] (8,11) [0.0] (8,12) [0.0] (8,13) [0.0] (8,14) [0.0] (8,15) [0.0] (8,16) [0.0] (8,17) [0.0] (8,18) [0.0] (8,19) [0.0] 

(9,0) [0.0] (9,1) [0.0] (9,2) [0.0] (9,3) [0.0] (9,4) [0.0] (9,5) [0.0] (9,6) [0.0] (9,7) [0.0] (9,8) [0.0] (9,9) [0.8505050505050505] (9,10) [0.0] (9,11) [0.0] (9,12) [0.0] (9,13) [0.0] (9,14) [0.0] (9,15) [0.0] (9,16) [0.0] (9,17) [0.0] (9,18) [0.0] (9,19) [0.0] 

(10,0) [0.0] (10,1) [0.005376344086021506] (10,2) [0.0] (10,3) [0.0] (10,4) [0.0] (10,5) [0.0] (10,6) [0.0] (10,7) [0.0] (10,8) [0.0] (10,9) [0.0] (10,10) [0.5443548387096774] (10,11) [0.0] (10,12) [0.0] (10,13) [0.0] (10,14) [0.0] (10,15) [0.0] (10,16) [0.0] (10,17) [0.0] (10,18) [0.0] (10,19) [0.0] 

(11,0) [0.0] (11,1) [0.0] (11,2) [0.0] (11,3) [0.0] (11,4) [0.0] (11,5) [0.0] (11,6) [0.0] (11,7) [0.0] (11,8) [0.0] (11,9) [0.0] (11,10) [0.0] (11,11) [0.9203778677462888] (11,12) [0.0] (11,13) [0.0] (11,14) [0.0] (11,15) [0.0] (11,16) [0.0] (11,17) [0.0] (11,18) [0.0] (11,19) [0.0] 

(12,0) [0.0006775067750677507] (12,1) [0.0] (12,2) [0.0] (12,3) [0.0] (12,4) [0.0] (12,5) [0.0] (12,6) [0.0] (12,7) [0.0] (12,8) [0.0] (12,9) [0.0] (12,10) [0.0] (12,11) [0.0] (12,12) [0.8644986449864499] (12,13) [0.0] (12,14) [0.0] (12,15) [0.0] (12,16) [0.0] (12,17) [0.0] (12,18) [0.0] (12,19) [0.0] 

(13,0) [0.0] (13,1) [0.0] (13,2) [0.0] (13,3) [0.0] (13,4) [0.0] (13,5) [0.0] (13,6) [0.0] (13,7) [0.0] (13,8) [0.0] (13,9) [0.0] (13,10) [0.0] (13,11) [0.0] (13,12) [0.0] (13,13) [0.7686116700201208] (13,14) [0.03085177733065057] (13,15) [0.0] (13,16) [0.0] (13,17) [0.0] (13,18) [0.0] (13,19) [0.0] 

(14,0) [0.0] (14,1) [0.0] (14,2) [0.0] (14,3) [0.0] (14,4) [0.0] (14,5) [0.0] (14,6) [0.0] (14,7) [0.0] (14,8) [0.0] (14,9) [0.0] (14,10) [0.0] (14,11) [0.0] (14,12) [0.0] (14,13) [0.0006734006734006734] (14,14) [0.8208754208754209] (14,15) [0.0] (14,16) [0.0] (14,17) [0.0] (14,18) [0.0] (14,19) [0.0] 

(15,0) [0.0] (15,1) [0.0] (15,2) [0.0] (15,3) [0.0] (15,4) [0.0] (15,5) [0.0] (15,6) [0.0] (15,7) [0.0] (15,8) [0.0] (15,9) [0.0] (15,10) [0.0] (15,11) [0.0] (15,12) [0.0] (15,13) [0.0] (15,14) [0.0] (15,15) [0.9479377958079783] (15,16) [0.0] (15,17) [0.0] (15,18) [0.0] (15,19) [0.0] 

(16,0) [0.0] (16,1) [0.0] (16,2) [0.0] (16,3) [0.0] (16,4) [0.0] (16,5) [0.0] (16,6) [0.0] (16,7) [0.0] (16,8) [0.0] (16,9) [0.0] (16,10) [0.0] (16,11) [0.0] (16,12) [0.0] (16,13) [0.0] (16,14) [0.0] (16,15) [0.0] (16,16) [0.9676767676767677] (16,17) [0.0] (16,18) [0.0006734006734006734] (16,19) [0.0] 

(17,0) [0.0] (17,1) [0.0] (17,2) [0.0] (17,3) [0.0] (17,4) [0.0] (17,5) [0.0] (17,6) [0.0] (17,7) [0.0] (17,8) [0.0] (17,9) [0.0] (17,10) [0.0] (17,11) [0.0] (17,12) [0.0] (17,13) [0.0] (17,14) [0.0] (17,15) [0.0] (17,16) [0.0] (17,17) [0.9112903225806451] (17,18) [0.0] (17,19) [0.0] 

(18,0) [0.0] (18,1) [0.0] (18,2) [0.0] (18,3) [0.0] (18,4) [0.0] (18,5) [0.0] (18,6) [0.0] (18,7) [0.0] (18,8) [0.0] (18,9) [0.0] (18,10) [0.0] (18,11) [0.0] (18,12) [0.0006720430107526882] (18,13) [0.0] (18,14) [0.0] (18,15) [0.0] (18,16) [0.008064516129032258] (18,17) [0.0] (18,18) [0.6586021505376344] (18,19) [0.0] 

(19,0) [0.0] (19,1) [0.0] (19,2) [0.0] (19,3) [0.0] (19,4) [0.0] (19,5) [0.0] (19,6) [0.0] (19,7) [0.0] (19,8) [0.0] (19,9) [0.0] (19,10) [0.0] (19,11) [0.0] (19,12) [0.0] (19,13) [0.0] (19,14) [0.0] (19,15) [0.0] (19,16) [0.0006706908115358819] (19,17) [0.0] (19,18) [0.0] (19,19) [0.9282360831656606] 

};
\end{axis}
\end{tikzpicture}
    \caption{Without Augmentation}
    \label{fig:cm_4}
    \end{subfigure}
    \centering
    \captionsetup{justification=centering}
  
    \setlength\belowcaptionskip{-.2cm}
    \caption{Bottom 20 devices of the OTA20, OTA50 and OTA100 union with 200 tags and 100\% of data}
        \label{fig:cms_200tags}
\end{figure}

Figure \ref{fig:res_augment100percent} shows the results with 100\% of the OTA subdatasets. First, the results indicate that with more data, the accuracy decreases in all the federated \gls{rfp} scenarios considered. This effect is particularly observed as the population size increases, where the accuracy drops by 20\% in the case of 200-tag when going from 10\% to 100\% of the data. While this result may seem counter intuitive and in contrast with prior art on ML, it is rooted in the principle that \textit{more data does not necessarily mean a better model, if it causes more confusion in the classifier}. Indeed, we see that also in the Union dataset, the learning curves show erratic trends, especially in the 100- and 200-tag datasets. This tells us that by adding more data, we are introducing more confusion into the models, which try to ``overfit the different channels'' without improving the generalization power. This effect is felt stronger in federated \gls{rfp}, which trains on local datasets only and therefore it takes much more time to converge than with less data.
\textbf{Nevertheless, we notice that \gls{da} on the Union dataset returns impressive results, with an accuracy improvement that increases with the population size, and reaching a 38\% improvement in the case of 200 tags.}
This is also represented in Figure \ref{fig:cms_200tags}, where we depict a subportion of the 200-device confusion matrix for the bottom (i.e., lowest accuracy) 20 devices in the dataset, with and without data augmentation.

Finally, in Figure \ref{fig:res_porcinemeat_100a} we report the results obtained by applying \gls{da} on the PM0 and PM1 datasets. With PM{0,1} we mean a federated environment where one client uses PM{0,1}-20 as dataset and the other one uses PM{0,1}-50. Even in this case, we verify that federated \gls{rfp} is able to generalize by reaching an accuracy of 83\% in the PM0 datasets, within 10\% of the Union accuracy. While with PM0 the federated approach obtains acceptable results with a smaller input size of 1024 I/Q samples, PM1 needs to increase the window size to 3072 to obtain results close to Union. The reason is that the PM1 were collected with much thicker porcine meat, which appears to introduce more distortion to the received waveform and thus partially hide the impairments, thus requiring more I/Q samples to achieve an accuracy similar to PM0. 
Figure \ref{fig:cms_20tags} provides a visual indication of how impactful \gls{da} is on the PM1 dataset. Finally, in line with the OTA results, we observe that \gls{da} does not benefit federated \gls{rfp} when the population size is low (PM0 and PM1 have only 20 tags). 
\textbf{Consistently with the OTA case, these results confirm that \gls{da} achieves better performance in more complex scenarios such as PM1, where the Union accuracy goes from 87\% to 97\%}.

\begin{figure}[t!p]
         \begin{subfigure}[b]{0.48\columnwidth}
                 \centering
                 \includegraphics[width=\textwidth]{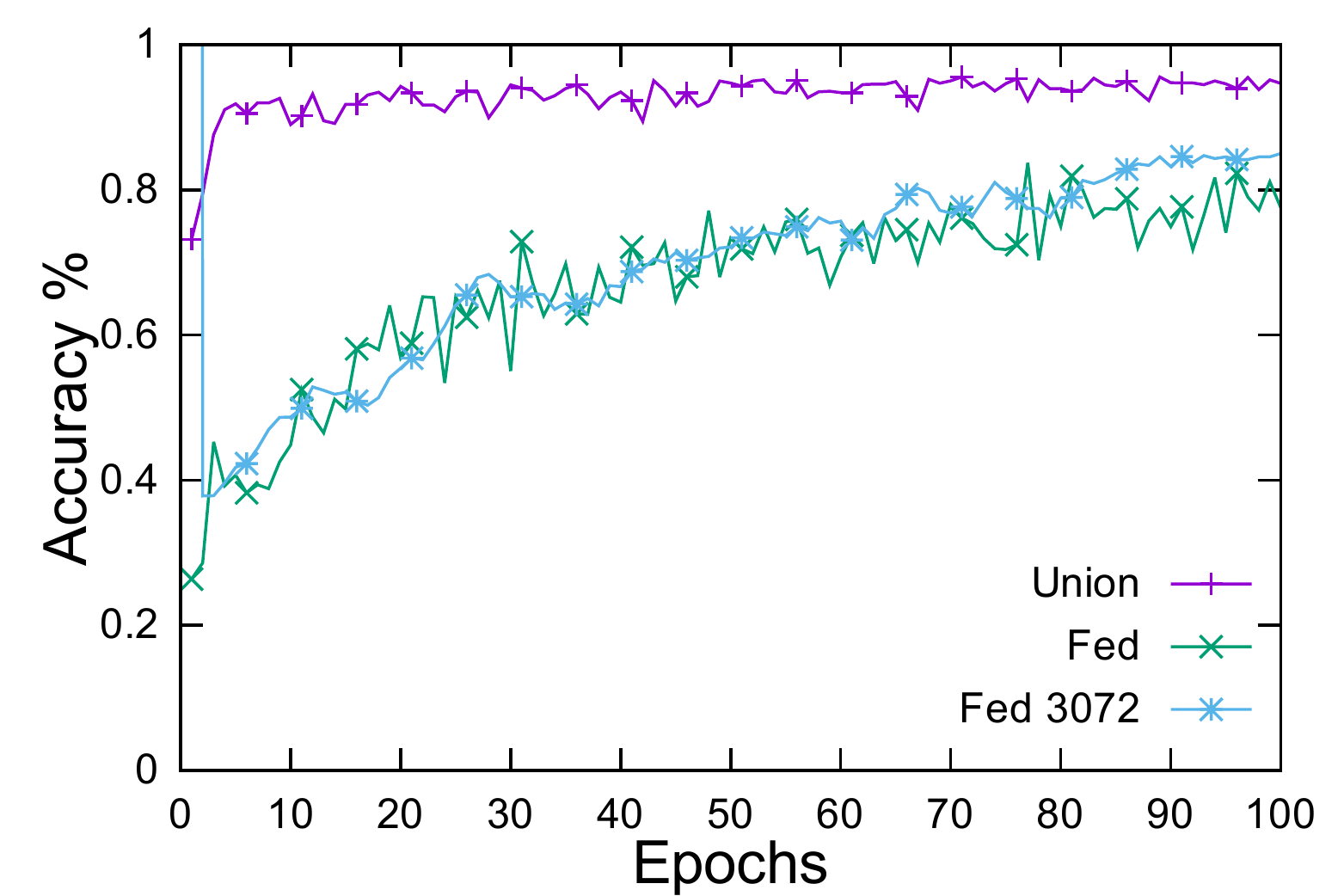}
                 \caption{PM0}
                 \label{fig:res_porcinemeat_100a}
         \end{subfigure}
         \begin{subfigure}[b]{0.48\columnwidth}
                 \centering
                 \includegraphics[width=\textwidth]{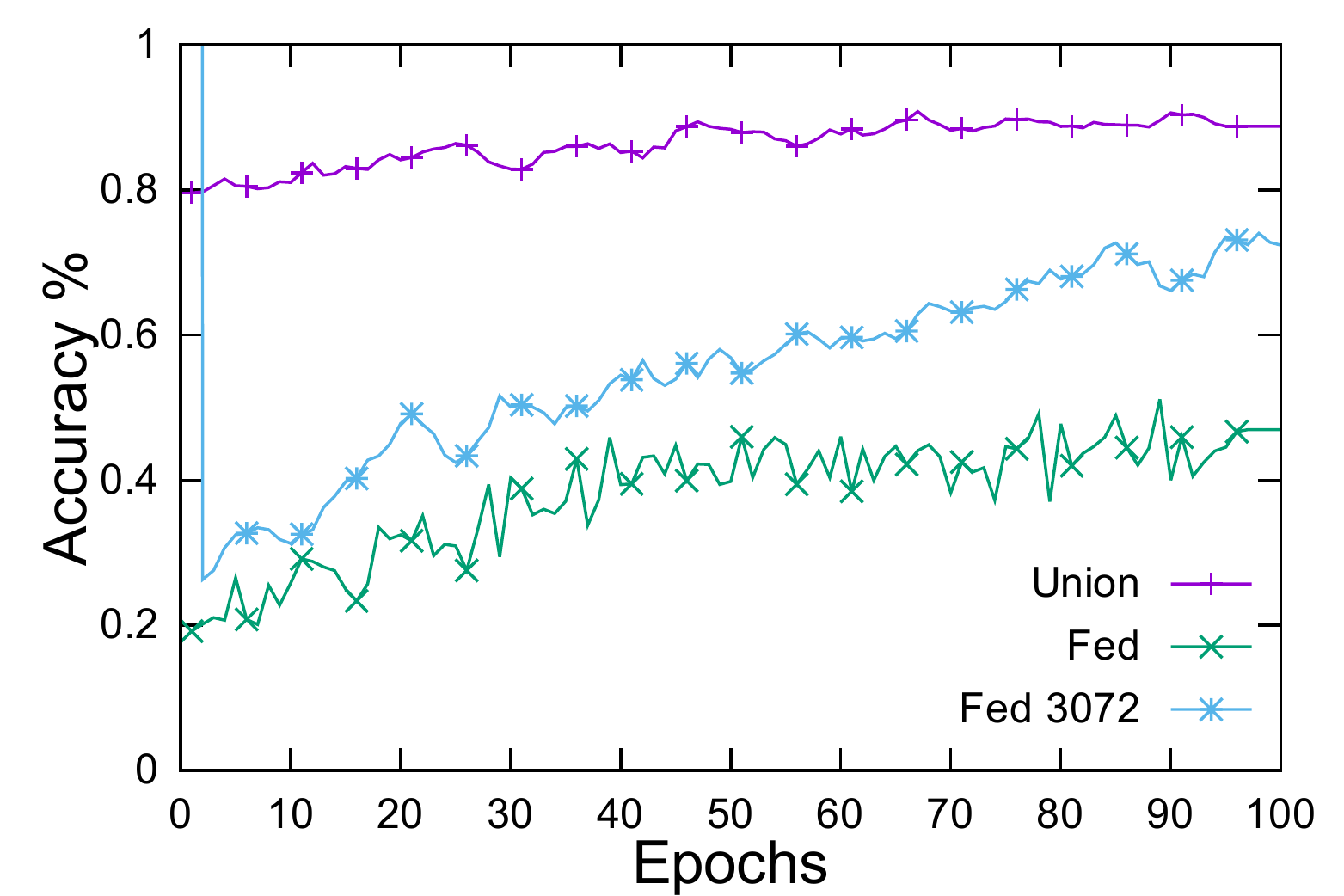}
                 \caption{PM1}
                 %\label{}
         \end{subfigure}
% leave a blank line to change row         

     \begin{subfigure}[b]{0.48\columnwidth}
             \centering
             \includegraphics[width=\textwidth]{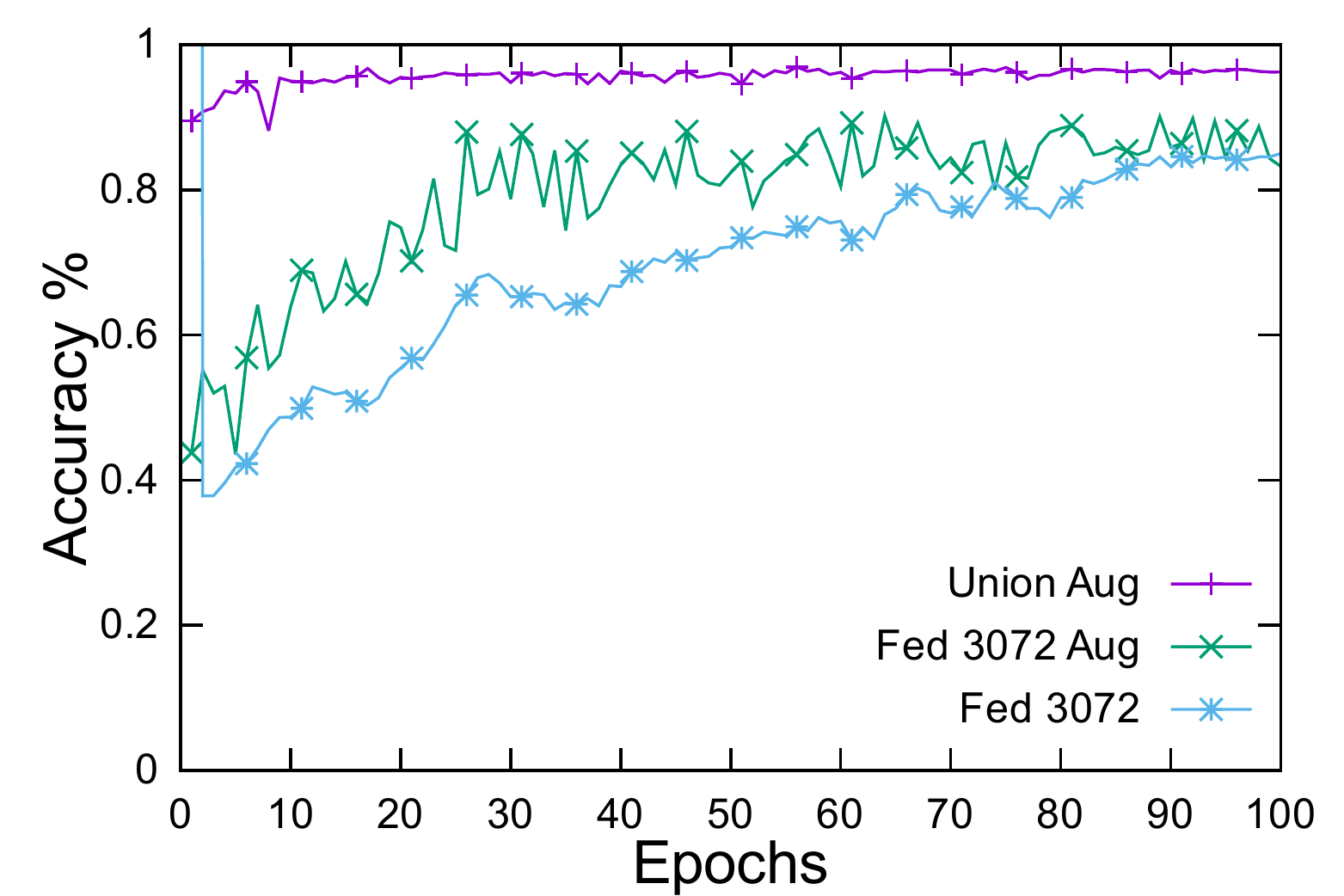}
             \caption{PM0 with Data Augmentation}
             %\label{}
     \end{subfigure}
     \begin{subfigure}[b]{0.48\columnwidth}
             \centering
             \includegraphics[width=\textwidth]{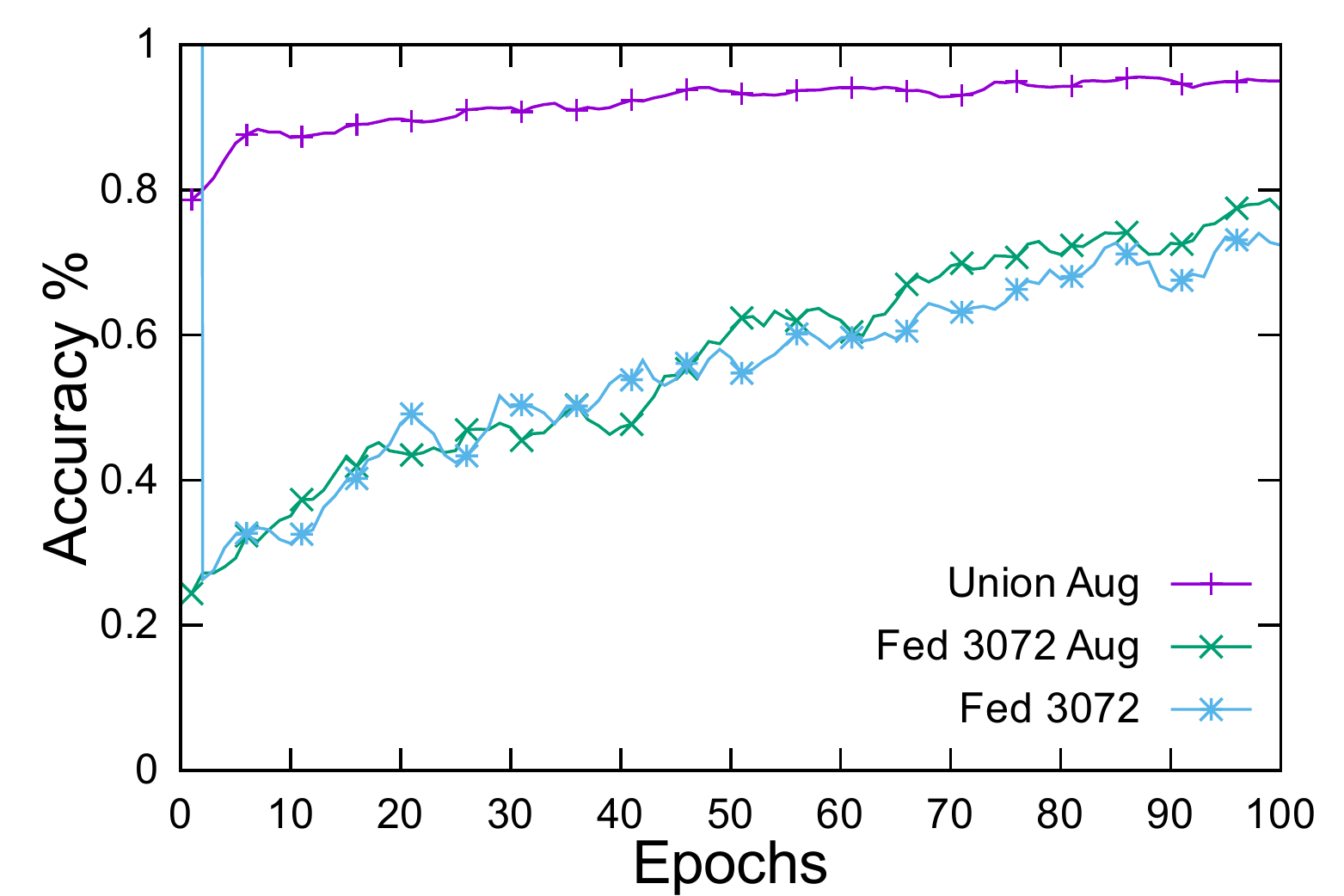}
             \caption{PM1 with Data Augmentation}
             %\label{}
      \end{subfigure}

\caption{Results with porcine meat with 100\% of data.} \label{fig:res_porcinemeat_100}
\end{figure}
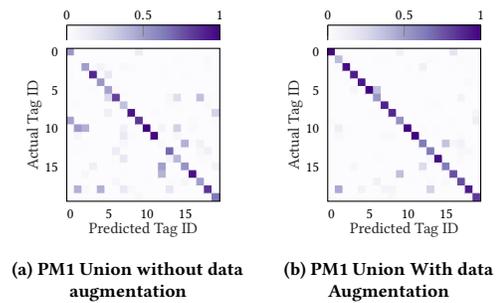
\begin{figure}[h]
    \centering
    
    \captionsetup{justification=centering}
    \begin{subfigure}[t]{0.40\columnwidth}
    \centering
    \setlength\fwidth{.6\columnwidth}
    \setlength\fheight{.6\columnwidth}
    \begin{tikzpicture}
\pgfplotsset{every tick label/.append style={font=\tiny}}

\begin{axis}[
enlargelimits=false,
colorbar,
colormap/Purples,
width=\fwidth,
height=\fheight,
at={(0\fwidth,0\fheight)},
scale only axis,
tick align=inside,
xlabel={Predicted Tag ID},
xmin=-0.5,
xmax=19.5,
xtick style={draw=none},
xlabel style={font=\scriptsize\color{white!15!black}},
ylabel style={font=\scriptsize\color{white!15!black}},
ylabel={Actual Tag ID},
ymin=-0.5,
ymax=19.5,
xlabel shift=-5pt,
ylabel shift=-5pt,
ytick style={draw=none},
axis background/.style={fill=white},
colorbar horizontal,
colorbar style={
at={(0,1.05)},               % <-- (changed)
anchor=below south west,    % <-- (changed)
% change the width of the colorbar relative to the main `axis' environment
width=\pgfkeysvalueof{/pgfplots/parent axis width},
xtick={0, 0.5, 1},
xmin=0,
xmax=1,
axis x line*=top,
xticklabel shift=-1pt,
point meta min=0,
point meta max=1,
},
colorbar/width=2mm,
]\addplot [matrix plot,point meta=explicit]
 coordinates {
(0,0) [0.5159362549800797] (0,1) [0.0] (0,2) [0.0] (0,3) [0.013280212483399735] (0,4) [0.0] (0,5) [0.0] (0,6) [0.0] (0,7) [0.0] (0,8) [0.0] (0,9) [0.4707835325365206] (0,10) [0.0] (0,11) [0.0] (0,12) [0.0] (0,13) [0.0] (0,14) [0.0] (0,15) [0.0] (0,16) [0.0] (0,17) [0.0] (0,18) [0.0] (0,19) [0.0] 

(1,0) [0.0] (1,1) [0.003360215053763441] (1,2) [0.0] (1,3) [0.0] (1,4) [0.0003360215053763441] (1,5) [0.0] (1,6) [0.003696236559139785] (1,7) [0.01478494623655914] (1,8) [0.0] (1,9) [0.0] (1,10) [0.5] (1,11) [0.054771505376344086] (1,12) [0.0] (1,13) [0.008064516129032258] (1,14) [0.0] (1,15) [0.0] (1,16) [0.0] (1,17) [0.0] (1,18) [0.41498655913978494] (1,19) [0.0] 

(2,0) [0.0] (2,1) [0.0] (2,2) [0.4842387659289068] (2,3) [0.0] (2,4) [0.0013413816230717639] (2,5) [0.0] (2,6) [0.0] (2,7) [0.0] (2,8) [0.004024144869215292] (2,9) [0.0] (2,10) [0.4242119382964453] (2,11) [0.0] (2,12) [0.01039570757880617] (2,13) [0.0] (2,14) [0.0] (2,15) [0.0] (2,16) [0.04560697518443997] (2,17) [0.0] (2,18) [0.030181086519114688] (2,19) [0.0] 

(3,0) [0.0] (3,1) [0.0] (3,2) [0.0] (3,3) [0.8709677419354839] (3,4) [0.0] (3,5) [0.020161290322580645] (3,6) [0.0006720430107526882] (3,7) [0.0] (3,8) [0.0] (3,9) [0.0] (3,10) [0.0] (3,11) [0.0] (3,12) [0.0] (3,13) [0.0] (3,14) [0.0] (3,15) [0.1081989247311828] (3,16) [0.0] (3,17) [0.0] (3,18) [0.0] (3,19) [0.0] 

(4,0) [0.0] (4,1) [0.0] (4,2) [0.07224462365591398] (4,3) [0.0] (4,4) [0.5077284946236559] (4,5) [0.0] (4,6) [0.0] (4,7) [0.0] (4,8) [0.0020161290322580645] (4,9) [0.0] (4,10) [0.0] (4,11) [0.0] (4,12) [0.0] (4,13) [0.0] (4,14) [0.0] (4,15) [0.0] (4,16) [0.0] (4,17) [0.0] (4,18) [0.41801075268817206] (4,19) [0.0] 

(5,0) [0.0] (5,1) [0.0003360215053763441] (5,2) [0.005376344086021506] (5,3) [0.15356182795698925] (5,4) [0.0006720430107526882] (5,5) [0.4734543010752688] (5,6) [0.24899193548387097] (5,7) [0.0] (5,8) [0.11323924731182795] (5,9) [0.0] (5,10) [0.0] (5,11) [0.0] (5,12) [0.0] (5,13) [0.004032258064516129] (5,14) [0.0] (5,15) [0.0] (5,16) [0.0003360215053763441] (5,17) [0.0] (5,18) [0.0] (5,19) [0.0] 

(6,0) [0.0] (6,1) [0.0003360215053763441] (6,2) [0.0] (6,3) [0.0010080645161290322] (6,4) [0.0003360215053763441] (6,5) [0.01881720430107527] (6,6) [0.7906586021505376] (6,7) [0.0] (6,8) [0.0] (6,9) [0.0] (6,10) [0.0] (6,11) [0.0] (6,12) [0.0] (6,13) [0.1364247311827957] (6,14) [0.009408602150537635] (6,15) [0.0] (6,16) [0.0] (6,17) [0.043010752688172046] (6,18) [0.0] (6,19) [0.0] 

(7,0) [0.0] (7,1) [0.001342732460557234] (7,2) [0.03591809331990601] (7,3) [0.0] (7,4) [0.0] (7,5) [0.0] (7,6) [0.00906344410876133] (7,7) [0.3544813695871098] (7,8) [0.0003356831151393085] (7,9) [0.0] (7,10) [0.09231285666330984] (7,11) [0.14602215508559918] (7,12) [0.031889895938234304] (7,13) [0.0] (7,14) [0.16381336018798254] (7,15) [0.0] (7,16) [0.008056394763343404] (7,17) [0.0030211480362537764] (7,18) [0.1537428667338033] (7,19) [0.0] 

(8,0) [0.0] (8,1) [0.0] (8,2) [0.01478494623655914] (8,3) [0.0] (8,4) [0.0] (8,5) [0.0010080645161290322] (8,6) [0.0] (8,7) [0.0] (8,8) [0.9307795698924731] (8,9) [0.0] (8,10) [0.0] (8,11) [0.021169354838709676] (8,12) [0.0] (8,13) [0.0] (8,14) [0.0] (8,15) [0.0] (8,16) [0.0] (8,17) [0.0] (8,18) [0.03125] (8,19) [0.0010080645161290322] 

(9,0) [0.13723511604439959] (9,1) [0.0] (9,2) [0.0] (9,3) [0.0] (9,4) [0.0] (9,5) [0.0] (9,6) [0.0] (9,7) [0.0] (9,8) [0.0] (9,9) [0.8213925327951564] (9,10) [0.0] (9,11) [0.0] (9,12) [0.0] (9,13) [0.0] (9,14) [0.0] (9,15) [0.0013454423141607804] (9,16) [0.0003363605785401951] (9,17) [0.0] (9,18) [0.0] (9,19) [0.03969054826774302] 

(10,0) [0.0] (10,1) [0.0] (10,2) [0.0003370407819346141] (10,3) [0.0] (10,4) [0.0] (10,5) [0.0] (10,6) [0.0] (10,7) [0.0003370407819346141] (10,8) [0.0] (10,9) [0.0] (10,10) [0.9915739804516347] (10,11) [0.0] (10,12) [0.0] (10,13) [0.0] (10,14) [0.0] (10,15) [0.0] (10,16) [0.0006740815638692282] (10,17) [0.007077856420626896] (10,18) [0.0] (10,19) [0.0] 

(11,0) [0.0] (11,1) [0.0] (11,2) [0.0] (11,3) [0.0] (11,4) [0.0] (11,5) [0.0] (11,6) [0.004368279569892473] (11,7) [0.004032258064516129] (11,8) [0.017809139784946238] (11,9) [0.0] (11,10) [0.0] (11,11) [0.9704301075268817] (11,12) [0.0] (11,13) [0.0] (11,14) [0.0] (11,15) [0.0] (11,16) [0.0] (11,17) [0.0] (11,18) [0.003360215053763441] (11,19) [0.0] 

(12,0) [0.0] (12,1) [0.0] (12,2) [0.0053655264922870555] (12,3) [0.005700871898054996] (12,4) [0.0] (12,5) [0.0] (12,6) [0.0] (12,7) [0.0] (12,8) [0.0] (12,9) [0.0] (12,10) [0.07142857142857142] (12,11) [0.0] (12,12) [0.029845741113346747] (12,13) [0.0] (12,14) [0.0] (12,15) [0.4782025486250838] (12,16) [0.409121395036888] (12,17) [0.00033534540576794097] (12,18) [0.0] (12,19) [0.0] 

(13,0) [0.0] (13,1) [0.004372687521022536] (13,2) [0.0] (13,3) [0.0] (13,4) [0.0] (13,5) [0.0] (13,6) [0.04742684157416751] (13,7) [0.020517995290951902] (13,8) [0.0] (13,9) [0.0] (13,10) [0.0] (13,11) [0.0] (13,12) [0.0] (13,13) [0.6737302388160108] (13,14) [0.06121762529431551] (13,15) [0.0] (13,16) [0.0] (13,17) [0.015472586612848975] (13,18) [0.17726202489068282] (13,19) [0.0] 

(14,0) [0.0] (14,1) [0.011772620248906828] (14,2) [0.0026908846283215607] (14,3) [0.0] (14,4) [0.0003363605785401951] (14,5) [0.0] (14,6) [0.21863437605112682] (14,7) [0.007063572149344097] (14,8) [0.0] (14,9) [0.0] (14,10) [0.28557013118062563] (14,11) [0.04305415405314497] (14,12) [0.004036326942482341] (14,13) [0.012108980827447022] (14,14) [0.39320551631348805] (14,15) [0.0] (14,16) [0.004036326942482341] (14,17) [0.012781701984527414] (14,18) [0.004709048099562731] (14,19) [0.0] 

(15,0) [0.0] (15,1) [0.0] (15,2) [0.0] (15,3) [0.007728494623655914] (15,4) [0.0] (15,5) [0.0] (15,6) [0.0] (15,7) [0.0] (15,8) [0.0] (15,9) [0.0] (15,10) [0.0] (15,11) [0.0] (15,12) [0.41397849462365593] (15,13) [0.0] (15,14) [0.0] (15,15) [0.5245295698924731] (15,16) [0.0] (15,17) [0.053763440860215055] (15,18) [0.0] (15,19) [0.0] 

(16,0) [0.005718129835183316] (16,1) [0.0] (16,2) [0.0026908846283215607] (16,3) [0.0] (16,4) [0.0003363605785401951] (16,5) [0.0] (16,6) [0.0] (16,7) [0.0] (16,8) [0.0] (16,9) [0.014799865455768583] (16,10) [0.028254288597376387] (16,11) [0.0] (16,12) [0.006727211570803902] (16,13) [0.0] (16,14) [0.0] (16,15) [0.0003363605785401951] (16,16) [0.9411368987554659] (16,17) [0.0] (16,18) [0.0] (16,19) [0.0] 

(17,0) [0.0] (17,1) [0.0] (17,2) [0.0] (17,3) [0.0] (17,4) [0.0] (17,5) [0.0] (17,6) [0.3648875461564283] (17,7) [0.0] (17,8) [0.0] (17,9) [0.0] (17,10) [0.007049345417925478] (17,11) [0.0003356831151393085] (17,12) [0.08996307485733468] (17,13) [0.03356831151393085] (17,14) [0.019805303793219202] (17,15) [0.0023497818059751594] (17,16) [0.0] (17,17) [0.482040953340047] (17,18) [0.0] (17,19) [0.0] 

(18,0) [0.0] (18,1) [0.023162134944612285] (18,2) [0.005035246727089627] (18,3) [0.0] (18,4) [0.0] (18,5) [0.0] (18,6) [0.01342732460557234] (18,7) [0.02685464921114468] (18,8) [0.0] (18,9) [0.0] (18,10) [0.0] (18,11) [0.0] (18,12) [0.0] (18,13) [0.09298422289358846] (18,14) [0.007385028533064787] (18,15) [0.0] (18,16) [0.0] (18,17) [0.0] (18,18) [0.8311513930849278] (18,19) [0.0] 

(19,0) [0.0] (19,1) [0.0] (19,2) [0.0] (19,3) [0.0] (19,4) [0.0] (19,5) [0.0] (19,6) [0.0] (19,7) [0.0] (19,8) [0.27251344086021506] (19,9) [0.014448924731182795] (19,10) [0.0] (19,11) [0.0] (19,12) [0.0] (19,13) [0.0] (19,14) [0.0] (19,15) [0.08333333333333333] (19,16) [0.0] (19,17) [0.0] (19,18) [0.0] (19,19) [0.6297043010752689] 
};
\end{axis}
\end{tikzpicture}
    \caption{PM1 Union without data augmentation}
    \setlength\belowcaptionskip{-.2cm}
    \label{fig:cm_1}
    \end{subfigure}
    \begin{subfigure}[t]{0.40\columnwidth}
    \centering
    \setlength\fwidth{.6\columnwidth}
    \setlength\fheight{.6\columnwidth}
    \begin{tikzpicture}
\pgfplotsset{every tick label/.append style={font=\tiny}}

\begin{axis}[
enlargelimits=false,
colorbar,
colormap/Purples,
width=\fwidth,
height=\fheight,
at={(0\fwidth,0\fheight)},
scale only axis,
tick align=inside,
xlabel={Predicted Tag ID},
xmin=-0.5,
xmax=19.5,
xtick style={draw=none},
xlabel style={font=\scriptsize\color{white!15!black}},
ylabel style={font=\scriptsize\color{white!15!black}},
ylabel={Actual Tag ID},
ymin=-0.5,
ymax=19.5,
xlabel shift=-5pt,
ylabel shift=-5pt,
ytick style={draw=none},
axis background/.style={fill=white},
colorbar horizontal,
colorbar style={
at={(0,1.05)},               % <-- (changed)
anchor=below south west,    % <-- (changed)
% change the width of the colorbar relative to the main `axis' environment
width=\pgfkeysvalueof{/pgfplots/parent axis width},
xtick={0, 0.5, 1},
xmin=0,
xmax=1,
axis x line*=top,
xticklabel shift=-1pt,
point meta min=0,
point meta max=1,
},
colorbar/width=2mm,
]\addplot [matrix plot,point meta=explicit]
 coordinates {
(0,0) [0.9681274900398407] (0,1) [0.0] (0,2) [0.0] (0,3) [0.0] (0,4) [0.0] (0,5) [0.0] (0,6) [0.0] (0,7) [0.0] (0,8) [0.0] (0,9) [0.01693227091633466] (0,10) [0.0] (0,11) [0.00099601593625498] (0,12) [0.0] (0,13) [0.0] (0,14) [0.0] (0,15) [0.0] (0,16) [0.0] (0,17) [0.0] (0,18) [0.0] (0,19) [0.013944223107569721] 

(1,0) [0.0] (1,1) [0.37348790322580644] (1,2) [0.015120967741935484] (1,3) [0.0005040322580645161] (1,4) [0.0] (1,5) [0.0] (1,6) [0.0025201612903225806] (1,7) [0.0] (1,8) [0.0] (1,9) [0.0] (1,10) [0.125] (1,11) [0.038306451612903226] (1,12) [0.0010080645161290322] (1,13) [0.0005040322580645161] (1,14) [0.0] (1,15) [0.0] (1,16) [0.0] (1,17) [0.0] (1,18) [0.4435483870967742] (1,19) [0.0] 

(2,0) [0.0] (2,1) [0.014084507042253521] (2,2) [0.9567404426559356] (2,3) [0.0] (2,4) [0.011569416498993963] (2,5) [0.0] (2,6) [0.0] (2,7) [0.0] (2,8) [0.0] (2,9) [0.0] (2,10) [0.013078470824949699] (2,11) [0.0] (2,12) [0.002012072434607646] (2,13) [0.0] (2,14) [0.0] (2,15) [0.0005030181086519115] (2,16) [0.0] (2,17) [0.0] (2,18) [0.002012072434607646] (2,19) [0.0] 

(3,0) [0.0] (3,1) [0.0] (3,2) [0.0] (3,3) [0.9022177419354839] (3,4) [0.0] (3,5) [0.055443548387096774] (3,6) [0.0] (3,7) [0.0] (3,8) [0.0] (3,9) [0.0] (3,10) [0.0] (3,11) [0.0] (3,12) [0.0005040322580645161] (3,13) [0.0] (3,14) [0.0] (3,15) [0.041834677419354836] (3,16) [0.0] (3,17) [0.0] (3,18) [0.0] (3,19) [0.0] 

(4,0) [0.0] (4,1) [0.0] (4,2) [0.0005040322580645161] (4,3) [0.0] (4,4) [0.9213709677419355] (4,5) [0.0] (4,6) [0.0] (4,7) [0.0] (4,8) [0.0] (4,9) [0.0] (4,10) [0.0] (4,11) [0.0] (4,12) [0.0] (4,13) [0.0005040322580645161] (4,14) [0.0] (4,15) [0.0] (4,16) [0.0] (4,17) [0.0] (4,18) [0.07762096774193548] (4,19) [0.0] 

(5,0) [0.0] (5,1) [0.0] (5,2) [0.0010080645161290322] (5,3) [0.0005040322580645161] (5,4) [0.0] (5,5) [0.9909274193548387] (5,6) [0.0] (5,7) [0.0] (5,8) [0.0045362903225806455] (5,9) [0.0] (5,10) [0.0] (5,11) [0.0025201612903225806] (5,12) [0.0] (5,13) [0.0] (5,14) [0.0005040322580645161] (5,15) [0.0] (5,16) [0.0] (5,17) [0.0] (5,18) [0.0] (5,19) [0.0] 

(6,0) [0.0] (6,1) [0.009072580645161291] (6,2) [0.0] (6,3) [0.0] (6,4) [0.0] (6,5) [0.36189516129032256] (6,6) [0.5186491935483871] (6,7) [0.0010080645161290322] (6,8) [0.0045362903225806455] (6,9) [0.0010080645161290322] (6,10) [0.0] (6,11) [0.0025201612903225806] (6,12) [0.0] (6,13) [0.005040322580645161] (6,14) [0.0045362903225806455] (6,15) [0.0] (6,16) [0.0] (6,17) [0.0907258064516129] (6,18) [0.0010080645161290322] (6,19) [0.0] 

(7,0) [0.0] (7,1) [0.006042296072507553] (7,2) [0.0] (7,3) [0.0] (7,4) [0.0005035246727089627] (7,5) [0.0] (7,6) [0.0] (7,7) [0.8877139979859013] (7,8) [0.0] (7,9) [0.0] (7,10) [0.0005035246727089627] (7,11) [0.030715005035246726] (7,12) [0.0025176233635448137] (7,13) [0.0] (7,14) [0.007049345417925478] (7,15) [0.0] (7,16) [0.0] (7,17) [0.0] (7,18) [0.0649546827794562] (7,19) [0.0] 

(8,0) [0.0] (8,1) [0.0] (8,2) [0.0010080645161290322] (8,3) [0.0] (8,4) [0.0030241935483870967] (8,5) [0.0005040322580645161] (8,6) [0.0] (8,7) [0.0] (8,8) [0.873991935483871] (8,9) [0.0] (8,10) [0.0] (8,11) [0.028225806451612902] (8,12) [0.0] (8,13) [0.0005040322580645161] (8,14) [0.0] (8,15) [0.0] (8,16) [0.0] (8,17) [0.0] (8,18) [0.05342741935483871] (8,19) [0.03931451612903226] 

(9,0) [0.09838546922300706] (9,1) [0.0] (9,2) [0.0] (9,3) [0.0] (9,4) [0.0] (9,5) [0.0] (9,6) [0.0] (9,7) [0.0] (9,8) [0.0] (9,9) [0.5383451059535822] (9,10) [0.0] (9,11) [0.0] (9,12) [0.0] (9,13) [0.0] (9,14) [0.0] (9,15) [0.0015136226034308778] (9,16) [0.3617558022199798] (9,17) [0.0] (9,18) [0.0] (9,19) [0.0] 

(10,0) [0.0] (10,1) [0.0010111223458038423] (10,2) [0.008088978766430738] (10,3) [0.0] (10,4) [0.0] (10,5) [0.0] (10,6) [0.0] (10,7) [0.0] (10,8) [0.0] (10,9) [0.0] (10,10) [0.9762386248736097] (10,11) [0.0] (10,12) [0.0015166835187057635] (10,13) [0.0] (10,14) [0.0] (10,15) [0.0] (10,16) [0.0] (10,17) [0.01314459049544995] (10,18) [0.0] (10,19) [0.0] 

(11,0) [0.0] (11,1) [0.0005040322580645161] (11,2) [0.0] (11,3) [0.0] (11,4) [0.0005040322580645161] (11,5) [0.0] (11,6) [0.0015120967741935483] (11,7) [0.0] (11,8) [0.0] (11,9) [0.0] (11,10) [0.0] (11,11) [0.9954637096774194] (11,12) [0.0] (11,13) [0.0020161290322580645] (11,14) [0.0] (11,15) [0.0] (11,16) [0.0] (11,17) [0.0] (11,18) [0.0] (11,19) [0.0] 

(12,0) [0.0] (12,1) [0.0] (12,2) [0.1006036217303823] (12,3) [0.004527162977867203] (12,4) [0.0] (12,5) [0.0] (12,6) [0.0] (12,7) [0.0] (12,8) [0.0] (12,9) [0.0] (12,10) [0.019617706237424547] (12,11) [0.0] (12,12) [0.6262575452716298] (12,13) [0.0] (12,14) [0.0] (12,15) [0.2369215291750503] (12,16) [0.012072434607645875] (12,17) [0.0] (12,18) [0.0] (12,19) [0.0] 

(13,0) [0.0] (13,1) [0.017658930373360242] (13,2) [0.0030272452068617556] (13,3) [0.0] (13,4) [0.0] (13,5) [0.0] (13,6) [0.008072653884964682] (13,7) [0.007063572149344097] (13,8) [0.0005045408678102926] (13,9) [0.0010090817356205853] (13,10) [0.0] (13,11) [0.0015136226034308778] (13,12) [0.0] (13,13) [0.627648839556004] (13,14) [0.034308779011099896] (13,15) [0.0] (13,16) [0.0] (13,17) [0.01866801210898083] (13,18) [0.2805247225025227] (13,19) [0.0] 

(14,0) [0.0] (14,1) [0.005045408678102927] (14,2) [0.0010090817356205853] (14,3) [0.0] (14,4) [0.0] (14,5) [0.0] (14,6) [0.0035317860746720484] (14,7) [0.02119071644803229] (14,8) [0.0] (14,9) [0.0] (14,10) [0.005549949545913219] (14,11) [0.0025227043390514633] (14,12) [0.0030272452068617556] (14,13) [0.0015136226034308778] (14,14) [0.955095862764884] (14,15) [0.0] (14,16) [0.0] (14,17) [0.0015136226034308778] (14,18) [0.0] (14,19) [0.0] 

(15,0) [0.0] (15,1) [0.0] (15,2) [0.0] (15,3) [0.0] (15,4) [0.0] (15,5) [0.0] (15,6) [0.0] (15,7) [0.0] (15,8) [0.0] (15,9) [0.0] (15,10) [0.0] (15,11) [0.0] (15,12) [0.3467741935483871] (15,13) [0.0] (15,14) [0.0] (15,15) [0.6234879032258065] (15,16) [0.0] (15,17) [0.029737903225806453] (15,18) [0.0] (15,19) [0.0] 

(16,0) [0.19122098890010092] (16,1) [0.0] (16,2) [0.006054490413723511] (16,3) [0.0] (16,4) [0.0] (16,5) [0.0] (16,6) [0.0] (16,7) [0.0010090817356205853] (16,8) [0.0] (16,9) [0.0075681130171543895] (16,10) [0.0005045408678102926] (16,11) [0.0] (16,12) [0.04238143289606458] (16,13) [0.0] (16,14) [0.0] (16,15) [0.0] (16,16) [0.7512613521695257] (16,17) [0.0] (16,18) [0.0] (16,19) [0.0] 

(17,0) [0.0] (17,1) [0.017119838872104734] (17,2) [0.0] (17,3) [0.0] (17,4) [0.0] (17,5) [0.0] (17,6) [0.07150050352467271] (17,7) [0.0] (17,8) [0.0] (17,9) [0.0] (17,10) [0.0010070493454179255] (17,11) [0.022155085599194362] (17,12) [0.081067472306143] (17,13) [0.03272910372608258] (17,14) [0.013595166163141994] (17,15) [0.004028197381671702] (17,16) [0.0] (17,17) [0.756797583081571] (17,18) [0.0] (17,19) [0.0] 

(18,0) [0.0] (18,1) [0.010070493454179255] (18,2) [0.006042296072507553] (18,3) [0.0] (18,4) [0.0256797583081571] (18,5) [0.0] (18,6) [0.0015105740181268882] (18,7) [0.0010070493454179255] (18,8) [0.0] (18,9) [0.0] (18,10) [0.0] (18,11) [0.0] (18,12) [0.0005035246727089627] (18,13) [0.0377643504531722] (18,14) [0.0] (18,15) [0.0] (18,16) [0.0] (18,17) [0.0] (18,18) [0.9174219536757301] (18,19) [0.0] 

(19,0) [0.0] (19,1) [0.0010080645161290322] (19,2) [0.0] (19,3) [0.006048387096774193] (19,4) [0.0] (19,5) [0.0] (19,6) [0.0] (19,7) [0.0005040322580645161] (19,8) [0.055443548387096774] (19,9) [0.1184475806451613] (19,10) [0.0] (19,11) [0.0] (19,12) [0.0] (19,13) [0.0] (19,14) [0.0] (19,15) [0.0020161290322580645] (19,16) [0.0] (19,17) [0.0] (19,18) [0.0] (19,19) [0.8165322580645161] 

};
\end{axis}
\end{tikzpicture}
    \caption{PM1 Union With data Augmentation}
    \label{fig:cm_2}
    \end{subfigure}
    \centering
    \captionsetup{justification=centering}
    \setlength\belowcaptionskip{-.2cm}
    \caption{Porcine Meat}
        \label{fig:cms_20tags}
\end{figure}
\section{Conclusions}

In this paper we propose the first large-scale investigation into \gls{rfp} of RFID tags with dynamic channel conditions.
This investigation is composed by three main contributions: first, a massive data collection campaign on a testbed composed by 200 off-the-shelf identical RFID tags and a \gls{sdr} tag reader, even with porcine meat as communication obstacle to emulate an implanted RFID. 
Second, the definition and the relative evaluation on the dataset of several \gls{cnn}-based classifiers in a variety of channel conditions. Our investigation reveals that training and testing on different channel conditions drastically degrades the classifier's accuracy.
Third, a novel training framework based on \gls{fml} and \gls{da} to boost the accuracy.
Extensive experimental results indicate that (i) our \gls{fml} approach improves accuracy by up to 48\% with respect to the single-dataset scenario; and (ii) our \gls{da} approach improves the accuracy by up to 31\%.
Results also highlight how \gls{fml} and \gls{da} should be applied to tackle different problems, and have different results depending on the scenario and the number of tags.
To the best of our knowledge, this is the first paper demonstrating the efficacy of \gls{fml} and data augmentation on such rich datasets and large population. Conversely from existing work, in this work we describe and share with the research community our fully-labeled 200-GB RFID waveform dataset, as well as our code and trained models to allow complete replicability and verification of results.


\begin{acks}
This work was carried out within the research project "SMARTOUR: intelligent platform for tourism" funded by the Ministry of University and Research with the Regional Development Fund of European Union (PON Research and Competitiveness 2007-2013).
\end{acks}

\bibliographystyle{ieeetr}
\bibliography{acmart}

\end{document}